\documentclass[manuscript,screen,author]{acmart}
\usepackage{graphicx}
\usepackage[compact]{titlesec}

\titlespacing{\subsubsection}{0pt}{*0}{*0}
\usepackage{wrapfig}
\usepackage{lscape}
\usepackage{caption}
\usepackage{subcaption}
\usepackage{multirow}
\usepackage{ragged2e}
\usepackage{booktabs}
\usepackage{tabularx}
\usepackage{rotating}
\usepackage{algorithmic}
\usepackage{tikz}
\usepackage{amsmath,amsfonts,amsthm}
\usepackage{mathtools,nccmath}
\usepackage{mathrsfs}

\usepackage{xcolor}
\def\BibTeX{{\rm B\kern-.05em{\sc i\kern-.025em b}\kern-.08em
    T\kern-.1667em\lower.7ex\hbox{E}\kern-.125emX}}
\usepackage{mfirstuc}
\DeclareMathOperator{\E}{\mathbb{E}}

\newcommand{\Barlow}{Barlow-Twins}

\newcommand{\tikzcircle}[2][red,fill=red]{\tikz[baseline=-0.5ex]\draw[#1,radius=#2] (0,0) circle ;}%
\newcommand{\modimage}{\tikzcircle[cyan, fill=cyan]{4pt} }
\newcommand{\modtext}{\tikzcircle[pink, fill=pink]{4pt} }
\newcommand{\modvideo}{\tikzcircle[gray, fill=gray]{4pt} }
\newcommand{\modaudio}{\tikzcircle[yellow, fill=yellow]{4pt} }
\newcommand{\modsensors}{\tikzcircle[teal, fill=teal]{4pt} }

\AtBeginDocument{%
  \providecommand\BibTeX{{%
    \normalfont B\kern-0.5em{\scshape i\kern-0.25em b}\kern-0.8em\TeX}}}

\setcopyright{acmcopyright}
\copyrightyear{2022}
\acmYear{2022}
\acmDOI{00.0000/0000000.0000000}

\begin{document}

\title[Beyond Vision: A Review on SSRL]{Beyond Just Vision: A Review on Self-Supervised Representation Learning on Multimodal and Temporal Data}

\author{Shohreh Deldari}
\email{shohreh.deldari@student.rmit.edu.au}
\orcid{0000-0001-8150-120X}
\affiliation{%
  \institution{School of Computing and Technologies, RMIT University}
  \city{Melbourne}
  \state{Victoria}
  \country{Australia}
  \postcode{3000}
}

\author{Hao Xue}
\affiliation{%
  \institution{School of Computer Science and Engineering, University of New South Wales}
  \streetaddress{}
  \city{Sydeny}
  \state{NSW}
  \country{Australia}}
\email{hao.xue1@unsw.edu.au}
\orcid{0000-0003-1700-9215}

\author{Aaqib Saeed}
\affiliation{%
  \institution{Philips Research}
  \city{Eindhoven}
  \country{Netherlands}
}
\orcid{0000-0003-1473-0322}
\email{aaqib.saeed@philips.com}

\author{Jiayuan He}
\affiliation{%
 \institution{School of Computing and Technologies, RMIT University}
 \city{Melbourne}
 \state{Victoria}
 \country{Australia}}
 \email{jiayuan.he@rmit.edu.au}
 \orcid{0000-0002-8994-9532}

\author{Daniel V. Smith}
\affiliation{%
  \institution{Data61, CSIRO}
  \city{Hobart}
  \state{Tasmania}
  \country{Australia}}
\email{daniel.v.smith@csiro.au}

\author{Flora D. Salim}
\affiliation{%
  \institution{School of Computer Science and Engineering, University of New South Wales}
  \city{Sydney}
  \state{NSW}
  \country{Australia}
  \postcode{2000}}
\email{flora.salim@unsw.edu.au}
\orcid{0000-0002-1237-1664}

\renewcommand{\shortauthors}{Deldari et al.}

\begin{abstract}

Recently, Self-Supervised Representation Learning (SSRL) has attracted much attention in the field of computer vision, speech, natural language processing (NLP), and recently, with other types of modalities, including time series from sensors.
The popularity of self-supervised learning is driven by the fact that traditional models typically require a huge amount of well-annotated data for training. Acquiring annotated data can be a difficult and costly process. Self-supervised methods have been introduced to improve the efficiency of training data through discriminative pre-training of models using supervisory signals that have been freely obtained from the raw data. Unlike existing reviews of SSRL that have pre-dominately focused upon methods in the fields of CV or NLP for a single modality, we aim to provide the first comprehensive review of multimodal self-supervised learning methods for temporal data. To this end, we 1) provide a comprehensive categorization of existing SSRL methods, 2) introduce a generic pipeline by defining the key components of a SSRL framework, 3) compare existing models in terms of their objective function, network architecture and potential applications, and 4) review existing multimodal techniques in each category and various modalities. Finally, we present existing weaknesses and future opportunities. We believe our work develops a perspective on the requirements of SSRL in domains that utilise multimodal and/or temporal data.

\end{abstract}


\maketitle

\section{Introduction}
The past decades have witnessed the great success of deep learning techniques, which has led to a proliferation of applications in which large datasets are available. Despite the significant impact of deep learning, many state-of-the-art techniques still require human intervention, e.g. manual data preprocessing and data annotation. This is a major bottleneck for supervised learning models given acquiring data annotations can often be an unwieldy process that requires some domain knowledge. Self-supervised representation learning (SSRL) address this bottleneck by attempting to uncover meaningful information about the data through training the network model with a supervisory signal obtained from the data itself. This significantly increases the data bandwidth available for training the model and has been shown to reduce the reliance on manual data annotations \cite{henaff2020data}. This might also be considered an early step in the path towards general artificial intelligence, given the computer learns from the observed data with far less human input than a supervised learning approach. In what next, we give an overview of \emph{representation learning} and related \emph{self-supervised learning} techniques.

\begin{figure}
\small
    \centering
    \subfloat[]{\includegraphics[width=0.35\textwidth]{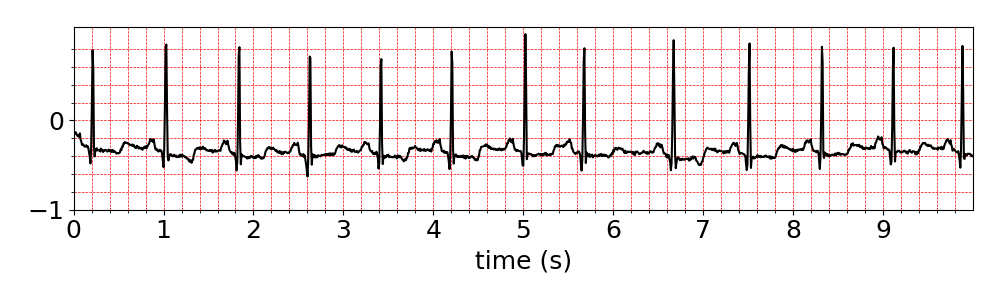}\label{fig:ecg_arr_wave}}
    \subfloat[]{\includegraphics[width=0.12\textwidth]{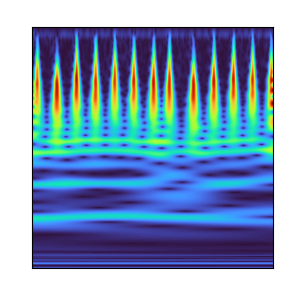}\label{fig:ecg_arr_scalo}}
    \hfill
    \subfloat[]{\includegraphics[width=0.35\textwidth]{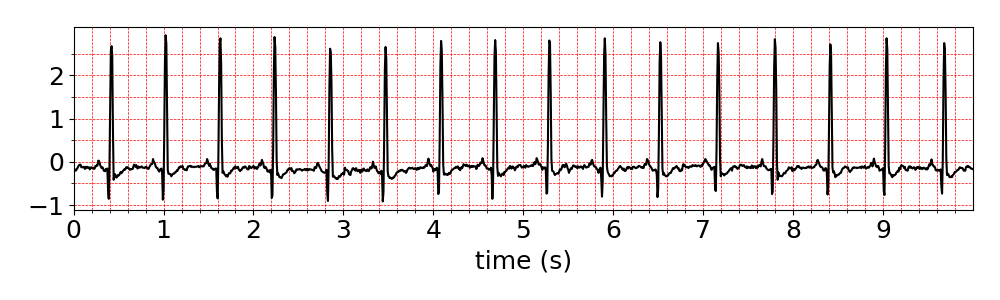}\label{fig:ecg_nsr_wave}}
    \subfloat[]{\includegraphics[width=0.12\textwidth]{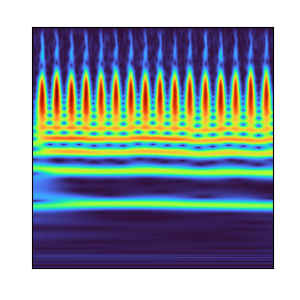}\label{fig:ecg_nsr_scalo}}
    \caption{Electrocardiogram (ECG) signals in two different views: (1) A sample ECG waveform of a) cardiac arrhythmia (ARR) and c)  Normal Sinus Rhythm (NSR); and (2) scalogram of ARR(b) and NSR (d) using continuous wavelet transform. }
    \label{fig:timeseries}
\end{figure}

\subsection{Representation Learning}
How input data is represented usually has a crucial impact on the performance of machine learning models. Therefore, the important ingredients of a successful machine learning model are preprocessing the input data and finding effective representation for it so that informative features can be easily extracted by a model later during training~\cite{bengio2013representation}. Fig.~\ref{fig:timeseries} presents an example of time-series data and shows two electrocardiogram (ECG) samples, one obtained from a patient with cardiac arrhythmia and the other with a normal sinus rhythm. Although representing the ECG samples in one-dimensional waveform (see Fig.~\ref{fig:ecg_arr_wave} and Fig.~\ref{fig:ecg_nsr_wave}) provides a straightforward view of the ECG signals, this representation is usually noisy due to the existence of various types of interference that can occur during ECG monitoring, such as baseline wanders, power line interference and machine interference. Performing the wavelet transform over the ECG sample and representing the signal in the time-frequency domain (see Fig.~\ref{fig:ecg_arr_scalo} and Fig.~\ref{fig:ecg_nsr_scalo}), however, can facilitate the model in attenuating this noise and extracting discriminative features from ECG signals~\cite{SAHOO201755,zhao2020ecg}.

Research on data representation initially relied on feature engineering, i.e. defining manual steps for transforming raw data into meaningful features. Examples of such transformation include the computation of statistical properties (e.g. mean and standard deviation) and shape- or pattern-based characteristics (e.g. wavelet transformation in Fig.~\ref{fig:timeseries}). However, designing a good selection of features for a complex task usually requires a huge amount of human effort and substantial domain knowledge to ensure the quality of the features. In addition, the features selected for one task may not be suitable for another task, making it infeasible to generalise using these approaches. 
To this end, representation learning, which aims to automatically extract discriminative features from data, has been proposed. The rationale is that dimensionality reduction algorithms, e.g. principle component analysis, manifold learning and deep learning will preserve only the most informative features of high-dimensional data and eliminate noise in data when mapping it into its lower-dimensional representations. The preserved and denoised features can thereafter be used as inputs to subsequent machine learning models. Since representation learning is minimally dependent on human intervention, it is no longer perceived as a task for a machine to solve but as learning a skill (Satinder Singh) to develop a generic understanding of the world and converge to create a general common understanding \cite{lecun2021dark} that can be generalised to new scenarios (transfer learning). Therefore, representation learning has attracted extensive attention in many domains, including computer vision (CV), speech recognition and natural language processing (NLP). 

\subsection{Self-Supervised Representation Learning}
A supervised model takes an annotated dataset for a given task (e.g. ImageNet for image classification) and learns data representations. Due to the generality of the trained task, the learned representation can serve as a good starting point for a new but similar task (e.g. object detection). 
However, the main limitation of supervised representation learning is the requirement for annotated datasets. Data annotation is an expensive and time-consuming process that requires significant human effort. To acquire high-quality datasets, it requires extensive expertise in the data domain (e.g. medical, legal or biology datasets). In addition, for some application domains (e.g. human-related applications), it is infeasible to obtain labelled datasets due to privacy issues. 

Self-supervised representation learning is proposed to overcome the above issues. It works directly on unlabelled data without the need for external annotation. Similar to human beings' self-learning process, which is a repeated cycle of `\textit{Observation, Action, Observation and Inference}', SSRL takes original data as the input, applies one or more predefined actions on the input data, observes the outcome of actions and infers the next optimisation steps for the model parameters. Repeating this procedure, SSRL models can extract the distribution and local/global features of data. 

A critical component of SSRL is the predefined action on the data input, which generates the pseudo labels to supervise the training process. An example of such an action is the pretext task of masked learning, which masks a proportion of data, uses the masked part as a label and guides the model to recover the masked part. 

Since SSRL does not need external supervision, it can leverage the vast amount of available unlabelled data (e.g. all images posted on the internet). This has led to the remarkable success of SSRL. In the 2000s, Hinton et al. (2006)~\cite{hinton2006fast} and Bengio et al. (2007)~\cite{bengio2007greedy} proposed a self-supervised algorithm for pretraining Deep Belief Networks before fine-tuning the model using ground truth labels. They showed that using pretrained weights can improve the performance of the supervised task. Nowadays, with deep learning-based models, SSRL not only achieves competitive accuracy in many downstream tasks when compared with other fully supervised rival methods but also has the advantage of far fewer parameters and smaller networks, when compared with similar supervised models; therefore, they are less prone to over-fitting \cite{jing2020self}. In addition, use cases for SSRL are not limited to serving as a pre-processing step in the machine learning pipeline (e.g. transfer learning) but also include applications in other fields, such as continual learning \cite{rao2019continual} and reinforcement learning \cite{schwarzer2020data} by improving the label/reward efficiency of those learning systems.

\subsection{Self-Supervised Representation Learning of Multimodal and Temporal Data}
Multimodal and temporal data are ubiquitous in daily life. The widespread use of digital devices and proliferation of digital applications, e.g. IoT applications, wireless communications and media consumption, have led to the increasing availability of temporal data. Examples of temporal data include sensor readings, stock prices and medical records, just to name a few. Temporal data analysis has widespread application in various domains, including the environment (e.g. climate modelling), public safety (e.g. crime prediction) and intelligent transportation (e.g. urban traffic management). In addition, a large quantity of data is generated in heterogeneous formats and thus covers multiple data modalities. For example, autonomous cars rely on both vision and sensor data for decision making. These data from different sources can be seen as different viewpoints on the same topic and provide complementary information. Therefore, utilising these two types of data simultaneously can enhance the effectiveness and reliability of machine learning models.

Despite the importance of multimodal and temporal data, analysing them is nontrivial. Temporal data has unique characteristics when compared with other data types (e.g. image and text data), as it is strongly correlated with the collection time. It is critical to extract dynamic temporal patterns, e.g. periodic patterns, to discover how the data evolves over time. When there are multiple modalities of data available, it is critical to learn representations that are not only effective but also semantically consistent across different modalities. 

To address these issues, many SSRL models have been proposed that leverage the large amount of unlabelled multimodal and/or temporal data available to learn data representations that can be transferred to downstream predictors. These models differ across various aspects in terms of data modalities (e.g. audio, image, text or time series or combinations of theses modalities), methods for generating supervision signals (e.g. pretext tasks, clustering, contrasting data samples and similarity computation) and objective functions (e.g. cross-entropy, triplet loss and InfoNCE). In this paper, we provide a systematic review of the recent works on SSRL of multimodal and temporal data. We present an overview of these works, position and compare them, and point out potential future research directions.

\subsection{Motivation and Contributions}
In this paper, we introduce the rapidly evolving field of representation learning, and review SSRL techniques. In particular, we discuss modalities that are less covered by existing surveys (temporal data) and specifically focus on cross-modal learning models. 
This paper will benefit the research community with the following contributions:
\begin{itemize}
    \item To the best of our knowledge, this is the \emph{first} comprehensive survey of SSRL that covers multiple modalities beyond vision data. We include audio and time series data types and their combinations with vision and text. This will provide a panorama which researchers can quickly apprehend the state-of-the-art works in these fields.
    
    \item We propose a classification scheme to position, categorise and compare the reviewed works. The comparative analysis may serve as a practical guide for readers on which model to use in real-life situations.
    
    \item We identify open challenges in the fields of SSRL for multimodal and temporal data. We further discuss possible future research directions that could potentially address these challenges.
\end{itemize}

The rest of this article is organised as follows: In Section \ref{sec:related_works}, we review existing review papers on SSRL 
In Section \ref{sec:def_backgrounds}, we explain important terms and background information to make it easy to read and understand the paper. In Section \ref{sec:single}, we describe existing categories and specifically cover SSRL methods proposed for a single mode of temporal data. We expand the existing frameworks to cross-modal approaches in Section \ref{sec:multimodal} and also investigate possible frameworks, requirements and challenges of cross-modal applications. In Section \ref{sec:loss_fn}, we focus on the evolving trend of objective functions employed for self-supervised learning. Finally, Section \ref{sec:discussion} discusses challenges and future research directions.

\section{Related Surveys}
\label{sec:related_works}
There are several comprehensive reviews on SSRL models based on Generative Adversarial Networks \cite{gao2022generative,liu2021genvscon}, autoencoders \cite{guo2019deepmultimodal} and contrastive learning  \cite{le2020contrastive,jing2020self,Jaiswal2020survey} in the fields of CV and NLP. However, none of those concentrates on either multimodal or temporal data. They also do not discuss the inspiring ideas behind self-supervised learning objective functions and the evolvement of architecture designs. In this work, we provide a detailed and up-to-date review of the evolution of self-supervised methods for representation learning. Although we cover only the ground-breaking studies in NLP and CV, we mainly focus on techniques and advances in temporal and multimodal data regimes in recent years.
\begin{table}[]

\centering
\small
\caption{Comparison of our survey with existing representation learning surveys}
\label{tab:related_surveys}
\begin{tabularx}{\linewidth}{p{2.2cm}|c|ccccc|cccc|X} 
\hline
\multirow{5}{*}{publication} &
\multirow{2}{*}{\begin{sideways}Multimomdal\end{sideways}} &
\multicolumn{5}{c|}{modality} & 
\multicolumn{4}{c|}{Category} &  
\multirow{5}{*}{Details}\\ 
\cline{3-11}
 &
 &
\multicolumn{1}{c}{\begin{sideways}Image\end{sideways}} & \multicolumn{1}{c}{\begin{sideways}Video\end{sideways}} & \multicolumn{1}{c}{\begin{sideways}Audio\end{sideways}} & \multicolumn{1}{c}{\begin{sideways}Text\end{sideways}} & \multicolumn{1}{c|}{\begin{sideways}Time series\end{sideways}} &
\multicolumn{1}{c}{\begin{sideways}Pretext\end{sideways}} &
\multicolumn{1}{c}{\begin{sideways}Generative\end{sideways}} & \multicolumn{1}{c}{\begin{sideways}Contrastive\end{sideways}} & \multicolumn{1}{c|}{\begin{sideways}Reg.-based\end{sideways}} & \\ 
\hline
Bengio et. al., 2013 \cite{bengio2013representation}& - & x & x & x & x & x &  x & x & - & - &  Explores the key factors in making a good representations and appropriate objective functions for three paradigms including probabilistic models, reconstruction-based, and manifold-learning. \\
Guo et. al., 2019 \cite{guo2019deepmultimodal}& x & x & x & x & x & - &  - & x & - & - & Specificaly explores multimodal representation learning. They investigate different frameworks in fusing modalities and divide RL models into four categories of probablistic, generative, auto-encoders, and attention-based.\\
Le-Khac et. al., 2020 \cite{le2020contrastive}& x & x & x & x & x & - & x  & - & x & - & The paper defines contrastive learning workflow and provide comprehensive taxonomy for different parts of the workflow. However, they consider all categories as contrastive which is not accurate. \\
Jing et. al., 2020 \cite{jing2020self}& x & x & x & - & - & - & x & x  & x & - & They divide SSRL models into 4 groups of Generation, Context, Free semantic label, and crossmodal techniques based on the employed tasks to generate pseudo-labels.\\
Latif et. al., 2020 \cite{latif2020speechsurvey}& - & - & - & x & - & - &  - & x &- &- & Covers representation learning approaches in Automatic Speech Recognition (ASR), Speaker Recognition (SR), and Speaker Emotion Recognition (SER). \\
Jaiswal et. al., 2020 \cite{Jaiswal2020survey} & - &x & -  & -  & x  & -  & -  & - & x & - & Contrastive learning for visual application\\
Mao et. al., 2020 \cite{mao2020survey} & - & x & x & x  & x & -  & x  & x & x & - & Divides SSRL approaches into bottleneck and prediction-based models. Bottleneck techniques includes dimenssinality reduction and auto-encoder models, while prediction-based techniques encompass methods that are categorized as pretext and contrastive models in our survey. \\

Liu et. al., 2021 \cite{liu2021genvscon}& - & x & - & - & x & - &  x & x & x & - & Compares contrastive vs. gen. adverserial models. \\

Ericsson et. al., 2021 \cite{ericsson2021survey}& - & x & x & x & x & x & x &  & x & x & The most recent survey paper with comprehensive review over different types SSRL models. However they reviewed limited number of multimodal and temporal methods.\\

\hline
Ours & x & x & x & x & x & x & x & - & x & x & Explores crossmodal and temporal SSRL models along with evolution of self-supervised objective functions\\
\hline
\end{tabularx}
\end{table}

Focusing on multimodal representation learning, Guo et. al. \cite{guo2019deepmultimodal} investigated existing supervised and unsupervised approaches in three categories based on how different modalities were fused together. They suggested three categories, i.e. joint representation, coordinated representation, and encoder–decoder.
In recent years, following the introduction of the contrastive objective functions, the representation learning research community has witnessed increasing attention being paid to and progress in this field. Jaiswal et. al. \cite{Jaiswal2020survey} and Le et. al. \cite{le2020contrastive} published comprehensive reviews investigating existing contrastive learning-based models mostly covering works in the CV and NLP areas. 
In their recent survey on self-supervised visual feature learning, Jing et. al. \cite{jing2020self} covered current methods and the relevant datasets. They divided existing SSRL approaches into four different categories based on the pretext tasks: 1) generation-based, 2) context-based, 3) free semantic labels and 4) cross-modal pretext tasks. However, free semantic label-based methods cannot be categorised as fully self-supervised techniques since they require human annotation to design and extract semantic labels for pretext tasks, such as moving object segmentation and relative depth prediction. They also treated contrastive-based and clustering methods as contextual pretext tasks and did not cover other well-known non-contrastive methods, such as self-distillation approaches. 
In contrast, Liu et. al. \cite{liu2021genvscon} divided existing SSRL methods into three main categories based on the final objective function, namely contrastive, generative or generative–contrastive (adversarial) models. They focused on SSRL methods across various domains, including vision, NLP and graphs, but didn't cover techniques related to temporal (e.g. audio and time-series) and multimodal data. 

Similar to \cite{liu2021genvscon}, we divided current SSRL methods based on the objective function; however, we will extend the categories to cover pretext-, contrastive-, clustering- and regularization-based methods.
To provide a full comparison between the existing surveys in this topic and highlight the contribution of our work, Table \ref{tab:related_surveys} lists all existing review papers and compares them based on the modalities and categories they cover. 
Note that we do not cover Generative-adversarial Network (GAN) and Variational AutoEncoder (VAE) models as there are many works that reviewed generative and adversarial models for self-supervised techniques \cite{yang2021survey,liu2021genvscon,gao2022generative}.

    
    
    
    
    
    
    
    
    


\section{Self-Supervised Representation Learning: Definitions and Background}
\label{sec:def_backgrounds}
To make this survey easy to read, this section explains the terms and notations that will be used throughout this work and the necessary background and existing learning paradigms.

\subsection{Definitions}
\label{sec:definition}
Given a set of $N$ unlabeled training samples $X_{1}, X_{2},\cdots,X_{N}$ from multimodal sources of data, $X_{n} = {x_{n}^{1},.., x_{n}^{M}}$ ($n \in N$) contains data captured from $M$ various input sources, and $X_{n}^{m}$ denotes the observation of the $n^{th}$ sample on modality $m$. In the case of supervised learning, models consume the correspondent ground truth labels $Y=\{y_{1},y_{2},...,y_{N}\}$ for training; however, in self-supervised training/pretraining, ground truth labels are discarded. 

\begin{table*}
\begin{minipage}{0.65\textwidth}
\setlength{\tabcolsep}{1pt}
\centering
\caption{Definition of notations used in describing the problem and method}
\label{tab:notations}
\begin{tabular}{c|l}
\hline
\textbf{Notation} & \textbf{Meaning}                                                     \\ 
\hline
$V$                 & Number of modalities, views, or features         \\
$X$ & unlabeled input data \\
$Y'$ & Pseudo-label generated based on intrinsic structures of $X$\\
$X_{v}$       & \textit{Set of input samples from modality $v$.}   \\
$x_{v}^{t}$       & \textit{The $t$th sample from modality $v$.}   \\
$z_{v}^{t}$       & \textit{The corresponding representation for the $t$th sample of modality $v$.} \\
$f(.)$            & \textit{The Encoder function}    \\
$S_{v,w}^{t,t'}$  & \textit{Similarity between sample $x_{v}^{t}$ and $x_{w}^{t'}$}                            \\
$\tau$              & Temperature parameter (scale adjustment)  \\
$\mathcal{L}$       & Loss                  \\
\hline
\end{tabular}
\end{minipage}
\hfill
\begin{minipage}{0.34\textwidth}
 \includegraphics[width=\textwidth]{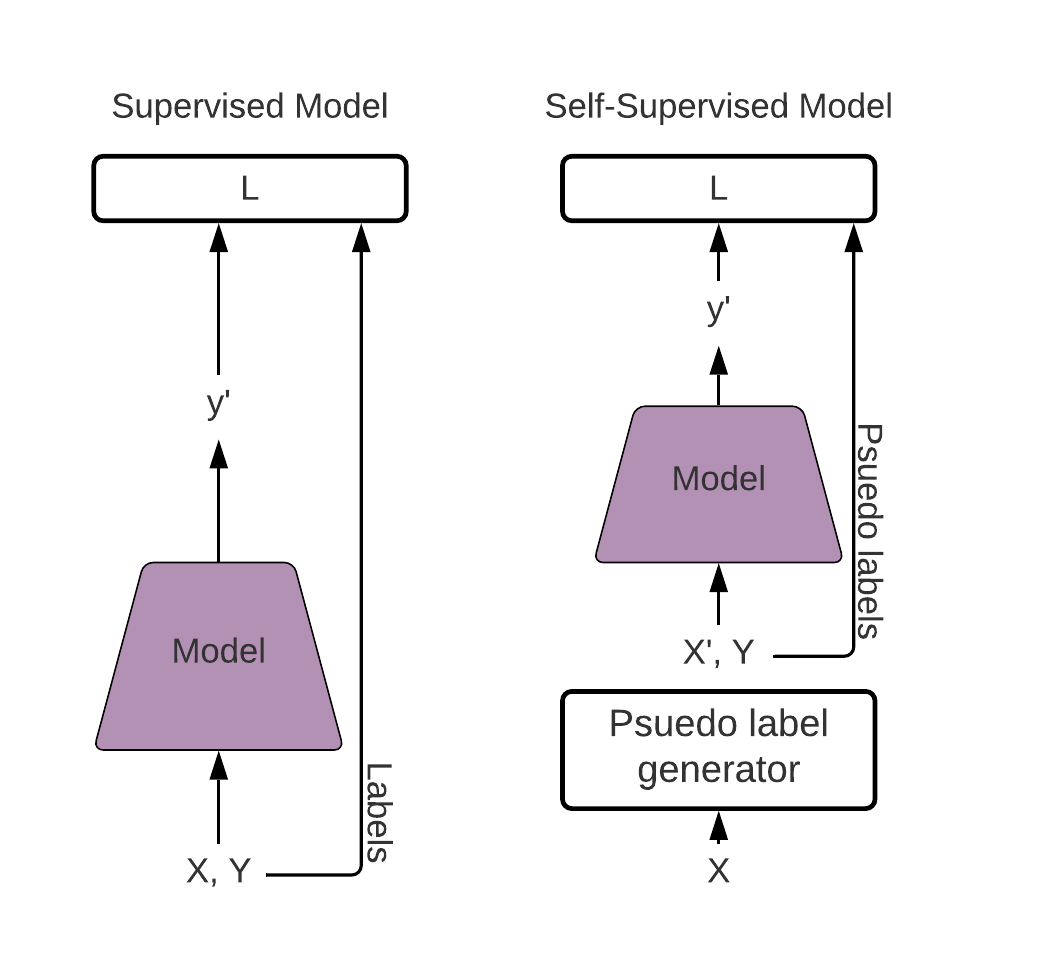}

\captionof{figure}{Supervised vs self-supervised}
\label{fig:sup_unsup}
\end{minipage}
\end{table*}

Figure \ref{fig:sup_unsup} compares the simple workflow in supervised (left) and self-supervised (right) models.
Since data can be in various modalities, such as video, audio or other sensor data, each modality comes with a specialised embedding module, $E_m(\cdot)$, that converts the input sample ${x_{i}^{m}}$ into a vectorised representation, $z_{i}^{m}$, and is followed by a generic model $f(\cdot)$ that calls either the predictor, classifier, encoder or other mapping functions to map the input vector $X \in R^{D}$ onto $y\prime$, i.e. $y^{'}=f(E(x))$. In the supervised model (\ref{fig:sup_unsup}[left]), the model is trained based on input labels $y$ to minimize the loss function $L(f(E(x)), y)$. However, in self-supervised approaches (right figure), the first pseudo-labels to be used as ground truth data $y=g(x)$ are generated based on the task and model $f(.)$. The model is trained to optimise the objective function $L(f(E(x)), g(x))$.
Table \ref{tab:notations} summarises the notation used throughout this paper.
Next, we introduce the keywords and terms used in SSRL approaches.

\begin{itemize}
    \item \textit{Supervised} learning is a subcategory of learning methods that uses ground truth labels (annotated data) to train the learning algorithms/models and tune their parameters with respect to accurately predicting the actual labels. 
    
    \item \textit{Weakly supervised} learning is a type of machine learning technique that works on either incomplete, inexact or inaccurate labels as the input training examples. This approach alleviates the burden of preparing carefully annotated datasets, which can be expensive or infeasible, as was discussed earlier. \textit{Semi-supervised} learning is a special case of weakly supervised approaches in which only a small subset of the training data is annotated.
    
    \item \textit{Self-supervised} learning is a class of machine learning algorithms that works on raw data input without any external ground truth labels. Although there is no external supervision, a self-supervised learning model obtains supervisory signals from the data itself based on extracting the structure, pattern and distribution of data by learning one or more predefined tasks (pretext tasks).
    
    \item \textit{Pretext tasks} are predesigned tasks for the self-supervised models to solve to avoid the need for data annotations. Specifically, the labels required by each task's objective function are automatically generated or extracted from raw training data. These auxiliary tasks are customised for each modality and potential downstream task so that the model can extract the most informative and generalisable embeddings to represent the specific modality. Predicting the next few frames of a time series, learning transformations, such as negating a signal, and rotating an image are examples of pretext tasks.

    \item \textit{Downstream/target tasks} are the main and final tasks for which we train a machine learning model. Employing a pretrained model, the model will be fine-tuned according to the downstream task's objective function. The fine-tuning can be done in either a self-supervised manner without requiring externally annotated data, such as forecasting, change point detection, segmentation and clustering tasks, or they may need labelled data for training, such as classification tasks.

    \item \textit{Modality}: Input data can be presented in various modalities, including text, static images or video, speech, audio, graphs or other sensor signals. Sensor modalities are defined as sensors collecting the same data modality (i.e. time series), but which are still collecting non-redundant data in each mode. For instance, the sensor modes could be measuring different phenomena or even the same phenoma but at different spatial locations. 
    
    \item {Cross-modal Learning}: Cross-modal or cross-view is a multimodal model learning approach where modalities are used as supervisory signal for each other or knowledge from one modality is distilled to another modality. 
    
    \item \textit{View}: Different views of the same sample are separate pieces of data presenting the original sample or event from different perspectives. Views should relate to the same topic but may come from different modalities, such as audio and text data along with visual frames in a video or sensor readings collected by different physiological sensors deployed on the human body. Views can also include similar modalities that are captured from different perspectives, such as videos recorded by different cameras from different angles or inertial signals captured by accelerometer sensors placed on different parts of a human body. 
    
    \item \textit{Augmentation}: Data augmentation involves applying various transformation techniques to the original data and has become an important module in many self-supervised learning methods, such as contrastive models. Using augmentation, we aim to obtain slightly different views of the same data that share similar semantics of the original raw data. 
    We denote the set of augmentation functions used by a model 
    as $\mathcal{A}$.
    
    \item \textit{Pseudo label}: To train self-supervised learning models, the supervisory signal is generated based on the raw data by the model during or before the training state. Depending on the self-supervised task and the model's objective function, pseudo labels can be generated in different ways. To name a few, they can show the temporal (spatial) order of frames (patches) in a video (image), the content of the masked part of the input or the expected outcome of applying a specific transformation function.
    
    \item \textit{Positive pair}: Positive samples $(x, x^{+})$ used in almost all kinds of SSRL methods are derived from the original sample $x$ through the pseudo-label generator function $(x, x^{+}) = g(x)$. Depending on the discrimination strategy, anchor $x$ and its positive pair $x^{+}$ share similar semantics by either 1) using different augmentation techniques so they both refer to the modality $v$ and time step $t$ ($x_{v}^{t}$), 2) referring to synchronous samples from different modalities $(x_{v}^{t}, x_{w}^{t})$ or 3) referring to temporally related samples (e.g. subsequent recordings in sensor data) from either the same or different modalities $(x_{v}^{t},x_{w}^{t'})$, where $v,w \in \mathcal{V}$.
    
    \item \textit{Negative pairs}, denoted as ($x,x^{-}$), are mainly used in contrastive learning approaches and refer to semantically irrelevant samples. 
    
    \item \textit{Sampling} is the process or strategy of selecting positive and negative pairs from all the available samples to create a batch or choosing hard negative pairs within a batch.
    
\end{itemize}



\begin{figure}
    \centering
    \includegraphics[width=\linewidth]{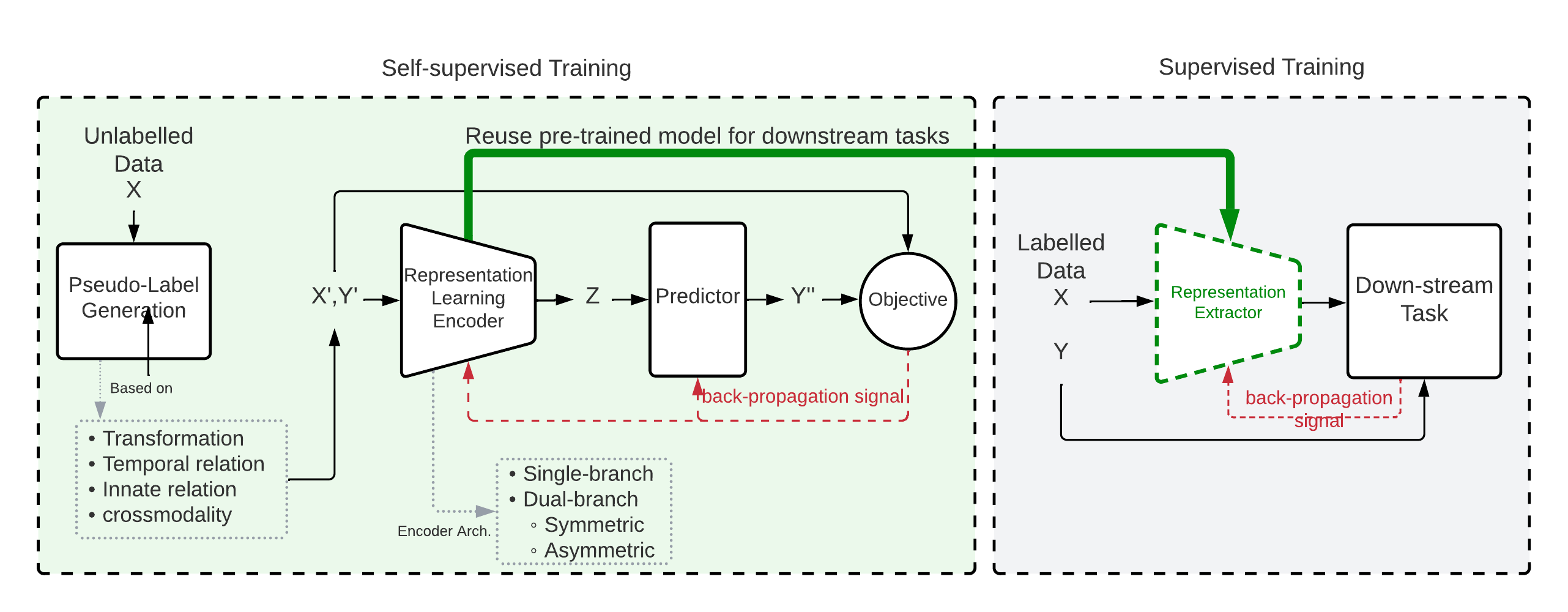}
    \caption{self-supervised representation learning (SSRL) workflow. First, SSRL methods take unlabeled data as inputs, extract new instances and their corresponding pseudo labels using various techniques, such as data transformation, temporal/spatial masking, innate relation and cross-modality matching. Next, representations are learned with the aim of predicting those extracted pseudo labels. Finally, a pre-trained encoder will be transferred to a supervised/unsupervised downstream task with limited labeled data.}
    \label{fig:workflow}
\end{figure}

With the above definitions, the general pipeline of SSRL techniques is illustrated in Figure \ref{fig:workflow}. Almost all self-supervised learning approaches can fit into the workflow depicted. The workflow contains the following major steps:

\begin{itemize}
    \item Self-supervised label generator: This step is mainly about defining pseudo-label generation strategies that are employed to train the representation learning encoder model. The self-supervised model first needs to extract a supervisory signal from the raw input, which will be used to train the model. Depending on the model, pseudo labels can be extracted based on (1) transformation or augmentation techniques, such as signal denoising or inversion, image colourisation or cropping. \cite{wen2021augmentation} explores existing techniques in create augmented samples from time-series data. (2) temporal/spatial relations (e.g. masked input or shuffling) and (3) cross-modal relations. All these approaches preserve similar semantics across the original data and generated samples. 
    
    \item Self-supervised encoder: The underlying network architecture can be constructed from a single branch, including an encoder, followed by a predictor network (for pretext and traditional clustering models) or a dual branch of similar networks that can be trained in either a symmetric or an asymmetric fashion. In symmetric models, both branches of the networks share the weights and are trained together based on the same feedback signal, but in asymmetric models, there are two branches of networks called teacher (online) and student (target) networks that are trained in a self-knowledge distillation manner. The encoder networks, which can be similar or different, are trained or updated through a shared/separate back-propagation signal.
    
    \item Self-supervised objective function: 
    To train the model, an appropriate objective function is needed to guide the model in parameter optimisation, which is usually considered to maximise the agreement between the predicted output ($Y^{\prime}$) and the generated pseudo label ($Y$). In approaches such as pretext or regularisation-based models, the objective functions are based on instance–instance discrimination, while contrastive (and some clustering) models usually discriminate each instance against multiple instances, proposing an instance–context discrimination objective function. In addition, contrastive models also propose a strategy (sampling technique) to find a suitable set of instances (context) to discriminate a given sample against. We will review the evolutionary trend of objective functions in Section \ref{sec:loss_fn}.
    
    \item Downstream task: After training the self-supervised model, the trained encoder module will be reused in solving other downstream tasks. Training for the downstream tasks may require external supervision (annotated data), such as tasks like classification, or not external supervision, such as forecasting, segmentation or change point detection applications. The encoder can be fixed and reused directly for the target problem over unseen data, or it may be fine-tuned according to a target optimisation function and preferences in an end-to-end manner.
\end{itemize}

 The first three factors are designed to work together, depending on the model category and structure of the data. In the following sections, we will review existing categories and specify each factor for existing works.

\subsection{Background and Frameworks}
\label{sec:frameworks}

As we mentioned, SSRL models aim to efficiently learn the hidden structure of the underlying data without requiring annotated datasets. These representations will later be reused in downstream tasks with comparatively less annotated data. Recently, SSRL approaches in different fields (such as CV, NLP, speech recognition and ubiquitous sensor applications) have made great progress and can even surpass their fully supervised rivals in a more data-efficient way.
There are also other frameworks, such as transfer learning, semi-supervised learning and active learning, that alleviate the limitation of labelled datasets. However, SSRL has been shown to be a serious competitor to these frameworks \cite{ericsson2021survey}.

Various classification schemes have been proposed for SSRL, depending on the area of focus (mostly CV-based approaches) or the training objectives. For example, Jing and Tian \cite{jing2020self} categorised self-supervised learning problems into the following four types: 1) generation models\footnote{Please note that the term generation-based model used by the authors of \cite{jing2020self} covers but is not limited to generative networks.} in which all or parts of the samples need to be generated or reconstructed; 2) context-based models that consider the contextual features of data, such as spatial or temporal relationships between different parts of each sample or inter-sample similarity (e.g. clustering methods); 3) semantic-based models that are trained based on automatically generated semantic labels, such as depth of image and moving objects and 4) cross-modality-based models in which the pretext tasks are to verify whether the two channels of input data are relevant/synchronised or not, such as RGB-flow \cite{toering2022self}, visual–audio \cite{arandjelovic2017look} or visual–text\cite{radford2021learning} correspondence correlation. 
However, this classification scheme is specifically designed for studies on CV problems. In addition, approaches that have recently emerged, such as models based on contrastive learning, either cannot properly fit into any of the four categories or span more than one category.
Liu et al. \cite{liu2021genvscon} divided the generation-based models in \cite{jing2020self} into generative\footnote{Please note the term `generative' here is not related to generative network models (e.g. GAN).} and adversarial categories and also considered a third category of contrastive models.

Le-Khak et al. \cite{le2020contrastive} divided SSRL methods into generative, discriminative and contrastive models. Based on this definition, generative models aim to learn good representation by modelling the distribution of input data and regenerating the data based on the inferred distribution, while discriminative approaches are mostly based on learning the mapping between the data and the self-defined pseudo labels. Therefore, the discriminative objective functions are usually formulated as regressive or cross/binary entropy loss functions. They also considered contrastive-based models and generative techniques as two independent categories, according to their objective functions. Unlike a generative model that reconstructs its input samples, in contrastive learning, a representation is learned by comparing the input samples (positive vs. negative pairs). Instead of learning representations from individual data samples one at a time, contrastive learning learns by comparing various samples. The comparison can be performed between positive pairs of `similar' inputs and negative pairs of `dissimilar' inputs. However, this definition does not cover other self-supervised representation learning frameworks, such as clustering, and regularisation-based methods in which there are no explicit negative pairs or automatically generated pseudo labels via pretext tasks.

\begin{figure}
    \centering
    \includegraphics[width=.9\textwidth]{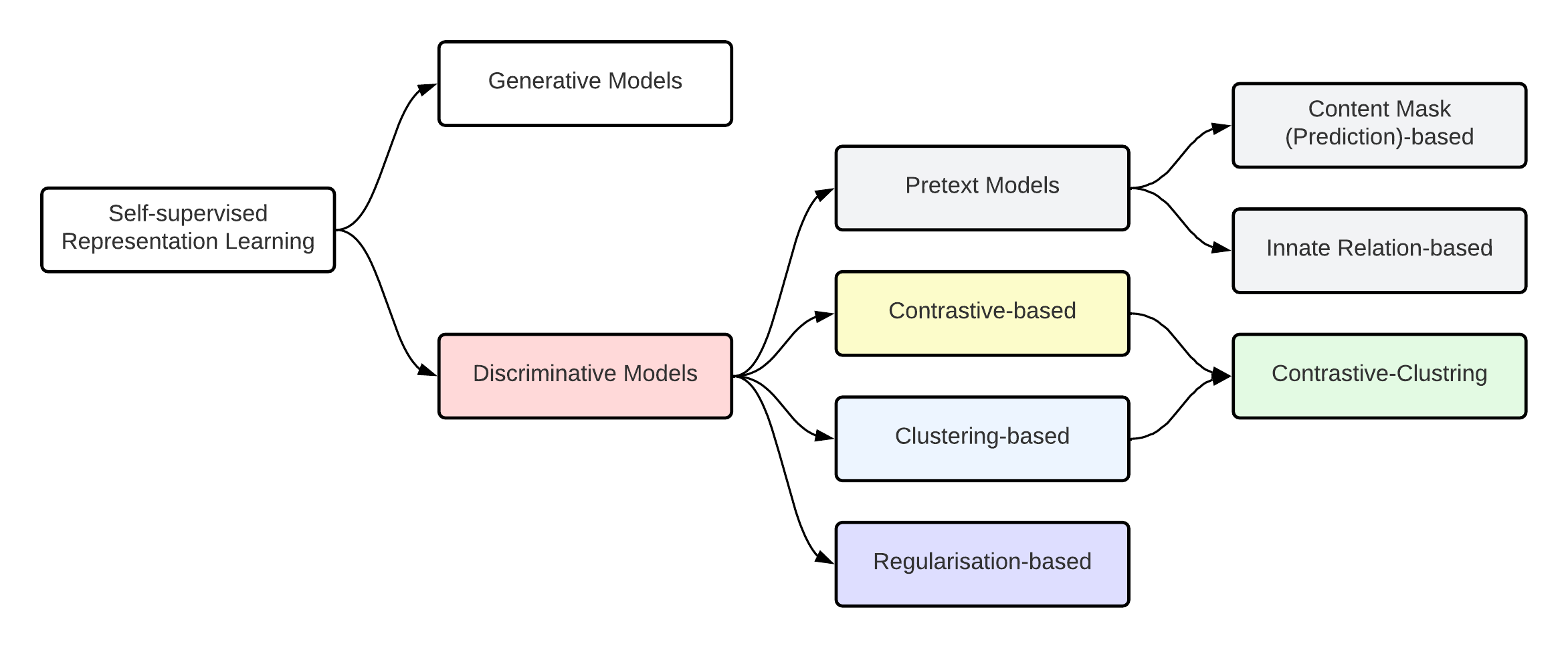}
    \caption{Categories of self-supervised representation learning frameworks applicable across all of underlying architecture and modalities.}
    \label{fig:category}
\end{figure}

In this work, we review existing SSRL methods across various \textit{data modalities} from a network architecture and objective function viewpoint. As shown in Figure \ref{fig:category}, regardless of the type and number of input data modalities, we divide SSRL models into two main groups, namely generative and discriminative models, based on their underlying learning process. Differences between self-supervised generative–adversarial models and discriminator models can be summarised as follows:

\begin{itemize}
    \item Objective function: While generative–adversarial models aim to estimate the distribution of data by leveraging distribution divergence (e.g. using Jensen–Shannon divergence or Wasserstein Distance), the discriminator models are mainly trained using reconstruction or contrastive loss functions.
 
    \item Parameters: GAN models need to train heavy networks to be able to distinguish fake instances from genuine ones, but discriminator models usually come with a simple projection head (e.g. a multi-layer perceptron).
    
    \item Representations: Although many researchers have tried to leverage the representations learnt by the GAN-based model for downstream tasks (BiGAN \cite{donahue2016bigan}, BigBiGAN \cite{donahue2019large}), the other groups of SSRL have outperformed those with fewer parameters \cite{liu2021genvscon}.
\end{itemize}

\begin{figure}
    \centering
    \includegraphics[width=\linewidth]{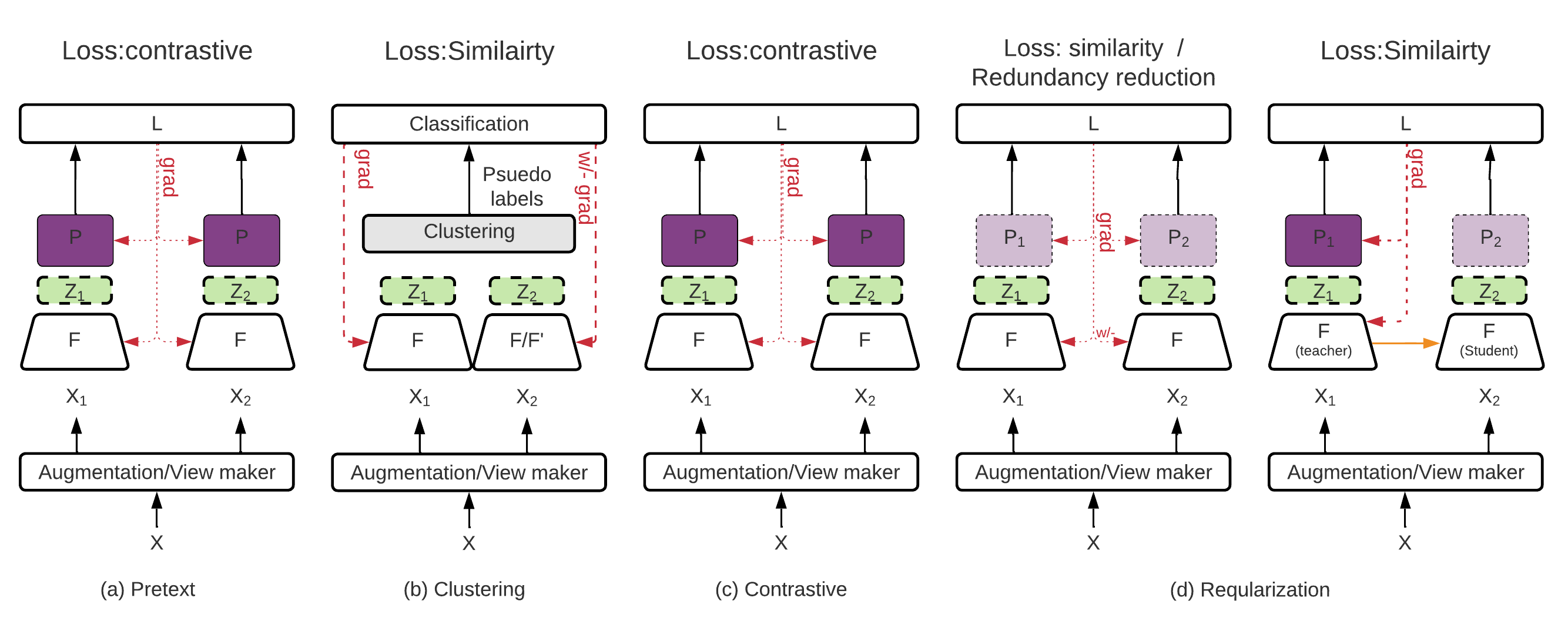}
    \caption{Comparing overall architecture of different self-supervised representation learning models. }
    \label{fig:models}
\end{figure}

In this work, we only cover the latter group of models, i.e. the \textit{discriminative} approaches. Many other works have extensively explored self-supervised generative-adversarial-based models \cite{gao2022generative,liu2021genvscon}. Based on the learning strategy they mainly use to generate the supervisory signal, we divide the discriminative models into four subcategories: 1) pretext-based, 2) contrastive-based, 3) cluster-based and 4) regularisation-based models. Each group can also be divided into more fine-grained subcategories according to their approach to representation learning (e.g. employing a novel objective function and designing a new learning architecture). Figure \ref{fig:models} presents a high-level overview of the architectures of each category.
In the next sections, we cover existing works in each category proposed for single modal data, and then we expand our review to multimodal applications in Section \ref{sec:multimodal}.

\section {Self-Supervised Representation Learning for Temporal Data}
\label{sec:single}
\subsection{Pretext Task}
\label{sec:pretext}
    


The resurgence of self-supervised learning~\cite{de1994learning} based on deep neural networks over the last decade has led to the proposal of various pretext tasks. Pretext tasks allow us to generate pseudo labels from the unlabelled data itself without any human intervention, so that we can train the model with those labels using a standard supervised learning objective function, e.g. cross-entropy loss (see Figure \ref{fig:models}.a). The prediction of Mel-frequency cepstral coefficients features from the raw audio waveforms~\cite{pascual2019learning} and the prediction of signal transformation~\cite{saeed2019multi} are prominent examples of pretext tasks in the temporal domain. These methods are also referred to as auxiliary, surrogate or pseudo tasks in the literature. A well-defined pretext task should be defined in a way that enables the model to learn semantic features effectively from data. 
Nevertheless, it is generally challenging to define a task that can lead to meaningful a priori representations; it is difficult to establish if the surrogate task provides enough training signals to extract features that can be broadly usable by downstream tasks \cite{zaiem2021pretext}. Therefore, many efforts have been made to design effective pretext tasks. In what follows, we describe a range of self-supervised pretext tasks that can be utilised at scale to train deep neural networks with a large amount of unlabelled temporal data, e.g. video, audio, accelerometer and electroencephalogram (EEG).     

We divide the pretext tasks based on their mechanisms for generating a supervisory signal into the following broad categories: \textit{prediction}-based tasks and those exploiting \textit{innate relationships} within the input. The methods centered around \textit{prediction} encompass techniques, such as input reconstruction and denoising. \textit{Innate relationships} include approaches that mainly exploit structure and context to learn features, such as odd sequence detection, correspondence and synchronisation. 

To alleviate the need to collect well-annotated data, the pretext tasks are largely utilised to pretrain deep models. In this realm, ~\cite{fernando2017self} proposed an odd-one-out network, intending to learn video representations by identifying video clips with frames in the wrong order. A multi-branch model is trained by presenting two clips in the correct order, and one clip containing shuffled frames forces the model to extract generalisable features through analogical reasoning. Mishra et al. developed a frame order prediction task ~\cite{misra2016shuffle} to learn video representations through sequential verification. They utilised a triplet Siamese network to predict whether the temporal order of the input tuple of three frames was correct. The temporal class-activation map network proposed by ~\cite{wei2018learning} utilised the arrow of time in videos as supervision. The task comprised detecting whether the video was played forwards or backwards to pretrain the network. 
 
~\cite{gao2018learning} developed an audio source separation method based on deep multi-instance, multi-label learning. They provided a scalable approach to leverage large-scale, in-the-wild videos with multiple audio sources, vision-guided separation and denoising. In addition, Pascual et al. ~\cite{pascual2019learning} developed a multi-task self-supervised learning framework for speech representations. Their problem-agnostic speech encoder was trained with a mix of seven feature prediction (e.g. log power spectrum) and reconstruction (e.g. audio waveform) tasks to obtain a better word error rate on downstream tasks via fine-tuning when compared with the supervised model.

For representation learning in a self-supervised manner from time series and signals, several auxiliary tasks have been proposed in recent years. The signal transformation prediction~\cite{saeed2019multi} method provides a multi-task approach to pretraining temporal models (e.g. convolutional networks) by learning to differentiate between the various transformations of the inputs. The objective is to solve multiple binary classification problems to detect if the sample has been transformed or is a clean example. The sense and learn framework ~\cite{saeed2021sense} proposed a range of pretext tasks to teach models based on a wide variety of inputs, as in some cases utilising transformations without domain knowledge can lead to unintended biases. Specifically, the tasks of blend detection, odd segment recognition, modality denoising and feature prediction from a masked window provide an effective, scalable and inexpensive way to acquire pseudo labels from the signal itself. The proposed techniques were evaluated using a diverse set of downstream tasks ranging from human activity recognition (HAR) to sleep stage scoring. In addition, the task of signal transformation prediction was formulated as a standard multi-class classification problem to simplify the task formulation and learning process from multi-modal data. Likewise, ~\cite{haresamudram2020masked} leveraged the masked reconstruction, similar to BERT~\cite{devlin2018bert} to learn representations from inertial measurement data for HAR applications. 
\begin{table*}[]
  \centering
  \small
  \caption{List of pretext models. Modalities are shown: Video \modvideo|Image \modimage|Audio \modaudio|Text \modtext|Inertial/Bio/Environmental Sensors \modsensors}
    \begin{tabularx}{\linewidth}{|p{1.5cm}|c|p{2.2cm}|p{2.1cm}|p{1.7cm}|X|}
    \toprule
    \textbf{Method} & \textbf{Mod} & \textbf{Label generation strategy} & \textbf{Architecture} & \textbf{Objective Function} & \textbf{Contribution/ Downstream Task} \\
    \hline
    Shuffle \cite{misra2016shuffle} & \modvideo & Temporal order & 3 parallel AlexNet & Cross entropy &  representation learning from videos for action recognition and pose estimation \\
    \hline
    O3N \cite{fernando2017self} & \modvideo & Temporal order & Odd-One-Out Net. & Cross entropy &  action recognition \\
    \hline
    AoT \cite{wei2018learning} & \modvideo & Temporal order & Convolution & Binary log. regression &  action recognition and video forensics\\
    \hline
    BERT \cite{devlin2018bert} & \modtext & Masked input / prediction & Bidirectional Transformers & Cross entropy      &   Bidirectional representation learning and can be applied to NLP tasks such as those defined in GLUE, MultiNLI, and SQuAD \\
    \hline
    PASE \cite{pascual2019learning} & \modaudio &  Feature prediction and reconstruction      & SincNet  &  Binary and regression losses & Speaker ID, emotion, and ASR \\
    \hline
    TPN \cite{saeed2019multi} & \modsensors & Signal transformation &  Temporal Convolutional  & Multiple BCE & Multiple pretext task learning / HAR\\
    \hline
    MRS \cite{haresamudram2020masked} & \modsensors & Masked input &   Transformer   &  Reconstruction    & HAR \\
    \hline
    Sense\&Learn \cite{saeed2021sense} & \modsensors & Signal transformation, masked input, shuffle, and others  & Temporal Convolutional & Cross entropy, reconstruction +metric losses & Multiple tasks in sensory domain \\
    \hline
    AALBERT \cite{chi2021audio} & \modaudio & Masked input & Transformers &   Reconstruction &  Phoneme and utterance-level speaker classification \\
    \hline
    PDR \cite{CarrBBTZ21} & \modaudio & Random permutation in spectrogram & Parallel AlexNet+ shared weight concatenated FCN & FenchelYoung loss \cite{blondel2020learning} & Employed a learning to rank loss function to  learn the order of shuffled parts of audio. \\
    \hline
    TERA \cite{liu2021tera} & \modaudio & Temporal, frequency, magnitude & Transformers & Reconstruction & Keyword spotting, speaker ID, and ASR. \\
    \hline
    PTS \cite{zaiem2021pretext} & \modaudio &  Prosody and spectral descriptors     &  Hybrid convolutional  STM     &   Multi-task losses    & Pretext task selection / ASR, speaker ID, emotion recognition \\
    \hline
    TST \cite{zerveas2020transformer} & \modsensors& Masked input & Transformers & MSE & Multivariate time series regression and classification \\
    \bottomrule
    \end{tabularx}%
  \label{tab:pretext_tab}%
\end{table*}%


\subsection{Contrastive Models}
\label{sec:contrastive_model}

Contrastive learning has become a prominent tool for unsupervised representation learning from various data modalities. To recap, it is utilised to pretrain a deep neural network using a contrastive or a metric learning objective on a large-scale unlabelled dataset. The network learns general-purpose representations by extracting similarities between embeddings of an input instance and its transformed (or augmented) variant and identifying dissimilarities with other examples. Figure \ref{fig:models}.c depicts the overall design of contrastive-based models. The deep metric learning loss \cite{chopra2005learning} is the earliest attempt to use discriminative learning with a focus on face verification. For similar applications, several loss functions have been proposed to address shortcomings of vanilla metric loss, such as triplet loss \cite{schroff2015facenet} and N-pair loss~\cite{sohn2016improved}. The noise–contrastive estimation framework was proposed by~\cite{gut2010NCE} to estimate the parameters of a statistical model using nonlinear logistic regression to differentiate observed data from artificially generated noise. Van den Oord et al.~\cite{oord2018representation} expanded the approach further and proposed the InfoNCE objective function to develop contrastive predictive coding, a universal unsupervised learning method. This method utilises a standard cross-entropy objective function to discriminate between positive samples and noisy samples in the latent space. The evaluation across a range of domains (including vision, NLP and reinforcement learning [RL]) highlighted the efficacy of the approach in learning useful representations for downstream tasks.~\cite{wang2021contrastive} extended this method and proposed contrastive separative coding for speaker verification (using audio) by separating the target signal from contrastive interfering signals.

For visual representation learning, Deep InfoMax~\cite{hjelm2018learning} exploits locality in the inputs into the objective function to simultaneously estimate and maximise the mutual information between the input and its high-level representations. SimCLR ~\cite{chen2020simple} further simplified contrastive learning without requiring a memory bank by carefully selecting data augmentation for positive generation and scaling the batch size to increase the negative samples. ~\cite{he2020momentum} presented momentum contrast (MoCo), which trains an encoder by matching a query (embedding) with keys from a momentum encoder. The momentum encoder is a dictionary comprising encoded representations of current and former mini-batches, decoupled from the batch size. Similarly, Saeed et al. ~\cite{saeed2021contrastive} proposed contrastive learning for audio without explicit data augmentation; it has been shown that an encoder can be pretrained by leveraging random cropping from the same clip for anchor and positive generation. The prototypical contrastive learning approach~\cite{li2021prototypical} combined instance discrimination with clustering to learn low-level and semantic features using a ProtoNCE objective. It encourages representations to be closer to the prototypes modelled as latent variables. Similarly, Divide and Contrast~\cite{tian2021divide} also leveraged clustering with any self-supervised learning approach to cluster representations for training independent expert models that were distilled to create a single model by predicting representations. 

DeCLUTR~\cite{giorgi2020declutr} presented an approach for learning sentence-level embeddings from a large amount of unlabelled text corpora using multiple techniques, such as overlapping, for efficiently conducting anchor-positive sample generation. In the time-series realm, ~\cite{saeed2021sense} used symmetric triple-based objectives for learning representations from various sensor input modalities. ~\cite{haresamudram2021contrastive} applied contrastive predictive coding \cite{oord2018representation} for HAR. Similarly,~\cite{deldari2021tscp2} extended contrastive predictive coding to detect change points in multidimensional time-series data.~\cite{eldele2021tstcc} proposed a similar approach to SimCLR for time-series representation learning using a transformer by maximising context similarity between strongly and weakly augmented inputs. In addition, temporal neighbourhood coding~\cite{tonekaboni2021unsupervised} was proposed to train an encoder by predicting whether segments were from the same neighbourhood, where the neighbourhood distribution was modeled as a Gaussian distribution to capture the gradual transitions in temporal data. Contrastive learning was also leveraged in a federated learning setting to exploit unlabelled decentralised data~\cite{saeed2020fed} and was shown to learn useful representations from a variety of sensory data, e.g. EEG and IMUs. 

\begin{table*}[htbp]
  \centering
  \caption{List of contrastive-based models. Various modalities are shown as: Video \modvideo|Image \modimage|Audio \modaudio|Text \modtext|Inertial/Bio/Environmental Sensors \modsensors}
  \small
    \begin{tabularx}{\linewidth}{|p{1.8cm}|c|p{2cm}|p{2cm}|p{1.2cm}|X|}
    \toprule
    \textbf{Method} & \textbf{Mod} & \textbf{Label Generaton Strategy} & \textbf{Architecture} & \textbf{Objective Function} & \textbf{Contribution/ Application} \\
    \hline
    CPC \cite{oord2018representation} & \modimage\modtext\modaudio &      Temporal locality &   Multiple    &  InfoNCE     & Modality-agnostic, considering temporal context  \\
    \hline
    InfoMax \cite{hjelm2018learning} & \modimage &   -    &  Convolutional     &    DeepInfoMax   &  Incorporating locality information and maximize the MI between input and extracted features\\
    \hline
    SimCLR \cite{chen2020simple} & \modimage & Augmentation & Dual Symmetric Encoder & NT-Xent & Simple yet powerful image representations \\
    \hline
    MOCO \cite{he2020momentum} & \modimage & Augmentation & Momentum Encoder & InfoNCE & Self-distillation combined with memory-bank for negative selection \\
    \hline
    DeCLUTR \cite{giorgi2020declutr} & \modtext  & Spatial locality      & Pretrained Transformer-based model       & InfoNCE      & Proposed a span sampling strategy to sample multiple positive samples along with easy/hard negatives. \\
    \hline
    Wav2Vec \cite{baevski2020wav2vec} & \modaudio & Temporal locality & Convolution & InfoNCE & Replaced the RNN in CPC with convolution for context head.  \\
    \hline
    VQ-Wav2Vec \cite{BaevskiSA20} & \modaudio & Temporal locality & Convolution & InfoNCE & Combining Wav2Vec and BERT \\
    \hline
    COLA \cite{saeed2021contrastive} & \modaudio & Implicit augmentation via random cropping &      EfficientNet-B0 &  NT-Xent &  General-purpose audio representations \\
    \hline
    CLOCS \cite{kiyasseh2020clocs} & \modsensors& Temp. locality + augmentation &  Temporal CNN     & InfoNCE &  Electrocardiogram representation learning \\
    \hline
    SCN \cite{saeed2020fed} & \modsensors& Scalogram &  Temporal CNN     &  Contrastive & Leveraged CL in federated learning \\
    \hline
    SACL \cite{cheng2020subject} & \modsensors & Augmentation       &    1D ResNet   & InfoNCE & Biosignals (EEG, ECG) \\
    \hline
    IIC \cite{tao2020self} & \modvideo & RGB + opt. flow & R3D (ResNet)  &    InfoNCE & Inter-intra contrastive approach \\
    \hline
    CoCLR \cite{han2020coclr} & \modvideo & Augmentation and  cross-modal  & S3D \cite{xie2017rethinking}     & InfoNCE & Multiple positives| Two stages of training: 1) Initialization (individual modality training), 2) Alternation (positive terms are chosen from the other modality).   \\
    \hline
    TSCP2 \cite{deldari2021tscp2} & \modsensors& Temporal locality & Symmetric Conv-based networks & InfoNCE & Used predictive coding for change point detection \\
    \hline
    CPC (HAR)\cite{haresamudram2021contrastive} & \modsensors& Temporal locality & Symmetric branches with Conv+GRU & InfoNCE & Employed CPC for HAR application \\
    \hline
    CPC (EEG)\cite{banville2021uncovering} & \modsensors& Temporal locality &       & InfoNCE & Employed CPC for Sleep Stage detection \\
    \hline
    TSTCC \cite{eldele2021tstcc} & \modsensors& Temporal locality + augmentation & SimCLR + Transformer & InfoNCE & Strong and weak augmentaions \\
    \hline
    TCN\cite{tonekaboni2021unsupervised} & \modsensors& Temporal locality &  ConvLSTM     & Tripletloss & CPC with novel sampling method \\
    \hline
    SleepDPC \cite{xiao2021sleep} & \modsensors &   Temporal locality     &   ConvLSTM    &  InfoNCE     & Sleep stage detection \\
    \hline
    CSC \cite{wang2021contrastive} & \modaudio &  Input mixing     &  SincNet  &  Multiple losses     & Distinguishing the target audio from inferring signals \\
    \hline
    CVRL \cite{qian2021spatiotemporal} & \modvideo & Temporal locality + augmentation &   3D-ResNet-50    &  InfoNCE & Employed both temporal and spatial augmentations \\
    \hline
    CC \cite{li2021contrastive} & \modimage & Augmentation &  ResNet-34    & InfoNCE & Instance- and cluster-level contrasting \\
    \bottomrule
    \end{tabularx}%
  \label{tab:contrastive_tab}%
\end{table*}%

In RL,~\cite{nguyen2021temporal} developed temporal predictive coding to extract features from an environment that were temporally consistent and could be predicted across time. The learned model was leveraged for planning in latent space on DMControl tasks. The backgrounds were modified with natural videos for injecting task-irrelevant information to evaluate the robustness of the learned agent. 
Time-contrastive networks~\cite{Sermanet2017TCN} employed third-person video observations for learning representations of robotic behaviours with metric loss. They demonstrated that the trained encoder could enable robots to directly mimic human poses and included a reward function in the RL algorithm. For video representation learning with three-dimensional (3D) convolutional networks, \cite{tao2020self} proposed an inter–intra contrastive approach. The different views of the same video clip (such as RGB frames and optical flow) were treated as positive, while those from other videos were used as negative samples for the training model with contrastive loss. A similar approach was suggested by~\cite{qian2021spatiotemporal} for contrastive video representation learning. It works by sampling temporal intervals that represent a number of frames between the starting points of two clips selected from a video. The clips are transformed using a temporally consistent spatial augmentation, and a contrastive objective is used to train the model by repelling embeddings of the clips from different videos and attracting clips from the same video. Along similar lines of exploiting multiple views,~\cite{han2020coclr} proposed a self-supervised co-training scheme to improve video representation learning with contrastive loss.

\begin{table*}
 \centering
\label{tab:clustering_tab}
\caption{List of clustering models. Various modalities are shown as Image \modimage}
\small
\begin{tabularx}{\linewidth}{|c|p{1.7cm}|c|p{1.6cm}|p{1.6cm}|p{2.35cm}|X|}
\toprule
\rotatebox[origin=c]{90}{Category}				& Method                         & \rotatebox[origin=c]{90}{Modality}            & Label generation strategy & Architecture            & Objective Function & Contribution / Application   \\ \hline
    \multirow{3}[6]{*}{\rotatebox[origin=c]{90}{Clustering}} & DeepCluster~\cite{caron2018deep} & \modimage & clustering & \multicolumn{1}{l|}{AlexNet, VGG-16} & \multirow{2}{*}{\shortstack{classification loss (CE)\\ + clustering loss}} & \multirow{2}{*}{\shortstack{Alternate between clustering and \\ cluster assignment}} \\
\cline{2-5}          & SeLA \cite{YM2020Selflabelling} & \modimage & clustering & \multicolumn{1}{l|}{AlexNet, ResNet-50} & \multicolumn{1}{l|}{} &  \\
\cline{2-7}          & ODC \cite{zhan2020ODC} & \modimage & NA    & Single branch ResNet-50 &  "loss re-weighting" technique  & No need to alternate between clustering and cluster refinement by introducing memory bank of centroids \\
    \hline
    \multirow{5}[6]{*}{\rotatebox[origin=c]{90}{Cont. + Clustering}} & SwAV \cite{caron2020swav} & \modimage & Augmentation / Clustering & ResNet-50 & Swapped prediction problem & contrasting cluster assignments \\
\cline{2-7}          & DnC \cite{tian2021divide} & \modimage & Augmentation & SimCLR+BYOL & InfoNCE + MSE & Representations clustering followed by cluslter-specific CL \\
\cline{2-7}          & PCL \cite{li2021prototypical} & \modimage & Augmentation / Clustering & ResNet-50 & ProtoNCE (general form of InfoNCE) & Alternate between clustering and contrastive learning \\
    \bottomrule
    \end{tabularx}%
  \label{tab:addlabel}%
\end{table*}%

\subsection{Clustering}
\label{sec:clustering_model}
As a typical class of unsupervised learning approaches, clustering has been widely studied in machine learning and deep learning domains.
Integrating clustering with representation learning has become another popular method of self-supervised learning.
The core idea is to leverage clustering techniques to provide pseudo labels as training signals (see Figure~\ref{fig:models}.b).
This line of work mainly focus on learning effective visual features for various downstream tasks, including image classification, object detection and semantic segmentation.
In order to drive similar samples into the same class and learn visual representations, DeepCluster~\cite{caron2018deep} adopts the k-means clustering assignments as pseudo labels to iteratively label each image.
Simply using k-means assignments during training could result in a degenerate solution in which all samples tend to be mapped within the same cluster. Targeting this issue, Asano et. al.~\cite{YM2020Selflabelling} considered the pseudo-label assignment problem as the optimal transport problem and further proposed using a fast version of the Sinkhorn–Knopp algorithm~\cite{cuturi2013sinkhorn} to efficiently find the approximate solution to the transport problem.
Similarly, Caron et. al. ~\cite{caron2020swav} proposed SwAV, which extends this clustering setting using a contrastive learning setting. Under the scalable online assignments, this could simultaneously cluster the data and enforce consistency between different views.
In order to further improve stability in the training of the clustering-based self-supervised learning paradigm, ~\cite{zhan2020ODC} developed ODC, and online deep clustering. Instead of alternating between feature clustering and the updating of network parameters, ODC decomposes the clustering process into mini batch-wise label updates that are integrated into iterations of network updates (e.g. convolution networks for extracting features). As a result, the feature clustering process and the updating of network parameters could be performed simultaneously.



\subsection{Regularisation-Based Models}
\label{sec:regularization_model}
Like contrastive models, these models work on distorted views of the same sample with the aim of maximising the similarity between representations of the augmented views. However, unlike contrastive models, they do not require  negative pairs to be included in the cost function. These models are comprised of two branches of encoders with similar architecture. These two branches, usually known as the student and teacher networks (or online and target networks) are trained to learn similar latent representations (see Figure \ref{fig:models}.d). 

As mentioned in the previous section, sampling plays a critical role in determining the quality of the learned representations. In contrastive models, negative sampling helps to learn the distinguishing features of data and prevent representation collapse where all embedding vectors collapse to a single trivial solution. However, using an effective sampling technique can be computationally expensive. As a result, \textit{regularisation}-based techniques have been proposed. They have gained much attention as they remove the need to use negative samples, whilst still avoiding representation collapse \cite{hua2021feature}. Regularisation approaches also remove the problem of learning representations with false negative pairs.

First, Bootstrap Your Own Latent (BYOL) \cite{niizumi2021byol} proposed a self-distillation model that employed asymmetrical encoding networks as online and target networks. The online network was trained based on backpropagation of the objective function, while the target network was updated based on a slowly moving exponential average of the online network. Consequently, the target network used what was referred to as a momentum encoder. Although, as we discussed earlier, negative examples in a contrastive loss function prevent representation collapse, BYOL does not consider negative examples, and the objective function is entirely based on the similarity of representations learnt from the online and target networks. \cite{fetterman2020understanding,tian2020understanding} conducted experiments to show that batch normalisation implicitly plays the role of negative pairs by forcing the representations of each positive pair to be far from the other samples in the batch. However, later authors of BYOL claimed that their approach was still effective even without batch normalisation \cite{richemond2020byol} by replacing it with group normalisation and weight standardisation in the network layers. 

The Simple Siamese Network (SimSiam) \cite{chen2021exploring} simplified the BYOL architecture by removing the need for a momentum encoder. \cite{chen2021exploring} showed that a stop gradient on the student network and a predictor on the top of the teacher network were crucial to avoid representation collapse. Therefore, both network branches update together based on the gradient of the predictor's output. DINO \cite{caron2021dino} expanded the momentum encoder from BYOL to avoid representation collapse by `centering' and `sharpening' the outputs of the teacher network. It was claimed that centering prevents one dimension from dominating the representation but encourages collapse to a uniform distribution, whilst sharpening has the opposite effect on the extracted representations. Therefore, together they can successfully replace batch normalisation.

\begin{table*}[htbp]
  \centering
  \caption{List of regularisation-based models. Various modalities are shown as Image \modimage}
  \small
    \begin{tabularx}{\linewidth}{|p{1.5cm}|c|p{1.5cm}|p{1.9cm}|X|X|}
    \toprule
    \textbf{Method} & \textbf{Mod} & \textbf{Label Generation Strategy} & \textbf{Architecture} & \textbf{Objective Function} & \textbf{Contribution/ Application} \\
    \hline
    
    BYOL \cite{niizumi2021byol} & \modimage & Augmentation & Sym. encoders + Momentum &   Mean Squared Error    & Removed the need for memory bank in MOCO  \\
    \hline
    SimSiam \cite{chen2021exploring} & \modimage & Augmentation & Dual Sym. encoders + one branch predictor&  Mean Squared Error  & Removed the need for Momentum update of BYOL \\
    \hline
    DINO \cite{caron2021dino} & \modimage & Augmentation & Self-Distil. + transformer & Cross entropy between distribution of representations from teacher and student networks     & Momentum encoders combined with ``Centring'' and ``sharpening'' of Teacher network's output \\
    \hline
    \multirow{2}{*}{\shortstack{Barlow-Twins\\ \cite{zbontar2021barlow}}} & \modimage & Augmentation & Dual Symmetric Encoder & Maximize agreement between the correlation and Identity matrix & A novel loss to maximize the information content and reduce redundancy of embeddings \\
    \hline
    W-MSE \cite{ermolov2021wmse} & \modimage & Augmentation & symmetric encoders  & Whitening MSE & Proposed Whitening MSE to remove the need for negative pairs \\
    \hline
    VicReg \cite{bardes2022vicreg} & \modimage & Augmentation & Asymmetric encoders      &       Regulaizing Variance, Invariance, Covariance of embeddings & Not require symmetric architecture suitable for multimodality\\
    
    \bottomrule
    \end{tabularx}%
  \label{tab:reg_tab}%
\end{table*}%

Another category of methods \cite{zbontar2021barlow,ermolov2021wmse,bardes2022vicreg} have introduced new objective functions that have been shown to achieve state of the art SSRL performance, whilst removing the need for negative samples, a momentum encoder and a `stop gradient'. Barlow Twins \cite{zbontar2021barlow} replaced the simple cosine similarity-based loss function with a new cost function that learns an invariant representation that minimises the redundancy of samples. The cost function achieves this by ensuring the cross correlation matrix of positive embedding pairs are brought as close as possible to the identity matrix. Similarly, W-MSE \cite{ermolov2021wmse} tried to force the spread of representations in a batch over the unit sphere. Very recently, Variance–Invariance–Covariance Regularisation (VICReg) \cite{bardes2022vicreg} was proposed based on the Barlow Twins approach and improved the model in various ways. VICReg proposed a novel objective function that considers three criteria, including the distance between representations, and bound the standard deviation of each representation variable along and the covariance of each pair of representations (this one has been borrowed from Barlow Twins \cite{zbontar2021barlow}). In addition, VICReg does not require that the architecture of the two branches be similar, and therefore, they will be trained separately based on the gradient. This recent line of designs is depicted in a high-level illustration in Figure \ref{fig:models}(d).
We will discuss the evolutionary trend of objective functions in SSRL models in Section \ref{sec:loss_fn}.
Almost all of the aforementioned approaches have been proposed for vision applications, mostly images. Therefore, it is an interesting direction for future studies to apply and customise these approaches for other modalities.


\section{Self-Supervised Representation Learning in Multimodal Data}
\label{sec:multimodal}
In this section, we explore existing approaches in multimodal self-supervised learning from different viewpoints. First, we explain the motivation and importance of multimodal self-supervised learning.
Second, we introduce the available categories of modality fusion techniques and the challenges of each model. Finally, we compare existing methods based on the number and type of modalities they can handle. 

\subsection{Why Self-Supervised Multimodal Learning?}
The human brain is constantly learning from the data it receives from our various sensory inputs (i.e. vision, hearing, smell, touch and taste). Inference and learning are the products of a series of actions, including correlating and contradicting the data received from various sources, and can be combined with our previous experience and knowledge (i.e. memory feedback). 
The manner in which a baby learns is a good example of our learning process. Experiences during our lives are mostly multimodal. Each mode of received data provides a complementary view of the same incident that can help us precisely perceive our environment. For example, if we consider an apple, we can recognise it by its smell, shape and taste. Learning a concept via multiple modalities is rooted in the concept of `degeneracy' in Neurobiology. Tononi et. al. \cite{tononi1999degeneracy} explained that `\textit{degeneracy is the ability of elements that are structurally different to perform the same function}'. Therefore, each modality explains the very same event in a different language and will reveal a different perspective with extra details about the very same experience. These overlapping and complementary views from various modalities help us to understand our environment, even in situations in which we lack one or more modalities. 
According to \cite{smith2005sixlesson}, babies learn from their multimodal sensory system, and surprisingly, they can acquire knowledge without needing any external supervision. The more they are exposed to different environments and experiences, the faster they learn and make precise connections between different modalities. Babies achieve this self-supervised multimodal synchronous perception \textit{incrementally} through their \textit{interactions} (i.e. series of \textit{actions} and \textit{observations}) with the world around them \cite{smith2005sixlesson}. 

Similarly, in the regime of artificial intelligence, fusing data from various modalities is a crucial factor in building successful applications. 
Multimodal learning is a well-established concept in many important tasks, such as visual question answering \cite{Kim_Jeong2021vqa}, cross-lingual speech translation \cite{khurana2020cstnet} and audio–visual retrieval \cite{morgado2021audio}.
We can summarise the reasons behind the recently growing interest in multi/cross-modal self-supervised learning as follows:
\begin{itemize}
    \item Annotated dataset: The success of traditional machine learning models depends heavily on the \textit{availability} and \textit{accuracy} of an annotated dataset. However, annotation of multimodal data is challenging due to (1) different levels of granularity of various modalities; (2) limitations in the number of modalities, size and generality of publicly available annotated datasets; (3) the requirement for domain experts in order to acquire annotated datasets in some domains, such as medicine, law and biology and (4) privacy issues in annotating datasets in some cases, especially in medical and other human-related applications. These challenges can result in a shortage of high-quality annotated datasets, which results in the poor performance of supervised models. However, self-supervised multimodal models can leverage an abundant amount of available unlabelled data.
    
    \item Supervisory effect: As we discussed earlier, modalities can provide supervisory signals for other modalities, such as the most well-known applications in the domain of text–speech \cite{baevski2019effectiveness} or text–vision \cite{radford2021learning} and recently in ubiquitous applications under the multi-device topic \cite{jain2022collossl}.
    
    
    \item Emerging applications: Due to the ever increasing amount of data being collected, stored and served, there is an increasing tendency towards using multimodal data within analytical applications. Modern intelligent systems are required to fuse heterogeneous data sources in order to makes predictions and inferences. With the emergence of the Internet of Things (IoT) where data is continuously being collected from a wide range of devices and individuals, there is an undeniable need for multimodal and \textit{cross-modal} learning frameworks \cite{jain2022collossl}.
    
    \item Diversity in data: Learning from a broad range of modalities not only yields a general-purpose solution but also provides the opportunity to uncover hidden and less-studied patterns in real-world data that may not be present in small annotated datasets, which are usually captured under fully controlled and restricted conditions.

\end{itemize}

\subsection{Cross-Modal Learning Categories}
Traditionally, cross-modal learning in machine learning has focused upon modeling different modalities independently prior to their fusion. For example, work integrating audio and video data have used visual networks to train the audio model (such as SoundNet \cite{aytar2016soundnet}) or vice versa \cite{owens2016ambient}. Previously, it was thought that our brains processes different sensing modalities independently (in separate unisensory regions) and the understanding obtained from each modality is then integrated after this individual processing. However, it has been shown that multisensory convergence occurs much earlier than previously thought \cite{schroeder2005multisensory}. This discovery has led to the exploration of multimodal machine learning techniques. Recent work has suggested that combining different data modalities in the early stages of modeling results in more accurate and robust deep neural networks \cite{barnum2020benefits}. 

Traditional supervised multimodal deep learning models trained modality-specific or modality-generic networks separately to extract modality-specific features, and then, to consider the relationships between modalities, they would use various techniques to combine the extracted modality-specific features. Based on how and when models combine and integrate different modalities, multimodal learning can be divided into (1) immediate fusion, (2) intermediate fusion and (3) late fusion. In immediate fusion, the extracted features are concatenated and fed into a shared model; however, in intermediate fusion and late fusion, the features are fed into different or shared processing models independently and are only combined in the middle or late layers of the model. 

None of these traditional models considers fusion of the input modalities before extracting representations (features)~\cite{barnum2020benefits}. We specifically focus on this type of fusion, which occurs before representations are extracted. This approach is also called \textit{cross-modal} learning in which we account for the possible relationship between modalities from the very beginning.
Hence, each modality can be considered as a supervisory signal for the other modalities. For example, text data can perfectly supervise the classification of captioned images \cite{radford2021learning} or audio with transcripts \cite{guzhov2021audioclip}. Another example of cross-modal self-supervised learning is instructional videos in which the sequence of the images is highly correlated with the audio (spoken words) \cite{miech2019howto100m,amrani2019learning,sun2019videobert,yang2021just}. Similarity ubiquitous computing applications with various sensor modalities can benefit significantly from self-supervised cross-modal learning \cite{jain2022collossl}. 
For example, in the field of robotics, integrating data from many different domains, such as visual, audio, speech and inertial data, is a popular research topic. Much work has been done in integrating visual and motion sensor data (visual–inertial odometry) \cite{chen2019selective,ALMALIOGLU2022119,kim2021unsupervised} to enable different motions and poses of mobile robots.

Amrani et. al. \cite{Amrani2021noise} provided a comprehensive taxonomy of \textit{cross-modal} representation learning models based on how they infer modality representations and capture cross-modal information. However, in this work, we only explore existing categories proposed for SSRL models as follows: 
\begin{enumerate}
    \item Joint representation learning models map all modalities into embedding vectors in a shared space. These models learn modality-specific representation under some constraints shared between all modalities. The most applied constraint in multimodal SSRL models is to maximise the agreement between various modalities' representation by maximising their correlation, similarity or mutual information \cite{tian2020contrastive,jain2022collossl,guzhov2021audioclip,yuan2021multimodal}.
    
    \item Translator representations are mostly based on encoder–decoder frameworks and are learned from a single modality that can translate the input sample into the corresponding sample or representations from another modality. As an example, VirTex \cite{desai2021virtex} proposed an encoder–decoder model to train visual representations by solving a proxy task based on the corresponding text from image captions. However, these methods do not exploit the full potential of each modality, as they do not consider the correlations between all modalities simultaneously.
    
\end{enumerate}

The former model has been used in the most recent frameworks (see Table \ref{tab:bimodality_comparison} and \ref{tab:modality_comparison}), especially in contrastive learning approaches in which they contrast different modalities instead of the augmented views of a single modality.

A good model for multimodal representation learning should be able to address the following criteria:
\begin{enumerate}
    
    \item Comprehensive representation: It is important to capture a compact but informative representation that is shared across all modalities. Usually, all modalities share the same high-level semantic but not necessarily the same phenomenon. For example, considering an image and its caption, they can be relevant, as they are pointing to the same topic, but their content may not be strictly the same. Some details in the image may not be covered in the text and vice versa \cite{shi2021relating,yang2017deep}. In contrast, in multimodal temporal data, the data usually comes from different sensors, such as cameras, motion sensors and wearable sensors, and all modalities represent the same phenomenon from different perspectives. Hence, a model needs to learn representations that reflect not only the features of each individual modality but also the global features at a high level. 
    
    \item Robustness against noise: Since there is no label to select negative pairs, the method should be robust against false negatives that may exist within a mini-batch. In most methods, different pairs within batches are treated as negative pairs while some of them may be semantically related. This problem causes more issues in cases with a limited range of classes, such as medical applications, human activity and emotion recognition \cite{deldari2021tscp2}. 
    
    \item Generic solution: The pretrained model should be extendable to various tasks, applications and modalities.
    \item Temporal structure: In the case of temporal modalities, the model should be able to capture the temporal structure of the data and the progression of the underlying temporal dynamics \cite{tonekaboni2021unsupervised}.
    
    \item Modality selection: In the case of multimodal temporal models, the model needs to be able to dynamically weigh each modality and emphasise the more important ones because irrelevant modalities (modalities that do not participate in the current class) can lead to a degenerate solution when the objective function maps the modalities more closely in the latent space.
\end{enumerate}

Next, we will first cover cross-modal approaches that work on only two modalities, then we explore the most recent works that have expanded the number of modalities to three or more. There is a large body of research on multimodal techniques; however, they are mostly dependent on annotated data. In this work, due to the main focus of the paper, we will highlight the self-supervised models proposed for temporal \textit{cross-modal} data.

\subsection{Dual-Modality Self-Supervised Recognition Learning}
Audio–visual correspondence (AVC) techniques \cite{arandjelovic2018objects,arandjelovic2017look} aim to learn shared representation between an image (a single frame of a video) and its corresponding sound segment using the contrastive learning framework. 
The AVC has been leveraged to learn representations for both modalities. ~\cite{arandjelovic2018objects} proposed AVE-Net to extract aligned visual and audio embeddings for inputs' correspondence with Euclidean distance between the features. 

Similarly, ~\cite{owens2018multisensory} presented an approach to predict whether video frames and audio are temporally aligned, where misaligned samples were generated simply by a shift operation in the audio domain. SoundNet ~\cite{aytar2016soundnet} capitalised on large-scale unlabelled data using a pretrained vision model to learn audio representations with knowledge distillation. Apart from pretext models,
Korbar et al. ~\cite{korbar2018cooperative} developed audio–visual temporal synchronisation (AVTS), a contrastive approach for AVTS that considers the temporal relationship and flow of video and audio and identifies whether they are synchronised. AVTS employs a regularised cross-entropy loss function similar to the loss functions used for Siamese networks \cite{koch2015siamese}.
AVID+CMA \cite{morgado2021audio} improved the AVC models by employing multiple positive and negative pairs from the same and different modalities using a noise contrastive estimations objective function.

Apart from audio–visual models, vision–text retrieval models were introduced, employing the shared semantic between the visual data and the corresponding text. ViLBERT \cite{lu2019vilbert} extended the BERT language model \cite{devlin2018bert} to jointly learn representations of images and text data through separate networks for each modality, which communicate using novel co-attentional transformer layers.
CLIP \cite{radford2021learning} exploited the use of natural language as a supervisory signal to learn representations from images. Similarly, \cite{yuan2021multimodal} proposed employing the semantic correlation between images and their captions to extract informative representations of each modality. In addition to inter-modality contrastive functions, they also suggested applying modality-specific contrastive objective functions to improve the encoders in learning representations. Hence, they trained encoders for each modality in a self-supervised manner and then mapped the representations onto a shared space using inter-modality contrastive learning. 

Later, many works extended the idea of leveraging the supervisory information embedded in text and images to learn representations from temporal modalities, such as video \cite{stroud2020learning,miech2019howto100m,Miech_2020_CVPR} and audio (HuBERT \cite{hsu2021hubert}). However, not all of these approaches consider temporal information in the input data. For example, \cite{stroud2020learning} learned to classify videos based on their title, description and tags, if available, but did not consider the temporality that can be extracted from the text and video. 
Due to the fact that video is usually paired with audio, and these two modalities can be highly correlated, the audio signal can play the same supervisory role for videos. In contrast to text, audio can be temporally correlated with different frames of the video, which provides a large amount of shared semantic information across both modalities over time. Alwassel et. al. proposed three different approaches to self-supervised learning from paired audio and video samples \cite{alwassel_2020_xdc}. They built their model based on DeepCluster \cite{caron2018deep}, which is a clustering and relabelling technique for learning from single-modality (video or image) data in a self-supervised manner. They proposed three models: 1) Multi-head deep clustering applies the DeepCluster model to each modality and adds a second classification head supervised by another modality. 2) Concatenation deep clustering obtains cluster vectors by concatenating the visual and audio features. 3) Cross-modal deep clustering (XDC), in which each encoder relies exclusively on the clusters learned from the other modality as the supervisory signal. Using modalities as supervisory signals for another modality, they showed that XDC can outperform other supervised and self-supervised baselines in activity recognition from videos.
It is worth noticing that XDC and similar approaches do not take the semantics of the accompanying audio into account. For example, XDC transforms the audio signal into a spectrogram image. Similarly, \cite{wang2021multimodal} learned general audio representations using video, spectrograms and raw waveforms, with contrastive objective functions rather than clustering. However, in this method, they exploited the use of low-resolution video as a supervisory signal for learning quality representations from audio signals. They evaluated the effectiveness of their method across different audio-related tasks, including speaker identification, keyword spotting, language or musical instrument identification.

\begin{table*}[]
\caption{Overview of existing methods considering two modalities. Various modalities are shown as Video \modvideo|Image \modimage|Audio \modaudio|Text \modtext|Inertial/Bio/Environmental Sensors \modsensors}
\label{tab:bimodality_comparison}
\small
\begin{tabularx}{\textwidth}{|p{1.3cm}|p{0.65cm}|p{1.35cm}|p{1.5cm}|p{1.5cm}|X|X|}
\hline
\textbf{Model}                                                   & \textbf{Mod}                                        & \textbf{Category}        & \textbf{Sampling}                                 & \textbf{Objective Function}                                                  & Application                                                                 & \textbf{Contribution}                                                                                                                      \\ \hline
TCN \cite{sermanet2017tcnmultiview}             & \modvideo                            & Contrastive              & cross-view (2-3 view points/camera)    & Triplet Loss                                                                 & Mimicing human actions by robots; Reward function in reinforcement learning & Learning from multiview videos from multiple cameras for robotic applications                                                              \\ \hline
L3-Net \cite{arandjelovic2017look}              & \modimage \modaudio   & Contrastive              & Image and corresponding audio                     & Binary NCE \cite{chopra2005face}                            & Audio source in the image, Cross modal retrieval                            & Detect image and audio correspondence and modality specific representations                                                               \\ \hline
AVC \cite{arandjelovic2018objects}              & \modimage \modaudio   & Contrastive              & Image and corresponding audio                     & Binary NCE                             & Audio source in the image, Cross modal retrieval                            & Improve L3-Net \cite{arandjelovic2017look} by employing Euclidean distance between modality embeddings instead of concatenation.              \\ \hline 
\cite{nagrani2018learnable}                     & \modvideo \modaudio   & Contrastive              & Face-Voice pairs                                  & Binary NCE                             & Biometrics (Person Identification, Face-Speech identification)              & Learns joint embedding between face and voices. Employs curriculum learning to weight hard negatives                                       \\ \hline
AVTS \cite{korbar2018cooperative}               & \multirow{2}{*}{\modvideo \modaudio}   & \multirow{2}{*}{Contrastive}              & \multirow{2}{*}{\shortstack{Video-audio \\ alignment}}                         & \multirow{2}{*}{Binary NCE}                            & \multirow{2}{*}{Audio source detection}                                                      & \multirow{2}{*}{\shortstack{Learning audio-visual repr-\\esentation via detecting vi-\\deos with not-aligned audio }}                                                          \\ \cline{1-1} 
Multisensory \cite{owens2018multisensory}       &   &               &    &     &     &  \\ \hline
VideoBERT \cite{sun2019videobert}               & \modvideo \modtext    &   Pretext                    &    Masked input           &   CrossEntropy                                                   & Action classification, Video captioning                                                                     &   Using bi-directional transformers to predict masked parts of text and video     \\ \hline
XDC\cite{alwassel_2020_xdc}                   & \modvideo \modaudio   & Clustering               & Cross modal                                       & K-means                                                                      & Action recognition, audio classification                                    & Iterate clustering (to generate pseudo-labels) and classification (to tune encoders)                                                      \\ \hline 
SelaVi \cite{asano2020selavi}                   & \modvideo \modaudio   & Clustering               & Cross modal                                       & Sinkhorn Clustering \cite{cuturi2013sinkhorn}               & Video clustering                                                            & Extends SeLA \cite{YM2020Selflabelling} by including audio and learning multiple clustering models. \\ \hline 
\cite{Miech_2020_CVPR}                        & \modvideo \modtext    & Contrastive               &  Crossmodal                                   & MIL-NCE        &  Action recognition / segmentation/localization  & Using instructional video                                                                                                                        \\ \hline 
CSTNet \cite{khurana2020cstnet}  & \modaudio \modtext    & Contrastive      & Crossmodal  & Triplet Loss  & Speech translation  & Semi-hard negative mining / Cross lingual learning                                                                                         \\ \hline
CLIP\cite{radford2021learning}                  & \modimage \modtext    & Contrastive              &  Crossmodal   &  Multi-class N-pair loss  &  OCR, action recognition in videos, geo-localization    &  Learning images from raw text directly   \\ \hline 
\cite{yuan2021multimodal}                       & \modimage \modtext    & Contrastive              &  Cross modal + augmentation  &   InfoNCE  & Image classification /tagging, segmentation, cross-modal retrieval  & Inter and Intra-modality contrastive loss function                                                                                         \\ \hline 
HuBERT \cite{hsu2021hubert}                     & \modaudio \modtext    &  Pretext   & Masked input  & CrossEntropy  & Speech recognition  &  Combined acoustic and language model learned from continuous inputs  \\ \hline 
AVID+CMA \cite{morgado2021audio}                & \modvideo \modaudio   & Contrastive              & Cross modal                                       & InfoNCE                                                                      & Audio-visual retrieval                                                      & Joint inter and intra-modal contrastive learning  
\\ \hline 
CM-CV \cite{jing2021crossview}                  & \modimage                            & Contrastive              & Augmentation, crossmodal                          & TL (views) + CE(modalities)   & Shape retrieval, Segmentation                                               & View and modality-invariant/TL:Triplet loss                        \\ \hline 
COALA \cite{favory2020coala,favory2021learning} & \modaudio \modtext    & Contrastive-Generative & Cross modal                                       & KL divergence + NT-Xent \cite{chen2020simple} & Sound event recognition, music genre recognition, Instrument classification & Reconstruction task + contrastive learning to learn audio from corresponding tags                                   \\ \hline
CRL \cite{ott2022cross}                         & \modimage \modsensors & Contrastive              & Cross modal                                       & Triplet loss                                                                 & Handwriting                                                                 & Novel application                                                                                            \\ \hline 
ViCC \cite{toering2022self}                     & \modvideo                            & Contrastive + Clustering & Augmentation for RGB + opt. flow & 12 x CE considers all positive pairs                               & Video Retrieval, Action Recognition                                         & Alternatively optimize each view based on prototypes learned from the other view                                                    \\ \hline
\end{tabularx}
\end{table*}

However, the abovementioned methods do not consider the temporal correlation across times (inter- or intra- modality). \cite{qian2021spatiotemporal} showed that considering temporal constraints can improve the accuracy and quality of the learned representations. Therefore, another line of research appeared that aimed to maintain the temporal consistency and correlation across time. In this regard, instructional videos became a popular source of video and language data, using automatic speech recognition to generate text from the narrations (audio signal) \cite{Miech_2020_CVPR,miech2019howto100m,sun2019videobert}.

\subsection{Multiple Modality Self-Supervised Recognition Learning}
Multimodal learning is considered a type of multiview learning technique in which data comes from different sources (modalities). For example, Time contrastive networks (TCN) \cite{sermanet2017tcnmultiview} have been proposed for learning representations from multiple views recorded simultaneously from multiple cameras. However, TCN is only capable of considering two sources of data at a time. 
Recently, the idea of cross-modal learning has been extended to more than two modalities. AudioCLIP \cite{guzhov2021audioclip} built on top of the well-known image–text model, CLIP \cite{radford2021learning}, by adding an audio branch and customising the contrastive objective function by contrasting each pair of text–image, text–audio and image–audio separately in a single objective function.
\cite{tian2020contrastive} proposed contrastive multiview coding (\textit{CMC}) with an objective function based on noise contrastive estimation \cite{gut2010NCE}. They proposed two different settings for considering multiple modalities: the core view and full view. In the core view, one view is considered as core and a pair-wise comparison between the core view and all the other views is performed. The full-view formulation of the \textit{CMC} objective function contrasts all pair-wise combinations of existing views. Therefore, the full view has an exponentially increasing cost with respect to the number of views, which means it fails to effectively scale to a large number of views. The idea of CMC has been used to extend contrastive learning models to consider multiple views (e.g. various augmentations of images and multiple sensory views, such as RGB, optical flow and depth) of the same modality. Although the model has been employed only for vision data, it is capable of considering other modalities as well. \cite{cha2021contrastive} applied the full view of CMC to synthetic aperture radar sensor data by extending the number of modalities and the number of augmentations that can be applied to each modality. 

Similarly, Alayrac et al. proposed the multimodal versatile network (MMV) \cite{alayrac2020self} that extended the dual-modal view (text–video) to three modalities and showed that including audio as an additional supervisory signal can help improve the quality of learned representations without annotated data. Therefore they employed a separate convolution-based encoder for each video, text (extracted speech from the audio) and raw audio signal to capture a shared representation between all modalities using a contrastive objective function. In contrast, VATT \cite{akbari2021vatt} suggested using a shared transformer-based backbone across all video, text and audio modalities and showed comparable results to MMV \cite{alayrac2020self}.

Apart from contrastive models, multimodal clustering networks (MCN) \cite{chen2021multimodal} extend the dual-modality clustering-based model (such as XDC \cite{alwassel_2020_xdc} and SeLaVi \cite{asano2020selavi}) to include three modalities (i.e. video, text and audio). However, MCN performs clustering across features from different modalities in the \textit{joint space} to learn multimodal clusters while XDC and SeLaVi apply clustering across each modality individually.

Extending the concept of self-supervised cross-modal learning to the domain of sensors and time-series applications is quite recent. Despite the undeniable need to consider data from many different sensor sources \cite{banos2021opportunistic} and the difficulty in annotating fast-generating time-series data in ubiquitous applications, applications of self-supervised contrastive learning in multi-modal time-series data have still not been extensively studied. In addition, applying existing contrastive methods to multimodal data requires pairwise contrasting mechanisms between modalities \cite{tian2020contrastive,wang2021multimodal}, which results in combinatorial computational complexity. Assuming the dot-product of sample pairs is the atomic operation of the contrastive objective function, the time complexity of contrasting each pair of modalities will be $\mathcal{O}(N^{2})$, where $N$ is the size of the input. Extending the number of modalities, $V \geq 2$ results in $\frac{V!}{(V-2)!}$ different pairs of modalities and a complexity of $\mathcal{O}(V^2.N^{2})$, which is quadratic with respect to the number of modalities $V$.
However, the majority of prior contrastive models required a large number of negative pairs, which significantly increases the space and time complexity, but larger batch sizes can potentially create a larger number of false negative pairs in ubiquitous applications, which can degrade performance \cite{deldari2021tscp2}.

Recently, ColloSSL \cite{jain2022collossl} proposed an end-to-end framework for multi-device sensor applications SSRL based on contrastive learning. ColloSSL is a collaborative device self-supervised learning model that considers data from multiple accelerometers and gyroscope sensors. To extract positive samples, ColloSSL selects the most suitable sensor(s) (device) based on the maximum mean discrepancy metric. The authors of ColloSSL proposed device selection, sampling and contrastive objective functions customised for the application of HAR using motion sensor data. 
Although there are several approaches to SSRL from multivariate sensor data  \cite{deldari2021tscp2,franceschi2019unsupervised}, they are not listed here as they do not consider cross-modal correlation. 
Table \ref{tab:bimodality_comparison} and \ref{tab:modality_comparison} summarise existing cross-modal models proposed for two and more than two modalities, respectively, and the category they belong to along with their main contribution and applications.

\begin{table}[]
\caption{Existing methods that consider more than two modalities. Comparing modalities, categories, target application and their contributions. Various modalities are shown as: Video \modvideo|Image \modimage|Audio \modaudio|Text \modtext|Inertial/Bio/Environmental Sensors \modsensors}
\label{tab:modality_comparison}
\begin{tabularx}{\linewidth}{|c|p{1.3cm}|c|ccccc|p{1.8cm}|X|}
\hline
\multicolumn{2}{|c|}{Model}        & \multicolumn{6}{c|}{Modality}         & \multicolumn{1}{c|}{\multirow{2}{*}{Category}} &
\multirow{2}{*}{Contribution}\\
\multicolumn{1}{|c|}{\begin{sideways}Year\end{sideways}} & Name & \begin{sideways}Joint\end{sideways} & \begin{sideways}Image\end{sideways} & \begin{sideways}Text\end{sideways} & \begin{sideways}Video\end{sideways} & \begin{sideways}Audio\end{sideways} & \begin{sideways}Sensor\end{sideways} &    \multicolumn{1}{c|}{}                       \\ \hline
\multirow{12}{*}{\begin{sideways}2020\end{sideways}} & CMC\cite{tian2020contrastive}& - &\modimage & - & \modvideo & - & - & Contrastive& They proposed two CL-based loss function (full-graph and core-view) to learn joint representation across various augmentations of single  video. They employed Info-NCE for every pair of views.\\ \cline{2-10}
 & MMV\cite{alayrac2020self}& Y & \modimage & \modtext  & \modvideo & \modaudio & - & Contrastive& Learns joint representation from video, audio, text by learning fine-grained audio-visual representations based on INfo-NCE, and contrasting them against text (video narrations) using MIL-NCE \cite{Miech_2020_CVPR}. \\ \cline{2-10}
&MVI \cite{jing2020invariant}& - & \modimage &-&-&-&-& Contrastive & Learns representations from 3D images by considering CL for all \textit{pairwise} combination of {mesh, point cloud, image} (cross-modal) and contrasting different augmentations of 2D images ( cross-view).\\\hline
\multirow{21}{*}{\begin{sideways}2021\end{sideways}}  & \cite{wang2021multimodal} & - & - & - & \modvideo & \modaudio & - & Contrastive & 3-way contrastive function (video, waveform, spectogram). Similar to \textit{full graph} in \cite{tian2020contrastive}, they consider all pair of modalities separately.\\ \cline{2-10}
 & \multirow{2}{*}{\shortstack{AudioCLIP \\  \cite{guzhov2021audioclip}}}& - & \modimage & \modtext & - & \modaudio & - & Contrastive & Applied symmetric cross entropy loss across each pair of modalities (img-txt, img-audio, txt-audio). \\ \cline{2-10}
 & VATT\cite{akbari2021vatt}& Y &  - & \modtext & \modvideo & \modaudio & - & Contrastive& Use the same architecture proposed by \cite{alayrac2020self}, however improved by introducing modality-specefic or modality-agnostic transformers for the encoders.\\ \cline{2-10}
 & MCN\cite{chen2021multimodal}& Y &  - & \modtext & \modvideo & \modaudio &-  &  \multirow{2}{*}{\shortstack{Clustering+\\Contrastive}}& Combining contrastive learning to learn instance based, and clustering to enforce grouping of semantically similar topics.\\ \cline{2-10}
 &  \multirow{2}{*}{\shortstack{ColloSSL \\ \cite{jain2022collossl}}}& Y & - & - & - & - & \modsensors & Contrastive & Offers an end-to-end solution including device selection, pair sampling, and cross-device contrastive objective function based on Info-NCE.\\ \cline{2-10}
 & ELo \cite{piergiovanni2020evolving} & Y &- & - & \modvideo & \modaudio & - & \multirow{2}{*}{\shortstack{Pretext + \\ Clustering}} & Multiple self-supervised tasks followed by clustering for fine-tuning the weights of the pretext tasks using an evolutionary loss function. Finally representations of different modalities (image, audio, optical flow) are infused into the main network by using distillation technique. \\ \hline
 
\multirow{2}{*}{\begin{sideways}2022\end{sideways}}  & \multirow{2}{*}{\shortstack{Data2vec \\  \cite{baevski2022data2vec}}} & Y & - & \modtext & \modvideo & \modaudio & - & \multirow{2}{*}{\shortstack{Pretext + \\ Regularization}}& Extends self-distillation training of BYOL \cite{niizumi2021byol} to 3 modalities combined with transformer encoders. \\
\hline
 
\end{tabularx}
\end{table}


\section {Evolving trend in Self-Supervised Recognition Learning Objective Functions}
\label{sec:loss_fn}
In the absence of annotated data, self-supervised learning models design a predefined function $\mathcal{F}$ to use the structure of data $X$ and generate pseudo labels $Y$. Depending on the function $\mathcal{F}$, the self-supervised model can be categorised into pretext task, contrastive, cluster-based or regularisation-based models. 
In this section, we cover the evolving trend of the objective functions used in each category by mentioning the most notable approaches. Table \ref{tab:loss_table} summarises the proposed objective functions and their main contributions to highlight the evolving trends in self-supervised objective functions. Please refer to Table \ref{tab:notations} for the list of notations used here.

\subsection{Pretext Task} As we discussed in Section \ref{sec:pretext}, pretext-based models are divided into two broad subcategories of \textit{self-prediction} and \textit{innate relation}-based models. In self-prediction (also called masked prediction) models, function $\mathcal{F}$ masks random or specific (usually future) parts of the data and forms a pair of masked and original data as the training input and ground truth labels $(X',Y') = \mathcal{F}(X)$. Other than masking the input data, $\mathcal{F}$ may generate the $X'$ and $Y'$ by generating noisy data from the original data. All these prediction-based approaches train to minimise a reconstruction loss, such as mean squared error (MSE), as their objective function to pretrain the underlying encoders $f$.

In innate relation-based models, the function $\mathcal{F}$ applies one or multiple transformations, $Y'$, over the original data, $X$, and feed the transformed data, $X'$, into the model. The model needs to learn the features of the data by trying to recognise the transformation function. For example, in vision applications, the transformation could be rotating the image, and the pseudo labels are the angle of rotation, or in a multi-channel sensor application, the transformation could be shuffling the data in either temporal or spatial order, and the model needs to predict the correct order. The learning objective function can be a cross-entropy loss function if there are multiple transformations. 

\subsection{Contrastive}
Since there is no information on the type and number of classes, in \textit{contrastive}-based models, each instance is considered as a separate class. However this assumption results in three main issues: 1) a large number of classes, 2) a limited number of examples per class (one instance per class) and 3) unseen classes during inference time. Due to these issues, it is not feasible or efficient to employ the cross-entropy loss function. 
To address the issue of unknown classes,  \cite{chopra2005face,hadsell2006dimensionality} proposed a similarity metric learning method to verify and recognise the related samples (positive pair) among artificially generated noise (negative pair) samples. 
Their idea was to map the samples to a lower-dimensional target space where the distances in the target space approximate the `semantic' distance of the samples in the original input space. Hence, instead of discriminating the negative samples in the original space, the model learned a similarity metric to differentiate between irrelevant samples in the target space. This formed the first line of research on the contrastive objective function, which required some pre-processing to generate the training set, including positive and negative pairs. This idea basically turns the problem of classification with a large number of classes into a problem with binary class labels, where the label is $one$ if a pair of inputs belong to the same sample (i.e. different augmentations of the same sample) and  the label is $zero$ if they belong to separate samples.

Triplet contrastive loss, proposed in \cite{chechik2010large} and \cite{schroff2015facenet}, expands the loss function to consider a triplet of samples (anchor, positive and negative) as $(x,x^{+},x^{-})$ simultaneously in the loss function. In general, the triplet loss function outperforms the contrastive loss because the relationship between the positive and negative pairs are both considered in the loss calculation, whereas the positive and negative pairs are considered separately in the contrastive loss function. However, both approaches suffer from slow convergence and falling into local optima \cite{sohn2016improved} mainly as a result of the following: First, since they consider only one negative pair per update of the loss, the objective function may push the negative pair apart, but they will be closer to other negative samples. Second, since there is only one negative sample per reference sample, the selection of a good negative pair plays an important role.

To address the issue of the limited number of examples per class, ExamplarCNN \cite{dosovitskiy2014examplarcnn} tried to solve the problem by applying multiple and randomly selected transformation functions on each input seed (sample) to create an augmented set of training samples for each instance (class). However, it is still infeasible to apply the cross-entropy objective function across large datasets \cite{ericsson2021survey}. 
Noise contrastive estimation (NCE) \cite{gut2010NCE} proposed a novel objective function with the aim of estimating the cross-entropy loss function by recognising the related samples (positive pair) among the artificially generated noise (negative pairs). Similar to NCE, n-paired \cite{sohn2016improved} and lifted structured \cite{oh2016deep} loss expanded the triplet loss function to consider multiple negative pairs instead of only one pair of positive and negative samples. It is worth noticing that n-paired loss is a special case of \cite{oh2016deep} that proposes a sampling strategy to choose positive pairs from disjoint classes. 
InfoNCE \cite{oord2018representation}, which is probably the most popular contrastive objective function in recent works, is inspired by the NCE loss function and uses categorical cross-entropy loss to identify the positive sample against a set of unrelated noise samples. MIL-NCE \cite{Miech_2020_CVPR} and soft nearest neighbor loss \cite{frosst2019analyzing} improve the InfoNCE objective function to consider multiple positive pairs at the same time.

These approaches, however, require expensive data sampling methods to select nontrivial negative instances for training. Recent studies have shown that harder negative samples (i.e. true negative samples that are close to the anchor) make it more challenging for the encoders to learn distinguishing features and lead to higher performance. However, while the quality of negative samples is crucial in contrastive learning, existing work has typically adopted random negative sampling. This requirement introduces the new concept of hard-negative instance mining, which has been shown to play a critical role in ensuring these cost functions are effective \cite{duan2019deep,wu2017sampling,kalantidis2020hard,NEURIPS2020_63c3ddcc}. Several different sampling strategies have been proposed to address this issue in different ways. Some of the recent works using random sampling strategies showed that covering a more diverse range of negative pairs can increase the performance of the model \cite{hjelm2018learning}. Therefore, these methods increase the number of negative pairs by either increasing the batch size (SimCLR \cite{chen2020simple}) or employing a memory bank that contains all the training samples (MoCo \cite{he2020momentum}). However, increasing the batch size or using memory can create other issues. 

Apart from computation and storage complexity introduced by larger batch sizes and memory usage, respectively, these techniques may increase the number of false negatives, but there is no guarantee that they introduce \textit{hard} negatives \cite{kalantidis2020hard,Mitrovic2020less}. Since we do not know the actual labels, random sampling may accidentally sample false negatives. Several methods addressed this problem by introducing a \textit{debiased} loss function to diminish the effect of fake negatives \cite{NEURIPS2020_63c3ddcc,robinson2021hardcontrastive}. Instead of proposing a new loss function, \cite{huynh2022boosting} tried to solve this issue by proposing a false negative cancellation method to detect and eliminate possible false negative pairs from the set of negative pairs and attract those as positive pairs. Other approaches include hard-negative sampling \cite{simo2015discriminative}, semi-hard mining \cite{schroff2015facenet}, distance weighted sampling \cite{wu2017sampling}, hard-negative class mining \cite{sohn2016improved} and rank-based negative mining \cite{wang2019ranked}. Recently, \cite{kalantidis2020hard} proposed a new hard-negative mining strategy that generates a synthetic hard-negative pair by mixing the most similar negative samples.

Recently, two critical aspects related to contrastive learning loss functions were introduced and evaluated by \cite{Wang_2021_understandingcl}. They introduced an optimisable metric based on (1) closeness (alignment) of positive pairs and (2) uniformity of the distribution of the representations on the hypersphere. They showed that optimising these two metrics leads to better representations for downstream tasks when compared with popular contrastive learning objective functions.
Over the last couple of years, contrastive objective functions have been extended to address cross-modal contrastive learning requirements \cite{akbari2021vatt,jain2022collossl,sermanet2017tcnmultiview,alayrac2020self,tian2020contrastive,wang2021multimodal}.


\begin{table*}[]
\centering
\small
\caption{Summarising existing self-supervised contrastive and regularisation-based learning loss functions (CM shows if the objective function is proposed for cross-modal).}
\label{tab:loss_table}
\begin{tabularx}{\linewidth}{p{1.8cm}|c|X|p{5cm}} 
\hline
\textbf{Method}  & \textbf{CM}*         & \textbf{Objective Function} &  \textbf{Details}   \\ 
\hline

\begin{tabular}[c]{@{}l@{}} Paired \cite{chopra2005learning}\\ (2005) \end{tabular}& - &  \begin{tabular}[c]{@{}l@{}} $ \frac{1}{|N|} \sum_{v,w} $\\ ${y_{vw}^{'} D(z_{v},z_{w})^{2} + (1-y_{vw}^{'}) max\{0,\alpha-D(z_{v},z_{w})\}^{2}} $ \end{tabular}& \multirow{2}{*}{\shortstack{$\alpha$ m is a hyperparameter, defining the lower \\ bound distance between negative pairs.}}
 \\
\cline{1-3}

\begin{tabular}[c]{@{}l@{}} Triplet \cite{schroff2015facenet}\\ (2015) \end{tabular}& - & $max (0, D(z, z^{+})^{2}-D(z, z^{-})^{2}+\alpha) $&   \\
\hline

\begin{tabular}[c]{@{}l@{}} Lifted structure \\ \cite{oh2016deep} (2016) \end{tabular}& - &  \begin{tabular}[c]{@{}l@{}} $\frac{1}{2|N|} \sum_{vw}{max(0, L)}^{2}$ \\
$ L =  D(z_{v},z_{w}) + \sum{}{(\alpha-D(z_{v},z_{v}^{-}))} + \sum{}{(\alpha-D(z_{w},z_{w}^{-}))}$ \end{tabular}
& Generalized the triplet loss to include multiple negative samples.
 \\
\hline
\multirow{2}{*}{ \shortstack{ InfoNCE \\ \cite{oord2018representation,henaff2020data} \\ (2018) }}& - & $- \sum_{x \in P}{log \frac{e^{f(x).f(x^{+})}}{e^{f(x).f(x^{+}))}+\sum_{x^{-}\in N}{e^{f(x).f(x^{-})}}}}$   &  Considers multiple negative pairs. Also called $CPC$ where $x_{v}$ and $x_{w}$ are history and future frames, respectively. \\
\hline

\begin{tabular}[c]{@{}l@{}}  MIL-NCE\cite{Miech_2020_CVPR} \\ (2020) \end{tabular}& - & $ - \sum_{x}{log \frac{\sum_{x^{+} \in P}{e^{f(x).f(x^{+})}}}{\sum_{x^{+} \in P}{e^{f(x).f(x^{+})}}+\sum_{x^{-} \in N}{e^{f(x).f(x^{-})}}}}$                            &  MIL-NCE considers multiple positive candidate pairs by replacing the single term $e^{f(x_{v}^{t}).f(x_{w}^{+})}$ in the standard NCE equation by a sum of scores over positive candidates.  \\
\hline

 \begin{tabular}[c]{@{}l@{}} MMV\cite{alayrac2020self} \\ (2020) \end{tabular}& yes & $ \lambda_{va} \mathcal{L}_{NCE}(x_{v},x_{a}^{+})+\lambda_{vat} \mathcal{L}_{MIL\_NCE}(x_{va},x_{t}^{+})$  & $va$ corresponds to video-audio pairs, and $vat$ points to  corresponding pairs from text and audio-visual modalities.  \\
\hline

\begin{tabular}[c]{@{}l@{}}  alignment-\\uniformity  \cite{wang2020understanding} \\ (2020) \end{tabular}& - & \begin{tabular}[c]{@{}l@{}}  $\mathcal{L}_{align} = \E_{(x,x^{+}) \in P} [||f(x)-f(x^{+})||_{\alpha}^{2}]$ \\  $\mathcal{L}_{uniformity} = log \E_{x,x' \in M} [e^{-t||f(x)-f(x')||_{2}^{2}}]$ \\
 $\mathcal{L}= \mathcal{L}_{align} + \lambda \mathcal{L}_{uniformity}$ \end{tabular}   & Optimizing metric based on aligning the representations of positive pairs and distributing the representations of all instances on the unit sphere.\\
\hline

\begin{tabular}[c]{@{}l@{}}DCL\cite{NEURIPS2020_63c3ddcc} \\ (2020) \end{tabular} & - &                         \begin{tabular}[c]{@{}l@{}} $g = \max \{\frac{1}{\tau} ( \frac{1}{N} \sum{e^{f(x).f(x_{w}^{-})}} + e^{f(x).f(x_{w}^{+})} ) , \epsilon \}$ \\
$\mathcal{L}_{DCL} = - \sum_{t}{log \frac{e^{f(x_{v}^{t}).f(x_{w}^{+})}}{e^{f(x_{v}^{t}).f(x_{w}^{+}))}+ Ng}} $
\end{tabular} 
&  Debiased Contrastive Learning for 1 positive sample and N negative samples across two views $v$ and $w$. The main purpose is to reduce the effect of false negative samples. \\
\hline

\begin{tabular}[c]{@{}l@{}}Hard-DCL\cite{robinson2021hardcontrastive} \\ (2021) \end{tabular}& - &
\begin{tabular}[c]{@{}l@{}} $ w = \frac{\beta e^{f(x,x^{-})}}{\sum_{x^{-}}{e^{f(x,x^{-})}}}$ \\
$g = \max \{\frac{1}{\tau} ( \frac{1}{N} \sum{w * e^{f(x).f(x^{-})}} + e^{f(x).f(x^{+})} ) , \epsilon \}$ \\
$\mathcal{L}_{Hard\_DCL} = - \sum_{t}{log \frac{e^{f(x^{t}).f(x^{+})}}{e^{f(x^{t}).f(x^{+}))}+ Ng}} $
\end{tabular} 
& Improves Debiased Contrastive Learning by increasing the weight of harder negative samples. \\
\hline

\begin{tabular}[c]{@{}l@{}} $CMC$\cite{tian2020contrastive}(2019)\\ \cite{wang2021multimodal} (2021) \end{tabular} &Yes & 
$\mathcal{L}_{CMC} = \sum_{v \in V}{\sum_{w \in V, w \ne v}{\mathcal{L}_{NCE}(v,w)}} $&
calculates the InfoNCE loss over every two-pairs of available views/modalities. \\ \hline

\begin{tabular}[c]{@{}l@{}} \Barlow \\ \cite{zbontar2021barlow} \\ (2021) \end{tabular}& -  &   
\begin{tabular}[c]{@{}l@{}}$C_{ij} = \frac{\sum_{t}{f(x_{i}).f(x_{j})}}{ \sqrt{\sum_{t}{f(x_{i}}^{2}} \cdot \sqrt{\sum_{t}{f(x_{i})}^{2}}}$\\ $\mathcal{L}_{\Barlow} = \sum_{i}{(1-C_{ii})^{2}} + \lambda \sum_{i}{\sum_{j \ne i}{C_{ij}^{2}}}$ \end{tabular}&  Non-contrastive objective function that only considers positive pairs with the aim of   increasing the similarity while reducing the redundancy by minimizing off-diagonal elements of representations' correlation matrix.  \\ 
\hline

\begin{tabular}[c]{@{}l@{}} $W-MSE$ \cite{ermolov2021wmse} \\(2021)\end{tabular} &- &  \begin{tabular}[c]{@{}l@{}} 
$W$ = Whitened representations\\
$D(.)$ = MSE between normalized whitened vectors\\
$x_{1}$ and $x_{2}$ are augmented versions of $x$\\
$ \mathcal{L}=\frac{2}{Pd(d-1)} \sum_{(x_{1},x_{2}) \in P }{D(W(f(x_{1})),W(f(x_{2})))}$ \end{tabular}&
Non-contrastive objective function with the aim of maximizing the similarity between positive pairs while distributing the representations across the unit sphere based on whitening-MSE. \\
\hline

\end{tabularx}
\end{table*}

\subsection{Regularisation}
As mentioned in Section \ref{sec:regularization_model}, unlike contrastive models, regularisation-based models do not require negative pairs. Their objective function is purely based on the similarity of positive pairs, and the negative part is excluded. Regularisation-based models mostly employ symmetric similarity loss functions based on cosine similarity or dot-product across representations of positive samples \cite{niizumi2021byol,chen2021exploring}. However, recent works have introduced a novel objective function along with their architecture-wise contributions \cite{zbontar2021barlow,bardes2022vicreg, ermolov2021wmse}. To summarise the contribution of each and highlight the differences when compared with contrastive models, Table \ref{tab:loss_table} compares their objective functions.

\section{Discussion and Future Directions}
\label{sec:discussion}
In this section, we analyse open challenges in applying self-supervised learning on multimodal time series and discuss future research directions that can potentially address these challenges.

\subsection{Constructing Contrastive Pairs}
An important ingredient for the contrastive learning pipeline is constructing the contrastive pairs. The quality of positive/negative pairs significantly affects the performance of a contrastive learning model.
One common and effective construction strategy in the CV domain is directly applying augmentation techniques (such as rotation and color distortion) on the raw data.
However, augmentation in the raw data space is challenging for time-series data, especially for multimodal and heterogeneous time-series data.
Considering the inherent characteristics of multimodal time-series data, including the complex temporal relationships between different modalities, it is difficult to design augmentation techniques that provide diverse views of the multimodal time-series data, whilst maintaining the representation invariance.
For example, we should not only consider constructing sample pairs from segments at different temporal points (as in single mode scenarios), but also sample pairs from different modalities. In addition, depending upon the specific downstream task, the sample pairs could be constructed from segments of the different modalities at either the same time or different time points. Consequently, there are additional complexities associated with designing the contrastive sampling process with multimodal time-series data.

Targeting this contrastive sampling challenge, we present two possible directions. (1) Contrasting in the latent space: 
Instead of augmenting the temporal data as is commonly done, augmentation could be carried out in latent space (e.g. after the raw data has been encoded) in order to construct contrastive pairs of samples. Such a construction process would provide a new perspective on contrastive learning. Techniques like temporal masking, adding noise or dropout operations (applied on the latent vectors) could be exploited.  
(2) Hierarchical contrasting:
To learn representations at various temporal resolutions, a novel hierarchical contrasting concept has recently been introduced~\cite{yue2021ts2vec}. A stack of convolutional layers was combined with the max pooling operator to generate a multi-scale temporal representation. Sample pairs of the temporal data were constructed from segments with different temporal resolutions and then contrasted against each other. This hierarchical contrasting could also potentially overcome the limitations of searching for optimal window sizes (used for generating samples of temporal clips in the sliding window manner) when constructing contrastive pairs of temporal samples. Further exploration is needed to expand it to multimodal data.

\subsection{Domain-Agnostic Representation Learning}

In this rapidly changing and uncertain context, one challenge related to gaining valuable information from temporal data is quickly adapting and applying a self-supervised learning model to new tasks and environments. 
For example, time-series anomaly detection tasks and forecasting tasks are often related. A forecasting system with the ability to detect anomalies would be beneficial in many applications.
Currently, in the self-supervised learning paradigm, pretraining (e.g. through pretext tasks or contrastive learning) is the initial step in the model training strategy. The pretrained models are then used (after being fine-tuned) for only one specific downstream task.
Instead, the goal of the general framework is that the unified representations learned from the framework are expected to be applied to different time-series tasks, such as classification, forecasting, change point detection and anomaly detection. A key consideration here is whether the general framework can effectively and efficiently learn high-quality and domain-agnostic features of the multimodal data that are flexible and robust \cite{tamkin2021dabs}.
In a recent study, ~\cite{yue2021ts2vec} first explored a flexible and universal representation contrastive learning-based framework for all kinds of tasks in the time-series domain. The proposed TS2Vec structure has demonstrated a generalisation capability in the single mode setting.
Nevertheless, how to design a universal representation learning framework for multimodal temporal data remains unexplored.
Under the multimodal setting, another valuable research question relates to dataset bias and the implication of this for transferring results across different tasks and datasets. For instance, the bias towards a specific modality under a particular task could undermine the quality of the learned representations under another task.
Further study into a general multimodal self-supervised learning-based architecture would be a promising research direction.

\subsection{Robustness to Irregular Data}
In real-world scenarios, temporal data collected in the wild often has issues, such as 
sequences having irregular sampling intervals and/or data gaps. Furthermore, multimodal temporal data
can be sampled at different rates leading to temporal misalignment between the samples of the different modes.
Such irregularities can be challenge when applying self-supervised learning-based models to multimodal temporal data.
By injecting time synchronisation errors into datasets,~\cite{jain2022collossl} recently analysed the robustness of its proposed SSRL model to temporal misalignment.
This experiment provided valuable insight into evaluating self-supervised learning frameworks 
with data in the wild. To handle the irregularity of temporal data, recent studies presented ordinary differential equation (ODE)-based models, such as ODE-RNNs~\cite{rubanova2019latent} and GRU-ODE-Bayes~\cite{de2019gru}, which have shown promising results. In addition, \cite{abushaqra2021piets} demonstrated that introducing irregularity encoders (for multimodal data sources) with attention mechanisms in parallelised structures can alleviate the influence of irregularity. Investigating approaches to integrate these techniques within the self-supervised learning paradigm would be an interesting direction for future investigation.

\subsection{Effective Augmentation and Positive Sampling }

In self-supervised learning, encoders are used to learn compact, informative feature vectors of the original data that can identify irrelevant samples. The quality of the final representation depends upon the use of data augmentations that are appropriate for the downstream analysis task and combination of views. If the input views (whether various augmented views or different modalities) are highly redundant, the encoders will learn trivial features that fail to provide useful representations \cite{minderer2020automatic}.  High levels of mutual information between positive pairs can lead to the model taking a short cut during training to learn redundant, uninformative features rather than distinguishing features \cite{tian2020makes}.
In the field of CV, much research has been conducted to carefully design or handcraft an appropriate set of augmentation techniques regarding the target task. However, in the field of cross-modal and temporal data, this issue has not yet been fully explored. For example, in the case of temporal data, we need to ensure that positive pairs that are in temporal order (e.g. subsequent frames) present a high amount of correlation. Similarly, in the case of multimodal data, different sources of temporal data may show high levels of mutual information (e.g. accelerometer and gyroscope sensors in human activity datasets). In both cases, although the encoder learns similar representations, they are not useful since they are only reflecting upon the similarities and not the distinguishing features of data. \cite{tian2020makes} explored the importance of view selection strategy and showed that there is an optimum amount of mutual information between views that is neither too high nor too low. For future works in cross-modal self-supervised learning across temporal data, it is critical to design appropriate modality selection (also known as device/channel selection) or augmentation techniques while considering downstream tasks.

\section{Conclusion}
In this work, we reviewed existing self-supervised approaches in learning representation vectors from various sources of data. We specifically concentrated on approaches proposed for multi-modal and temporal data, as these methods were not explored in existing review papers. To provide a comprehensive review, we introduced a new multi-attribute categorisation, which builds upon previous ones by including additional attributes; the input sampling technique, pseudo-label generation technique, training strategy and objective function. We then categorised discriminative (non-generative) SSRL models into four main categories: pretext, contrastive, clustering and regularisation-based models. Each category was introduced and outlined with respect to existing methods designed for single mode temporal data-sets. We then explored multimodal methods with respect to these existing categories and their specific requirements. 
Based on our review of existing SSRL approaches, we identified several challenges and weaknesses within existing methods, which need to be addressed in future research. We hope that this review helps the community in understanding the current limitations of SSRL for multi-modal temporal data, and where new research opportunities exist.


\begin{acks}
Authors would like to acknowledge the support from CSIRO Data61 Scholarship program (Grant number 500588), RMIT Research International Tuition Fee Scholarship and Australian Research Council (ARC) Discovery Project DP190101485.
\end{acks}


\bibliographystyle{ACM-Reference-Format}
\bibliography{main}


\begin{thebibliography}{162}


\ifx \showCODEN    \undefined \def \showCODEN     #1{\unskip}     \fi
\ifx \showDOI      \undefined \def \showDOI       #1{#1}\fi
\ifx \showISBNx    \undefined \def \showISBNx     #1{\unskip}     \fi
\ifx \showISBNxiii \undefined \def \showISBNxiii  #1{\unskip}     \fi
\ifx \showISSN     \undefined \def \showISSN      #1{\unskip}     \fi
\ifx \showLCCN     \undefined \def \showLCCN      #1{\unskip}     \fi
\ifx \shownote     \undefined \def \shownote      #1{#1}          \fi
\ifx \showarticletitle \undefined \def \showarticletitle #1{#1}   \fi
\ifx \showURL      \undefined \def \showURL       {\relax}        \fi
\providecommand\bibfield[2]{#2}
\providecommand\bibinfo[2]{#2}
\providecommand\natexlab[1]{#1}
\providecommand\showeprint[2][]{arXiv:#2}

\bibitem[\protect\citeauthoryear{Abushaqra, Xue, Ren, and Salim}{Abushaqra
  et~al\mbox{.}}{2021}]%
        {abushaqra2021piets}
\bibfield{author}{\bibinfo{person}{Futoon~M Abushaqra}, \bibinfo{person}{Hao
  Xue}, \bibinfo{person}{Yongli Ren}, {and} \bibinfo{person}{Flora~D Salim}.}
  \bibinfo{year}{2021}\natexlab{}.
\newblock \showarticletitle{PIETS: Parallelised Irregularity Encoders for
  Forecasting with Heterogeneous Time-Series}. In
  \bibinfo{booktitle}{\emph{2021 IEEE International Conference on Data Mining
  (ICDM)}}. IEEE, \bibinfo{pages}{976--981}.
\newblock


\bibitem[\protect\citeauthoryear{Akbari, Yuan, Qian, Chuang, Chang, Cui, and
  Gong}{Akbari et~al\mbox{.}}{2021}]%
        {akbari2021vatt}
\bibfield{author}{\bibinfo{person}{Hassan Akbari}, \bibinfo{person}{Linagzhe
  Yuan}, \bibinfo{person}{Rui Qian}, \bibinfo{person}{Wei-Hong Chuang},
  \bibinfo{person}{Shih-Fu Chang}, \bibinfo{person}{Yin Cui}, {and}
  \bibinfo{person}{Boqing Gong}.} \bibinfo{year}{2021}\natexlab{}.
\newblock \showarticletitle{Vatt: Transformers for multimodal self-supervised
  learning from raw video, audio and text}.
\newblock \bibinfo{journal}{\emph{arXiv preprint arXiv:2104.11178}}
  (\bibinfo{year}{2021}).
\newblock


\bibitem[\protect\citeauthoryear{Alayrac, Recasens, Schneider, Arandjelovic,
  Ramapuram, De~Fauw, Smaira, Dieleman, and Zisserman}{Alayrac
  et~al\mbox{.}}{2020}]%
        {alayrac2020self}
\bibfield{author}{\bibinfo{person}{Jean-Baptiste Alayrac},
  \bibinfo{person}{Adria Recasens}, \bibinfo{person}{Rosalia Schneider},
  \bibinfo{person}{Relja Arandjelovic}, \bibinfo{person}{Jason Ramapuram},
  \bibinfo{person}{Jeffrey De~Fauw}, \bibinfo{person}{Lucas Smaira},
  \bibinfo{person}{Sander Dieleman}, {and} \bibinfo{person}{Andrew Zisserman}.}
  \bibinfo{year}{2020}\natexlab{}.
\newblock \showarticletitle{Self-Supervised MultiModal Versatile Networks}.
\newblock \bibinfo{journal}{\emph{NeurIPS}} \bibinfo{volume}{2},
  \bibinfo{number}{6} (\bibinfo{year}{2020}), \bibinfo{pages}{7}.
\newblock


\bibitem[\protect\citeauthoryear{Almalioglu, Turan, Saputra, {de Gusmão},
  Markham, and Trigoni}{Almalioglu et~al\mbox{.}}{2022}]%
        {ALMALIOGLU2022119}
\bibfield{author}{\bibinfo{person}{Yasin Almalioglu}, \bibinfo{person}{Mehmet
  Turan}, \bibinfo{person}{Muhamad Risqi~U. Saputra},
  \bibinfo{person}{Pedro~P.B. {de Gusmão}}, \bibinfo{person}{Andrew Markham},
  {and} \bibinfo{person}{Niki Trigoni}.} \bibinfo{year}{2022}\natexlab{}.
\newblock \showarticletitle{SelfVIO: Self-supervised deep monocular
  Visual–Inertial Odometry and depth estimation}.
\newblock \bibinfo{journal}{\emph{Neural Networks}}  \bibinfo{volume}{150}
  (\bibinfo{year}{2022}), \bibinfo{pages}{119--136}.
\newblock


\bibitem[\protect\citeauthoryear{Alwassel, Mahajan, Korbar, Torresani, Ghanem,
  and Tran}{Alwassel et~al\mbox{.}}{2020}]%
        {alwassel_2020_xdc}
\bibfield{author}{\bibinfo{person}{Humam Alwassel}, \bibinfo{person}{Dhruv
  Mahajan}, \bibinfo{person}{Bruno Korbar}, \bibinfo{person}{Lorenzo
  Torresani}, \bibinfo{person}{Bernard Ghanem}, {and} \bibinfo{person}{Du
  Tran}.} \bibinfo{year}{2020}\natexlab{}.
\newblock \showarticletitle{Self-Supervised Learning by Cross-Modal Audio-Video
  Clustering}. In \bibinfo{booktitle}{\emph{Advances in Neural Information
  Processing Systems (NeurIPS)}}.
\newblock


\bibitem[\protect\citeauthoryear{Amrani, Ben-Ari, Hakim, and Bronstein}{Amrani
  et~al\mbox{.}}{2019}]%
        {amrani2019learning}
\bibfield{author}{\bibinfo{person}{Elad Amrani}, \bibinfo{person}{Rami
  Ben-Ari}, \bibinfo{person}{Tal Hakim}, {and} \bibinfo{person}{Alex
  Bronstein}.} \bibinfo{year}{2019}\natexlab{}.
\newblock \showarticletitle{Learning to detect and retrieve objects from
  unlabeled videos}. In \bibinfo{booktitle}{\emph{2019 IEEE/CVF International
  Conference on Computer Vision Workshop (ICCVW)}}. IEEE,
  \bibinfo{pages}{3713--3717}.
\newblock


\bibitem[\protect\citeauthoryear{Amrani, Ben-Ari, Rotman, and Bronstein}{Amrani
  et~al\mbox{.}}{2021}]%
        {Amrani2021noise}
\bibfield{author}{\bibinfo{person}{Elad Amrani}, \bibinfo{person}{Rami
  Ben-Ari}, \bibinfo{person}{Daniel Rotman}, {and} \bibinfo{person}{Alex
  Bronstein}.} \bibinfo{year}{2021}\natexlab{}.
\newblock \showarticletitle{Noise Estimation Using Density Estimation for
  Self-Supervised Multimodal Learning}.
\newblock \bibinfo{journal}{\emph{Proceedings of the AAAI Conference on
  Artificial Intelligence}} \bibinfo{volume}{35}, \bibinfo{number}{8}
  (\bibinfo{date}{May} \bibinfo{year}{2021}).
\newblock


\bibitem[\protect\citeauthoryear{Arandjelovic and Zisserman}{Arandjelovic and
  Zisserman}{2017}]%
        {arandjelovic2017look}
\bibfield{author}{\bibinfo{person}{Relja Arandjelovic} {and}
  \bibinfo{person}{Andrew Zisserman}.} \bibinfo{year}{2017}\natexlab{}.
\newblock \showarticletitle{Look, listen and learn}. In
  \bibinfo{booktitle}{\emph{Proceedings of the IEEE International Conference on
  Computer Vision}}.
\newblock


\bibitem[\protect\citeauthoryear{Arandjelovic and Zisserman}{Arandjelovic and
  Zisserman}{2018}]%
        {arandjelovic2018objects}
\bibfield{author}{\bibinfo{person}{Relja Arandjelovic} {and}
  \bibinfo{person}{Andrew Zisserman}.} \bibinfo{year}{2018}\natexlab{}.
\newblock \showarticletitle{Objects that sound}. In
  \bibinfo{booktitle}{\emph{Proceedings of the European conference on computer
  vision (ECCV)}}. \bibinfo{pages}{435--451}.
\newblock


\bibitem[\protect\citeauthoryear{Asano, Patrick, Rupprecht, and Vedaldi}{Asano
  et~al\mbox{.}}{2020}]%
        {asano2020selavi}
\bibfield{author}{\bibinfo{person}{Yuki Asano}, \bibinfo{person}{Mandela
  Patrick}, \bibinfo{person}{Christian Rupprecht}, {and}
  \bibinfo{person}{Andrea Vedaldi}.} \bibinfo{year}{2020}\natexlab{}.
\newblock \showarticletitle{Labelling unlabelled videos from scratch with
  multi-modal self-supervision}.
\newblock \bibinfo{journal}{\emph{Advances in Neural Information Processing
  Systems}}  \bibinfo{volume}{33} (\bibinfo{year}{2020}),
  \bibinfo{pages}{4660--4671}.
\newblock


\bibitem[\protect\citeauthoryear{Aytar, Vondrick, and Torralba}{Aytar
  et~al\mbox{.}}{2016}]%
        {aytar2016soundnet}
\bibfield{author}{\bibinfo{person}{Yusuf Aytar}, \bibinfo{person}{Carl
  Vondrick}, {and} \bibinfo{person}{Antonio Torralba}.}
  \bibinfo{year}{2016}\natexlab{}.
\newblock \showarticletitle{Soundnet: Learning sound representations from
  unlabeled video}.
\newblock \bibinfo{journal}{\emph{Advances in neural information processing
  systems}}  \bibinfo{volume}{29} (\bibinfo{year}{2016}),
  \bibinfo{pages}{892--900}.
\newblock


\bibitem[\protect\citeauthoryear{Baevski, Auli, and Mohamed}{Baevski
  et~al\mbox{.}}{2019}]%
        {baevski2019effectiveness}
\bibfield{author}{\bibinfo{person}{Alexei Baevski}, \bibinfo{person}{Michael
  Auli}, {and} \bibinfo{person}{Abdelrahman Mohamed}.}
  \bibinfo{year}{2019}\natexlab{}.
\newblock \showarticletitle{Effectiveness of self-supervised pre-training for
  speech recognition}.
\newblock \bibinfo{journal}{\emph{arXiv preprint arXiv:1911.03912}}
  (\bibinfo{year}{2019}).
\newblock


\bibitem[\protect\citeauthoryear{Baevski, Hsu, Xu, Babu, Gu, and Auli}{Baevski
  et~al\mbox{.}}{2022}]%
        {baevski2022data2vec}
\bibfield{author}{\bibinfo{person}{Alexei Baevski}, \bibinfo{person}{Wei-Ning
  Hsu}, \bibinfo{person}{Qiantong Xu}, \bibinfo{person}{Arun Babu},
  \bibinfo{person}{Jiatao Gu}, {and} \bibinfo{person}{Michael Auli}.}
  \bibinfo{year}{2022}\natexlab{}.
\newblock \showarticletitle{Data2vec: A general framework for self-supervised
  learning in speech, vision and language}.
\newblock \bibinfo{journal}{\emph{arXiv preprint arXiv:2202.03555}}
  (\bibinfo{year}{2022}).
\newblock


\bibitem[\protect\citeauthoryear{Baevski, Schneider, and Auli}{Baevski
  et~al\mbox{.}}{2020a}]%
        {BaevskiSA20}
\bibfield{author}{\bibinfo{person}{Alexei Baevski}, \bibinfo{person}{Steffen
  Schneider}, {and} \bibinfo{person}{Michael Auli}.}
  \bibinfo{year}{2020}\natexlab{a}.
\newblock \showarticletitle{vq-wav2vec: Self-Supervised Learning of Discrete
  Speech Representations}. In \bibinfo{booktitle}{\emph{8th International
  Conference on Learning Representations, {ICLR} 2020, Addis Ababa, Ethiopia,
  April 26-30, 2020}}.
\newblock


\bibitem[\protect\citeauthoryear{Baevski, Zhou, Mohamed, and Auli}{Baevski
  et~al\mbox{.}}{2020b}]%
        {baevski2020wav2vec}
\bibfield{author}{\bibinfo{person}{Alexei Baevski}, \bibinfo{person}{Yuhao
  Zhou}, \bibinfo{person}{Abdelrahman Mohamed}, {and} \bibinfo{person}{Michael
  Auli}.} \bibinfo{year}{2020}\natexlab{b}.
\newblock \showarticletitle{wav2vec 2.0: {A} Framework for Self-Supervised
  Learning of Speech Representations}. In \bibinfo{booktitle}{\emph{Advances in
  Neural Information Processing Systems (NeurIPS)}}.
\newblock


\bibitem[\protect\citeauthoryear{Banos, Calatroni, Damas, Pomares, Roggen,
  Rojas, and Villalonga}{Banos et~al\mbox{.}}{2021}]%
        {banos2021opportunistic}
\bibfield{author}{\bibinfo{person}{Oresti Banos}, \bibinfo{person}{Alberto
  Calatroni}, \bibinfo{person}{Miguel Damas}, \bibinfo{person}{Hector Pomares},
  \bibinfo{person}{Daniel Roggen}, \bibinfo{person}{Ignacio Rojas}, {and}
  \bibinfo{person}{Claudia Villalonga}.} \bibinfo{year}{2021}\natexlab{}.
\newblock \showarticletitle{Opportunistic activity recognition in IoT sensor
  ecosystems via multimodal transfer learning}.
\newblock \bibinfo{journal}{\emph{Neural Processing Letters}}
  \bibinfo{volume}{53}, \bibinfo{number}{5} (\bibinfo{year}{2021}),
  \bibinfo{pages}{3169--3197}.
\newblock


\bibitem[\protect\citeauthoryear{Banville, Chehab, Hyv{\"a}rinen, Engemann, and
  Gramfort}{Banville et~al\mbox{.}}{2021}]%
        {banville2021uncovering}
\bibfield{author}{\bibinfo{person}{Hubert Banville}, \bibinfo{person}{Omar
  Chehab}, \bibinfo{person}{Aapo Hyv{\"a}rinen},
  \bibinfo{person}{Denis-Alexander Engemann}, {and} \bibinfo{person}{Alexandre
  Gramfort}.} \bibinfo{year}{2021}\natexlab{}.
\newblock \showarticletitle{Uncovering the structure of clinical EEG signals
  with self-supervised learning}.
\newblock \bibinfo{journal}{\emph{Journal of Neural Engineering}}
  \bibinfo{volume}{18}, \bibinfo{number}{4} (\bibinfo{year}{2021}),
  \bibinfo{pages}{046020}.
\newblock


\bibitem[\protect\citeauthoryear{Bardes, Ponce, and LeCun}{Bardes
  et~al\mbox{.}}{2022}]%
        {bardes2022vicreg}
\bibfield{author}{\bibinfo{person}{Adrien Bardes}, \bibinfo{person}{Jean
  Ponce}, {and} \bibinfo{person}{Yann LeCun}.} \bibinfo{year}{2022}\natexlab{}.
\newblock \showarticletitle{{VICR}eg: Variance-Invariance-Covariance
  Regularization for Self-Supervised Learning}. In
  \bibinfo{booktitle}{\emph{International Conference on Learning
  Representations}}.
\newblock
\urldef\tempurl%
\url{https://openreview.net/forum?id=xm6YD62D1Ub}
\showURL{%
\tempurl}


\bibitem[\protect\citeauthoryear{Barnum, Talukder, and Yue}{Barnum
  et~al\mbox{.}}{2020}]%
        {barnum2020benefits}
\bibfield{author}{\bibinfo{person}{George Barnum}, \bibinfo{person}{Sabera
  Talukder}, {and} \bibinfo{person}{Yisong Yue}.}
  \bibinfo{year}{2020}\natexlab{}.
\newblock \showarticletitle{On the Benefits of Early Fusion in Multimodal
  Representation Learning}.
\newblock \bibinfo{journal}{\emph{arXiv preprint arXiv:2011.07191}}
  (\bibinfo{year}{2020}).
\newblock


\bibitem[\protect\citeauthoryear{Bengio, Courville, and Vincent}{Bengio
  et~al\mbox{.}}{2013}]%
        {bengio2013representation}
\bibfield{author}{\bibinfo{person}{Yoshua Bengio}, \bibinfo{person}{Aaron
  Courville}, {and} \bibinfo{person}{Pascal Vincent}.}
  \bibinfo{year}{2013}\natexlab{}.
\newblock \showarticletitle{Representation learning: A review and new
  perspectives}.
\newblock \bibinfo{journal}{\emph{IEEE transactions on pattern analysis and
  machine intelligence}} \bibinfo{volume}{35}, \bibinfo{number}{8}
  (\bibinfo{year}{2013}), \bibinfo{pages}{1798--1828}.
\newblock


\bibitem[\protect\citeauthoryear{Bengio, Lamblin, Popovici, and
  Larochelle}{Bengio et~al\mbox{.}}{2007}]%
        {bengio2007greedy}
\bibfield{author}{\bibinfo{person}{Yoshua Bengio}, \bibinfo{person}{Pascal
  Lamblin}, \bibinfo{person}{Dan Popovici}, {and} \bibinfo{person}{Hugo
  Larochelle}.} \bibinfo{year}{2007}\natexlab{}.
\newblock \showarticletitle{Greedy layer-wise training of deep networks}. In
  \bibinfo{booktitle}{\emph{Advances in neural information processing
  systems}}. \bibinfo{pages}{153--160}.
\newblock


\bibitem[\protect\citeauthoryear{Blondel, Martins, and Niculae}{Blondel
  et~al\mbox{.}}{2020}]%
        {blondel2020learning}
\bibfield{author}{\bibinfo{person}{Mathieu Blondel},
  \bibinfo{person}{Andr{\'e}~FT Martins}, {and} \bibinfo{person}{Vlad
  Niculae}.} \bibinfo{year}{2020}\natexlab{}.
\newblock \showarticletitle{Learning with Fenchel-Young losses.}
\newblock \bibinfo{journal}{\emph{J. Mach. Learn. Res.}} \bibinfo{volume}{21},
  \bibinfo{number}{35} (\bibinfo{year}{2020}), \bibinfo{pages}{1--69}.
\newblock


\bibitem[\protect\citeauthoryear{Caron, Bojanowski, Joulin, and Douze}{Caron
  et~al\mbox{.}}{2018}]%
        {caron2018deep}
\bibfield{author}{\bibinfo{person}{Mathilde Caron}, \bibinfo{person}{Piotr
  Bojanowski}, \bibinfo{person}{Armand Joulin}, {and} \bibinfo{person}{Matthijs
  Douze}.} \bibinfo{year}{2018}\natexlab{}.
\newblock \showarticletitle{Deep clustering for unsupervised learning of visual
  features}. In \bibinfo{booktitle}{\emph{Proceedings of the European
  Conference on Computer Vision (ECCV)}}. \bibinfo{pages}{132--149}.
\newblock


\bibitem[\protect\citeauthoryear{Caron, Misra, Mairal, Goyal, Bojanowski, and
  Joulin}{Caron et~al\mbox{.}}{2020}]%
        {caron2020swav}
\bibfield{author}{\bibinfo{person}{Mathilde Caron}, \bibinfo{person}{Ishan
  Misra}, \bibinfo{person}{Julien Mairal}, \bibinfo{person}{Priya Goyal},
  \bibinfo{person}{Piotr Bojanowski}, {and} \bibinfo{person}{Armand Joulin}.}
  \bibinfo{year}{2020}\natexlab{}.
\newblock \showarticletitle{Unsupervised learning of visual features by
  contrasting cluster assignments}. In \bibinfo{booktitle}{\emph{Advances in
  Neural Information Processing Systems (NeurIPS)}}.
\newblock


\bibitem[\protect\citeauthoryear{Caron, Touvron, Misra, J{\'{e}}gou, Mairal,
  Bojanowski, and Joulin}{Caron et~al\mbox{.}}{2021}]%
        {caron2021dino}
\bibfield{author}{\bibinfo{person}{Mathilde Caron}, \bibinfo{person}{Hugo
  Touvron}, \bibinfo{person}{Ishan Misra}, \bibinfo{person}{Herv{\'{e}}
  J{\'{e}}gou}, \bibinfo{person}{Julien Mairal}, \bibinfo{person}{Piotr
  Bojanowski}, {and} \bibinfo{person}{Armand Joulin}.}
  \bibinfo{year}{2021}\natexlab{}.
\newblock \showarticletitle{Emerging Properties in Self-Supervised Vision
  Transformers}. In \bibinfo{booktitle}{\emph{2021 {IEEE/CVF} International
  Conference on Computer Vision, {ICCV}}}. \bibinfo{publisher}{{IEEE}},
  \bibinfo{pages}{9630--9640}.
\newblock


\bibitem[\protect\citeauthoryear{Carr, Berthet, Blondel, Teboul, and
  Zeghidour}{Carr et~al\mbox{.}}{2021}]%
        {CarrBBTZ21}
\bibfield{author}{\bibinfo{person}{Andrew~N Carr}, \bibinfo{person}{Quentin
  Berthet}, \bibinfo{person}{Mathieu Blondel}, \bibinfo{person}{Olivier
  Teboul}, {and} \bibinfo{person}{Neil Zeghidour}.}
  \bibinfo{year}{2021}\natexlab{}.
\newblock \showarticletitle{Self-Supervised Learning of Audio Representations
  From Permutations With Differentiable Ranking}.
\newblock \bibinfo{journal}{\emph{IEEE Signal Processing Letters}}
  \bibinfo{volume}{28} (\bibinfo{year}{2021}), \bibinfo{pages}{708--712}.
\newblock


\bibitem[\protect\citeauthoryear{Cha, Seo, and Choi}{Cha et~al\mbox{.}}{2021}]%
        {cha2021contrastive}
\bibfield{author}{\bibinfo{person}{Keumgang Cha}, \bibinfo{person}{Junghoon
  Seo}, {and} \bibinfo{person}{Yeji Choi}.} \bibinfo{year}{2021}\natexlab{}.
\newblock \showarticletitle{Contrastive Multiview Coding With Electro-Optics
  for SAR Semantic Segmentation}.
\newblock \bibinfo{journal}{\emph{IEEE Geoscience and Remote Sensing Letters}}
  \bibinfo{volume}{19} (\bibinfo{year}{2021}), \bibinfo{pages}{1--5}.
\newblock


\bibitem[\protect\citeauthoryear{Chechik, Sharma, Shalit, and Bengio}{Chechik
  et~al\mbox{.}}{2010}]%
        {chechik2010large}
\bibfield{author}{\bibinfo{person}{Gal Chechik}, \bibinfo{person}{Varun
  Sharma}, \bibinfo{person}{Uri Shalit}, {and} \bibinfo{person}{Samy Bengio}.}
  \bibinfo{year}{2010}\natexlab{}.
\newblock \showarticletitle{Large Scale Online Learning of Image Similarity
  Through Ranking.}
\newblock \bibinfo{journal}{\emph{Journal of Machine Learning Research}}
  \bibinfo{volume}{11}, \bibinfo{number}{3} (\bibinfo{year}{2010}).
\newblock


\bibitem[\protect\citeauthoryear{Chen, Rouditchenko, Duarte, Kuehne, Thomas,
  Boggust, Panda, Kingsbury, Feris, Harwath, et~al\mbox{.}}{Chen
  et~al\mbox{.}}{2021}]%
        {chen2021multimodal}
\bibfield{author}{\bibinfo{person}{Brian Chen}, \bibinfo{person}{Andrew
  Rouditchenko}, \bibinfo{person}{Kevin Duarte}, \bibinfo{person}{Hilde
  Kuehne}, \bibinfo{person}{Samuel Thomas}, \bibinfo{person}{Angie Boggust},
  \bibinfo{person}{Rameswar Panda}, \bibinfo{person}{Brian Kingsbury},
  \bibinfo{person}{Rogerio Feris}, \bibinfo{person}{David Harwath},
  {et~al\mbox{.}}} \bibinfo{year}{2021}\natexlab{}.
\newblock \showarticletitle{Multimodal clustering networks for self-supervised
  learning from unlabeled videos}. In \bibinfo{booktitle}{\emph{ICCV}}.
\newblock


\bibitem[\protect\citeauthoryear{Chen, Rosa, Miao, Lu, Wu, Markham, and
  Trigoni}{Chen et~al\mbox{.}}{2019}]%
        {chen2019selective}
\bibfield{author}{\bibinfo{person}{Changhao Chen}, \bibinfo{person}{Stefano
  Rosa}, \bibinfo{person}{Yishu Miao}, \bibinfo{person}{Chris~Xiaoxuan Lu},
  \bibinfo{person}{Wei Wu}, \bibinfo{person}{Andrew Markham}, {and}
  \bibinfo{person}{Niki Trigoni}.} \bibinfo{year}{2019}\natexlab{}.
\newblock \showarticletitle{Selective sensor fusion for neural visual-inertial
  odometry}. In \bibinfo{booktitle}{\emph{Proceedings of the IEEE/CVF
  Conference on Computer Vision and Pattern Recognition}}.
  \bibinfo{pages}{10542--10551}.
\newblock


\bibitem[\protect\citeauthoryear{Chen, Kornblith, Norouzi, and Hinton}{Chen
  et~al\mbox{.}}{2020}]%
        {chen2020simple}
\bibfield{author}{\bibinfo{person}{Ting Chen}, \bibinfo{person}{Simon
  Kornblith}, \bibinfo{person}{Mohammad Norouzi}, {and}
  \bibinfo{person}{Geoffrey Hinton}.} \bibinfo{year}{2020}\natexlab{}.
\newblock \showarticletitle{A simple framework for contrastive learning of
  visual representations}. In \bibinfo{booktitle}{\emph{International
  conference on machine learning}}. PMLR, \bibinfo{pages}{1597--1607}.
\newblock


\bibitem[\protect\citeauthoryear{Chen and He}{Chen and He}{2021}]%
        {chen2021exploring}
\bibfield{author}{\bibinfo{person}{Xinlei Chen} {and} \bibinfo{person}{Kaiming
  He}.} \bibinfo{year}{2021}\natexlab{}.
\newblock \showarticletitle{Exploring simple siamese representation learning}.
  In \bibinfo{booktitle}{\emph{Proceedings of the IEEE/CVF Conference on
  Computer Vision and Pattern Recognition}}. \bibinfo{pages}{15750--15758}.
\newblock


\bibitem[\protect\citeauthoryear{Cheng, Goh, Dogrusoz, Tuzel, and Azemi}{Cheng
  et~al\mbox{.}}{2020}]%
        {cheng2020subject}
\bibfield{author}{\bibinfo{person}{Joseph~Y Cheng}, \bibinfo{person}{Hanlin
  Goh}, \bibinfo{person}{Kaan Dogrusoz}, \bibinfo{person}{Oncel Tuzel}, {and}
  \bibinfo{person}{Erdrin Azemi}.} \bibinfo{year}{2020}\natexlab{}.
\newblock \showarticletitle{Subject-aware contrastive learning for biosignals}.
\newblock \bibinfo{journal}{\emph{arXiv preprint arXiv:2007.04871}}
  (\bibinfo{year}{2020}).
\newblock


\bibitem[\protect\citeauthoryear{Chi, Chung, Wu, Hsieh, Chen, Li, and Lee}{Chi
  et~al\mbox{.}}{2021}]%
        {chi2021audio}
\bibfield{author}{\bibinfo{person}{Po-Han Chi}, \bibinfo{person}{Pei-Hung
  Chung}, \bibinfo{person}{Tsung-Han Wu}, \bibinfo{person}{Chun-Cheng Hsieh},
  \bibinfo{person}{Yen-Hao Chen}, \bibinfo{person}{Shang-Wen Li}, {and}
  \bibinfo{person}{Hung-yi Lee}.} \bibinfo{year}{2021}\natexlab{}.
\newblock \showarticletitle{Audio albert: A lite bert for self-supervised
  learning of audio representation}. In \bibinfo{booktitle}{\emph{2021 IEEE
  Spoken Language Technology Workshop (SLT)}}. IEEE, \bibinfo{pages}{344--350}.
\newblock


\bibitem[\protect\citeauthoryear{Chopra, Hadsell, and LeCun}{Chopra
  et~al\mbox{.}}{2005a}]%
        {chopra2005learning}
\bibfield{author}{\bibinfo{person}{Sumit Chopra}, \bibinfo{person}{Raia
  Hadsell}, {and} \bibinfo{person}{Yann LeCun}.}
  \bibinfo{year}{2005}\natexlab{a}.
\newblock \showarticletitle{Learning a similarity metric discriminatively, with
  application to face verification}. In \bibinfo{booktitle}{\emph{2005 IEEE
  Computer Society Conference on Computer Vision and Pattern Recognition
  (CVPR'05)}}, Vol.~\bibinfo{volume}{1}. IEEE, \bibinfo{pages}{539--546}.
\newblock


\bibitem[\protect\citeauthoryear{Chopra, Hadsell, and LeCun}{Chopra
  et~al\mbox{.}}{2005b}]%
        {chopra2005face}
\bibfield{author}{\bibinfo{person}{S. Chopra}, \bibinfo{person}{R. Hadsell},
  {and} \bibinfo{person}{Y. LeCun}.} \bibinfo{year}{2005}\natexlab{b}.
\newblock \showarticletitle{Learning a similarity metric discriminatively, with
  application to face verification}. In \bibinfo{booktitle}{\emph{2005 IEEE
  Computer Society Conference on Computer Vision and Pattern Recognition
  (CVPR'05)}}, Vol.~\bibinfo{volume}{1}. \bibinfo{pages}{539--546 vol. 1}.
\newblock
\urldef\tempurl%
\url{https://doi.org/10.1109/CVPR.2005.202}
\showDOI{\tempurl}


\bibitem[\protect\citeauthoryear{Chuang, Robinson, Lin, Torralba, and
  Jegelka}{Chuang et~al\mbox{.}}{2020}]%
        {NEURIPS2020_63c3ddcc}
\bibfield{author}{\bibinfo{person}{Ching-Yao Chuang}, \bibinfo{person}{Joshua
  Robinson}, \bibinfo{person}{Yen-Chen Lin}, \bibinfo{person}{Antonio
  Torralba}, {and} \bibinfo{person}{Stefanie Jegelka}.}
  \bibinfo{year}{2020}\natexlab{}.
\newblock \showarticletitle{Debiased Contrastive Learning}. In
  \bibinfo{booktitle}{\emph{Advances in Neural Information Processing
  Systems}}, \bibfield{editor}{\bibinfo{person}{H.~Larochelle},
  \bibinfo{person}{M.~Ranzato}, \bibinfo{person}{R.~Hadsell},
  \bibinfo{person}{M.~F. Balcan}, {and} \bibinfo{person}{H.~Lin}} (Eds.),
  Vol.~\bibinfo{volume}{33}. \bibinfo{publisher}{Curran Associates, Inc.}
\newblock


\bibitem[\protect\citeauthoryear{Cuturi}{Cuturi}{2013}]%
        {cuturi2013sinkhorn}
\bibfield{author}{\bibinfo{person}{Marco Cuturi}.}
  \bibinfo{year}{2013}\natexlab{}.
\newblock \showarticletitle{Sinkhorn distances: Lightspeed computation of
  optimal transport}.
\newblock \bibinfo{journal}{\emph{Advances in neural information processing
  systems}}  \bibinfo{volume}{26} (\bibinfo{year}{2013}).
\newblock


\bibitem[\protect\citeauthoryear{De~Brouwer, Simm, Arany, and
  Moreau}{De~Brouwer et~al\mbox{.}}{2019}]%
        {de2019gru}
\bibfield{author}{\bibinfo{person}{Edward De~Brouwer}, \bibinfo{person}{Jaak
  Simm}, \bibinfo{person}{Adam Arany}, {and} \bibinfo{person}{Yves Moreau}.}
  \bibinfo{year}{2019}\natexlab{}.
\newblock \showarticletitle{Gru-ode-bayes: Continuous modeling of
  sporadically-observed time series}.
\newblock \bibinfo{journal}{\emph{Advances in neural information processing
  systems}}  \bibinfo{volume}{32} (\bibinfo{year}{2019}).
\newblock


\bibitem[\protect\citeauthoryear{de~Sa}{de~Sa}{1994}]%
        {de1994learning}
\bibfield{author}{\bibinfo{person}{Virginia~R de Sa}.}
  \bibinfo{year}{1994}\natexlab{}.
\newblock \showarticletitle{Learning classification with unlabeled data}.
\newblock \bibinfo{journal}{\emph{Advances in neural information processing
  systems}} (\bibinfo{year}{1994}), \bibinfo{pages}{112--112}.
\newblock


\bibitem[\protect\citeauthoryear{Deldari, Smith, Xue, and Salim}{Deldari
  et~al\mbox{.}}{2021}]%
        {deldari2021tscp2}
\bibfield{author}{\bibinfo{person}{Shohreh Deldari}, \bibinfo{person}{Daniel~V.
  Smith}, \bibinfo{person}{Hao Xue}, {and} \bibinfo{person}{Flora~D. Salim}.}
  \bibinfo{year}{2021}\natexlab{}.
\newblock \showarticletitle{Time Series Change Point Detection with
  Self-Supervised Contrastive Predictive Coding}. In
  \bibinfo{booktitle}{\emph{Proceedings of The Web Conference 2021}}
  \emph{(\bibinfo{series}{WWW '21})}. \bibinfo{publisher}{Association for
  Computing Machinery}.
\newblock
\urldef\tempurl%
\url{https://doi.org/10.1145/3442381.3449903}
\showDOI{\tempurl}


\bibitem[\protect\citeauthoryear{Desai and Johnson}{Desai and Johnson}{2021}]%
        {desai2021virtex}
\bibfield{author}{\bibinfo{person}{Karan Desai} {and} \bibinfo{person}{Justin
  Johnson}.} \bibinfo{year}{2021}\natexlab{}.
\newblock \showarticletitle{Virtex: Learning visual representations from
  textual annotations}. In \bibinfo{booktitle}{\emph{Proceedings of the
  IEEE/CVF Conference on Computer Vision and Pattern Recognition}}.
  \bibinfo{pages}{11162--11173}.
\newblock


\bibitem[\protect\citeauthoryear{Devlin, Chang, Lee, and Toutanova}{Devlin
  et~al\mbox{.}}{2019}]%
        {devlin2018bert}
\bibfield{author}{\bibinfo{person}{Jacob Devlin}, \bibinfo{person}{Ming{-}Wei
  Chang}, \bibinfo{person}{Kenton Lee}, {and} \bibinfo{person}{Kristina
  Toutanova}.} \bibinfo{year}{2019}\natexlab{}.
\newblock \showarticletitle{{BERT:} Pre-training of Deep Bidirectional
  Transformers for Language Understanding}. In \bibinfo{booktitle}{\emph{Proc.
  of the Conf. of the North American Chapter of the Assoc. for Computational
  Linguistics: Human Lang. Tech., {NAACL-HLT} 2019}}.
\newblock


\bibitem[\protect\citeauthoryear{Donahue, Kr{\"a}henb{\"u}hl, and
  Darrell}{Donahue et~al\mbox{.}}{2016}]%
        {donahue2016bigan}
\bibfield{author}{\bibinfo{person}{Jeff Donahue}, \bibinfo{person}{Philipp
  Kr{\"a}henb{\"u}hl}, {and} \bibinfo{person}{Trevor Darrell}.}
  \bibinfo{year}{2016}\natexlab{}.
\newblock \showarticletitle{Adversarial feature learning}.
\newblock \bibinfo{journal}{\emph{arXiv preprint arXiv:1605.09782}}
  (\bibinfo{year}{2016}).
\newblock


\bibitem[\protect\citeauthoryear{Donahue and Simonyan}{Donahue and
  Simonyan}{[n.d.]}]%
        {donahue2019large}
\bibfield{author}{\bibinfo{person}{Jeff Donahue} {and} \bibinfo{person}{Karen
  Simonyan}.} \bibinfo{year}{[n.d.]}\natexlab{}.
\newblock \showarticletitle{Large scale adversarial representation learning}.
\newblock \bibinfo{journal}{\emph{NeurIPS}}  \bibinfo{volume}{32}
  (\bibinfo{year}{[n.\,d.]}).
\newblock


\bibitem[\protect\citeauthoryear{Dosovitskiy, Springenberg, Riedmiller, and
  Brox}{Dosovitskiy et~al\mbox{.}}{2014}]%
        {dosovitskiy2014examplarcnn}
\bibfield{author}{\bibinfo{person}{Alexey Dosovitskiy},
  \bibinfo{person}{Jost~Tobias Springenberg}, \bibinfo{person}{Martin
  Riedmiller}, {and} \bibinfo{person}{Thomas Brox}.}
  \bibinfo{year}{2014}\natexlab{}.
\newblock \showarticletitle{Discriminative unsupervised feature learning with
  convolutional neural networks}.
\newblock \bibinfo{journal}{\emph{Advances in neural information processing
  systems}}  \bibinfo{volume}{27} (\bibinfo{year}{2014}).
\newblock


\bibitem[\protect\citeauthoryear{Duan, Chen, Lu, and Zhou}{Duan
  et~al\mbox{.}}{2019}]%
        {duan2019deep}
\bibfield{author}{\bibinfo{person}{Yueqi Duan}, \bibinfo{person}{Lei Chen},
  \bibinfo{person}{Jiwen Lu}, {and} \bibinfo{person}{Jie Zhou}.}
  \bibinfo{year}{2019}\natexlab{}.
\newblock \showarticletitle{Deep embedding learning with discriminative
  sampling policy}. In \bibinfo{booktitle}{\emph{Proceedings of the IEEE
  Conference on Computer Vision and Pattern Recognition}}.
  \bibinfo{pages}{4964--4973}.
\newblock


\bibitem[\protect\citeauthoryear{Eldele, Ragab, Chen, Wu, Kwoh, Li, and
  Guan}{Eldele et~al\mbox{.}}{2021}]%
        {eldele2021tstcc}
\bibfield{author}{\bibinfo{person}{Emadeldeen Eldele}, \bibinfo{person}{Mohamed
  Ragab}, \bibinfo{person}{Zhenghua Chen}, \bibinfo{person}{Min Wu},
  \bibinfo{person}{Chee~Keong Kwoh}, \bibinfo{person}{Xiaoli Li}, {and}
  \bibinfo{person}{Cuntai Guan}.} \bibinfo{year}{2021}\natexlab{}.
\newblock \showarticletitle{Time-Series Representation Learning via Temporal
  and Contextual Contrasting}. In \bibinfo{booktitle}{\emph{Proceedings of the
  Thirtieth International Joint Conference on Artificial Intelligence,
  {IJCAI-21}}}.
\newblock


\bibitem[\protect\citeauthoryear{Ericsson, Gouk, Loy, and Hospedales}{Ericsson
  et~al\mbox{.}}{2021}]%
        {ericsson2021survey}
\bibfield{author}{\bibinfo{person}{Linus Ericsson}, \bibinfo{person}{Henry
  Gouk}, \bibinfo{person}{Chen~Change Loy}, {and} \bibinfo{person}{Timothy~M
  Hospedales}.} \bibinfo{year}{2021}\natexlab{}.
\newblock \showarticletitle{Self-Supervised Representation Learning:
  Introduction, Advances and Challenges}.
\newblock \bibinfo{journal}{\emph{arXiv preprint arXiv:2110.09327}}
  (\bibinfo{year}{2021}).
\newblock


\bibitem[\protect\citeauthoryear{Ermolov, Siarohin, Sangineto, and
  Sebe}{Ermolov et~al\mbox{.}}{2021}]%
        {ermolov2021wmse}
\bibfield{author}{\bibinfo{person}{Aleksandr Ermolov},
  \bibinfo{person}{Aliaksandr Siarohin}, \bibinfo{person}{Enver Sangineto},
  {and} \bibinfo{person}{Nicu Sebe}.} \bibinfo{year}{2021}\natexlab{}.
\newblock \showarticletitle{Whitening for Self-Supervised Representation
  Learning}. In \bibinfo{booktitle}{\emph{Proceedings of the 38th International
  Conference on Machine Learning}} \emph{(\bibinfo{series}{Proceedings of
  Machine Learning Research}, Vol.~\bibinfo{volume}{139})}.
  \bibinfo{publisher}{PMLR}, \bibinfo{pages}{3015--3024}.
\newblock


\bibitem[\protect\citeauthoryear{Favory, Drossos, Virtanen, and Serra}{Favory
  et~al\mbox{.}}{2020}]%
        {favory2020coala}
\bibfield{author}{\bibinfo{person}{Xavier Favory},
  \bibinfo{person}{Konstantinos Drossos}, \bibinfo{person}{Tuomas Virtanen},
  {and} \bibinfo{person}{Xavier Serra}.} \bibinfo{year}{2020}\natexlab{}.
\newblock \showarticletitle{Coala: Co-aligned autoencoders for learning
  semantically enriched audio representations}.
\newblock \bibinfo{journal}{\emph{arXiv preprint arXiv:2006.08386}}
  (\bibinfo{year}{2020}).
\newblock


\bibitem[\protect\citeauthoryear{Favory, Drossos, Virtanen, and Serra}{Favory
  et~al\mbox{.}}{2021}]%
        {favory2021learning}
\bibfield{author}{\bibinfo{person}{Xavier Favory},
  \bibinfo{person}{Konstantinos Drossos}, \bibinfo{person}{Tuomas Virtanen},
  {and} \bibinfo{person}{Xavier Serra}.} \bibinfo{year}{2021}\natexlab{}.
\newblock \showarticletitle{Learning contextual tag embeddings for cross-modal
  alignment of audio and tags}. In \bibinfo{booktitle}{\emph{ICASSP 2021-2021
  IEEE International Conference on Acoustics, Speech and Signal Processing
  (ICASSP)}}. IEEE, \bibinfo{pages}{596--600}.
\newblock


\bibitem[\protect\citeauthoryear{Fernando, Bilen, Gavves, and Gould}{Fernando
  et~al\mbox{.}}{2017}]%
        {fernando2017self}
\bibfield{author}{\bibinfo{person}{Basura Fernando}, \bibinfo{person}{Hakan
  Bilen}, \bibinfo{person}{Efstratios Gavves}, {and} \bibinfo{person}{Stephen
  Gould}.} \bibinfo{year}{2017}\natexlab{}.
\newblock \showarticletitle{Self-supervised video representation learning with
  odd-one-out networks}. In \bibinfo{booktitle}{\emph{Proceedings of the IEEE
  conference on computer vision and pattern recognition}}.
  \bibinfo{pages}{3636--3645}.
\newblock


\bibitem[\protect\citeauthoryear{Fetterman and Albrecht}{Fetterman and
  Albrecht}{2020}]%
        {fetterman2020understanding}
\bibfield{author}{\bibinfo{person}{Abe Fetterman} {and} \bibinfo{person}{Josh
  Albrecht}.} \bibinfo{year}{2020}\natexlab{}.
\newblock \showarticletitle{Understanding self-supervised and contrastive
  learning with bootstrap your own latent (BYOL)}.
\newblock  (\bibinfo{year}{2020}).
\newblock
\urldef\tempurl%
\url{https://generallyintelligent.ai/blog/2020-08-24-understanding-self-supervised-contrastive-learning/}
\showURL{%
\tempurl}


\bibitem[\protect\citeauthoryear{Franceschi, Dieuleveut, and Jaggi}{Franceschi
  et~al\mbox{.}}{2019}]%
        {franceschi2019unsupervised}
\bibfield{author}{\bibinfo{person}{Jean-Yves Franceschi},
  \bibinfo{person}{Aymeric Dieuleveut}, {and} \bibinfo{person}{Martin Jaggi}.}
  \bibinfo{year}{2019}\natexlab{}.
\newblock \showarticletitle{Unsupervised scalable representation learning for
  multivariate time series}. In \bibinfo{booktitle}{\emph{Advances in Neural
  Information Processing Systems}}. \bibinfo{pages}{4650--4661}.
\newblock


\bibitem[\protect\citeauthoryear{Frosst, Papernot, and Hinton}{Frosst
  et~al\mbox{.}}{2019}]%
        {frosst2019analyzing}
\bibfield{author}{\bibinfo{person}{Nicholas Frosst}, \bibinfo{person}{Nicolas
  Papernot}, {and} \bibinfo{person}{Geoffrey Hinton}.}
  \bibinfo{year}{2019}\natexlab{}.
\newblock \showarticletitle{Analyzing and improving representations with the
  soft nearest neighbor loss}. In \bibinfo{booktitle}{\emph{International
  conference on machine learning}}. PMLR, \bibinfo{pages}{2012--2020}.
\newblock


\bibitem[\protect\citeauthoryear{Gao, Xue, Shao, Zhao, Qin, Prabowo, Rahaman,
  and Salim}{Gao et~al\mbox{.}}{2022}]%
        {gao2022generative}
\bibfield{author}{\bibinfo{person}{Nan Gao}, \bibinfo{person}{Hao Xue},
  \bibinfo{person}{Wei Shao}, \bibinfo{person}{Sichen Zhao},
  \bibinfo{person}{Kyle~Kai Qin}, \bibinfo{person}{Arian Prabowo},
  \bibinfo{person}{Mohammad~Saiedur Rahaman}, {and} \bibinfo{person}{Flora~D
  Salim}.} \bibinfo{year}{2022}\natexlab{}.
\newblock \showarticletitle{Generative adversarial networks for spatio-temporal
  data: A survey}.
\newblock \bibinfo{journal}{\emph{ACM Transactions on Intelligent Systems and
  Technology (TIST)}} \bibinfo{volume}{13}, \bibinfo{number}{2}
  (\bibinfo{year}{2022}), \bibinfo{pages}{1--25}.
\newblock


\bibitem[\protect\citeauthoryear{Gao, Feris, and Grauman}{Gao
  et~al\mbox{.}}{2018}]%
        {gao2018learning}
\bibfield{author}{\bibinfo{person}{Ruohan Gao}, \bibinfo{person}{Rogerio
  Feris}, {and} \bibinfo{person}{Kristen Grauman}.}
  \bibinfo{year}{2018}\natexlab{}.
\newblock \showarticletitle{Learning to separate object sounds by watching
  unlabeled video}. In \bibinfo{booktitle}{\emph{Proceedings of the European
  Conference on Computer Vision (ECCV)}}. \bibinfo{pages}{35--53}.
\newblock


\bibitem[\protect\citeauthoryear{Giorgi, Nitski, Wang, and Bader}{Giorgi
  et~al\mbox{.}}{2021}]%
        {giorgi2020declutr}
\bibfield{author}{\bibinfo{person}{John~M. Giorgi}, \bibinfo{person}{Osvald
  Nitski}, \bibinfo{person}{Bo Wang}, {and} \bibinfo{person}{Gary~D. Bader}.}
  \bibinfo{year}{2021}\natexlab{}.
\newblock \showarticletitle{{DeCLUTR}: Deep Contrastive Learning for
  Unsupervised Textual Representations}. In \bibinfo{booktitle}{\emph{Proc. of
  the Annual Meeting of the Assoc. for Computational Linguistics and the
  International Joint Conf. on Natural Lang. Processing, {ACL/IJCNLP}}}.
\newblock


\bibitem[\protect\citeauthoryear{Grill, Strub, Altch{\'{e}}, Tallec, Richemond,
  Buchatskaya, Doersch, Pires, Guo, Azar, Piot, Kavukcuoglu, Munos, and
  Valko}{Grill et~al\mbox{.}}{2020}]%
        {niizumi2021byol}
\bibfield{author}{\bibinfo{person}{Jean{-}Bastien Grill},
  \bibinfo{person}{Florian Strub}, \bibinfo{person}{Florent Altch{\'{e}}},
  \bibinfo{person}{Corentin Tallec}, \bibinfo{person}{Pierre~H. Richemond},
  \bibinfo{person}{Elena Buchatskaya}, \bibinfo{person}{Carl Doersch},
  \bibinfo{person}{Bernardo~{\'{A}}vila Pires}, \bibinfo{person}{Zhaohan Guo},
  \bibinfo{person}{Mohammad~Gheshlaghi Azar}, \bibinfo{person}{Bilal Piot},
  \bibinfo{person}{Koray Kavukcuoglu}, \bibinfo{person}{R{\'{e}}mi Munos},
  {and} \bibinfo{person}{Michal Valko}.} \bibinfo{year}{2020}\natexlab{}.
\newblock \showarticletitle{Bootstrap Your Own Latent - {A} New Approach to
  Self-Supervised Learning}. In \bibinfo{booktitle}{\emph{Advances in Neural
  Information Processing Systems (NeurIPS)}}.
\newblock


\bibitem[\protect\citeauthoryear{Guo, Wang, and Wang}{Guo
  et~al\mbox{.}}{2019}]%
        {guo2019deepmultimodal}
\bibfield{author}{\bibinfo{person}{Wenzhong Guo}, \bibinfo{person}{Jianwen
  Wang}, {and} \bibinfo{person}{Shiping Wang}.}
  \bibinfo{year}{2019}\natexlab{}.
\newblock \showarticletitle{Deep Multimodal Representation Learning: A Survey}.
\newblock \bibinfo{journal}{\emph{IEEE Access}}  \bibinfo{volume}{7}
  (\bibinfo{year}{2019}).
\newblock


\bibitem[\protect\citeauthoryear{Gutman and Hyvarinen}{Gutman and
  Hyvarinen}{2010}]%
        {gut2010NCE}
\bibfield{author}{\bibinfo{person}{Michael Gutman} {and} \bibinfo{person}{Aapo
  Hyvarinen}.} \bibinfo{year}{2010}\natexlab{}.
\newblock \showarticletitle{Noise-contrastive estimation: A new estimation
  principle for unnormalized statistical models}. In
  \bibinfo{booktitle}{\emph{Proceedings of 13th International Conference on
  Artificial Intelligence and Statistics}}. \bibinfo{pages}{297--304}.
\newblock


\bibitem[\protect\citeauthoryear{Guzhov, Raue, Hees, and Dengel}{Guzhov
  et~al\mbox{.}}{2021}]%
        {guzhov2021audioclip}
\bibfield{author}{\bibinfo{person}{Andrey Guzhov}, \bibinfo{person}{Federico
  Raue}, \bibinfo{person}{J{\"o}rn Hees}, {and} \bibinfo{person}{Andreas
  Dengel}.} \bibinfo{year}{2021}\natexlab{}.
\newblock \showarticletitle{AudioCLIP: Extending CLIP to Image, Text and
  Audio}.
\newblock \bibinfo{journal}{\emph{arXiv preprint arXiv:2106.13043}}
  (\bibinfo{year}{2021}).
\newblock


\bibitem[\protect\citeauthoryear{Hadsell, Chopra, and LeCun}{Hadsell
  et~al\mbox{.}}{2006}]%
        {hadsell2006dimensionality}
\bibfield{author}{\bibinfo{person}{Raia Hadsell}, \bibinfo{person}{Sumit
  Chopra}, {and} \bibinfo{person}{Yann LeCun}.}
  \bibinfo{year}{2006}\natexlab{}.
\newblock \showarticletitle{Dimensionality reduction by learning an invariant
  mapping}. In \bibinfo{booktitle}{\emph{2006 IEEE Computer Society Conference
  on Computer Vision and Pattern Recognition (CVPR'06)}},
  Vol.~\bibinfo{volume}{2}. IEEE, \bibinfo{pages}{1735--1742}.
\newblock


\bibitem[\protect\citeauthoryear{Han, Xie, and Zisserman}{Han
  et~al\mbox{.}}{2020}]%
        {han2020coclr}
\bibfield{author}{\bibinfo{person}{Tengda Han}, \bibinfo{person}{Weidi Xie},
  {and} \bibinfo{person}{Andrew Zisserman}.} \bibinfo{year}{2020}\natexlab{}.
\newblock \showarticletitle{Self-supervised co-training for video
  representation learning}.
\newblock \bibinfo{journal}{\emph{Advances in Neural Information Processing
  Systems}}  \bibinfo{volume}{33} (\bibinfo{year}{2020}),
  \bibinfo{pages}{5679--5690}.
\newblock


\bibitem[\protect\citeauthoryear{Haresamudram, Beedu, Agrawal, Grady, Essa,
  Hoffman, and Pl{\"o}tz}{Haresamudram et~al\mbox{.}}{2020}]%
        {haresamudram2020masked}
\bibfield{author}{\bibinfo{person}{Harish Haresamudram},
  \bibinfo{person}{Apoorva Beedu}, \bibinfo{person}{Varun Agrawal},
  \bibinfo{person}{Patrick~L Grady}, \bibinfo{person}{Irfan Essa},
  \bibinfo{person}{Judy Hoffman}, {and} \bibinfo{person}{Thomas Pl{\"o}tz}.}
  \bibinfo{year}{2020}\natexlab{}.
\newblock \showarticletitle{Masked reconstruction based self-supervision for
  human activity recognition}. In \bibinfo{booktitle}{\emph{Proceedings of the
  2020 International Symposium on Wearable Computers}}.
  \bibinfo{pages}{45--49}.
\newblock


\bibitem[\protect\citeauthoryear{Haresamudram, Essa, and
  Pl{\"o}tz}{Haresamudram et~al\mbox{.}}{2021}]%
        {haresamudram2021contrastive}
\bibfield{author}{\bibinfo{person}{Harish Haresamudram}, \bibinfo{person}{Irfan
  Essa}, {and} \bibinfo{person}{Thomas Pl{\"o}tz}.}
  \bibinfo{year}{2021}\natexlab{}.
\newblock \showarticletitle{Contrastive predictive coding for human activity
  recognition}.
\newblock \bibinfo{journal}{\emph{Proceedings of the ACM on Interactive,
  Mobile, Wearable and Ubiquitous Technologies}} \bibinfo{volume}{5},
  \bibinfo{number}{2} (\bibinfo{year}{2021}), \bibinfo{pages}{1--26}.
\newblock


\bibitem[\protect\citeauthoryear{He, Fan, Wu, Xie, and Girshick}{He
  et~al\mbox{.}}{2020}]%
        {he2020momentum}
\bibfield{author}{\bibinfo{person}{Kaiming He}, \bibinfo{person}{Haoqi Fan},
  \bibinfo{person}{Yuxin Wu}, \bibinfo{person}{Saining Xie}, {and}
  \bibinfo{person}{Ross Girshick}.} \bibinfo{year}{2020}\natexlab{}.
\newblock \showarticletitle{Momentum contrast for unsupervised visual
  representation learning}. In \bibinfo{booktitle}{\emph{Proceedings of the
  IEEE/CVF Conference on Computer Vision and Pattern Recognition}}.
  \bibinfo{pages}{9729--9738}.
\newblock


\bibitem[\protect\citeauthoryear{Henaff}{Henaff}{2020}]%
        {henaff2020data}
\bibfield{author}{\bibinfo{person}{Olivier Henaff}.}
  \bibinfo{year}{2020}\natexlab{}.
\newblock \showarticletitle{Data-efficient image recognition with contrastive
  predictive coding}. In \bibinfo{booktitle}{\emph{International Conference on
  Machine Learning}}. PMLR.
\newblock


\bibitem[\protect\citeauthoryear{Hinton, Osindero, and Teh}{Hinton
  et~al\mbox{.}}{2006}]%
        {hinton2006fast}
\bibfield{author}{\bibinfo{person}{Geoffrey~E Hinton}, \bibinfo{person}{Simon
  Osindero}, {and} \bibinfo{person}{Yee-Whye Teh}.}
  \bibinfo{year}{2006}\natexlab{}.
\newblock \showarticletitle{A fast learning algorithm for deep belief nets}.
\newblock \bibinfo{journal}{\emph{Neural computation}} \bibinfo{volume}{18},
  \bibinfo{number}{7} (\bibinfo{year}{2006}).
\newblock


\bibitem[\protect\citeauthoryear{Hjelm, Fedorov, Lavoie-Marchildon, Grewal,
  Bachman, Trischler, and Bengio}{Hjelm et~al\mbox{.}}{2019}]%
        {hjelm2018learning}
\bibfield{author}{\bibinfo{person}{R~Devon Hjelm}, \bibinfo{person}{Alex
  Fedorov}, \bibinfo{person}{Samuel Lavoie-Marchildon}, \bibinfo{person}{Karan
  Grewal}, \bibinfo{person}{Phil Bachman}, \bibinfo{person}{Adam Trischler},
  {and} \bibinfo{person}{Yoshua Bengio}.} \bibinfo{year}{2019}\natexlab{}.
\newblock \showarticletitle{Learning deep representations by mutual information
  estimation and maximization}. In \bibinfo{booktitle}{\emph{International
  Conference on Learning Representations}}.
\newblock


\bibitem[\protect\citeauthoryear{Hsu, Bolte, Tsai, Lakhotia, Salakhutdinov, and
  Mohamed}{Hsu et~al\mbox{.}}{2021}]%
        {hsu2021hubert}
\bibfield{author}{\bibinfo{person}{Wei-Ning Hsu}, \bibinfo{person}{Benjamin
  Bolte}, \bibinfo{person}{Yao-Hung~Hubert Tsai}, \bibinfo{person}{Kushal
  Lakhotia}, \bibinfo{person}{Ruslan Salakhutdinov}, {and}
  \bibinfo{person}{Abdelrahman Mohamed}.} \bibinfo{year}{2021}\natexlab{}.
\newblock \showarticletitle{HuBERT: Self-Supervised Speech Representation
  Learning by Masked Prediction of Hidden Units}.
\newblock \bibinfo{journal}{\emph{arXiv preprint arXiv:2106.07447}}
  (\bibinfo{year}{2021}).
\newblock


\bibitem[\protect\citeauthoryear{Hua, Wang, Xue, Ren, Wang, and Zhao}{Hua
  et~al\mbox{.}}{2021}]%
        {hua2021feature}
\bibfield{author}{\bibinfo{person}{Tianyu Hua}, \bibinfo{person}{Wenxiao Wang},
  \bibinfo{person}{Zihui Xue}, \bibinfo{person}{Sucheng Ren},
  \bibinfo{person}{Yue Wang}, {and} \bibinfo{person}{Hang Zhao}.}
  \bibinfo{year}{2021}\natexlab{}.
\newblock \showarticletitle{On feature decorrelation in self-supervised
  learning}. In \bibinfo{booktitle}{\emph{Proceedings of the IEEE/CVF
  International Conference on Computer Vision}}. \bibinfo{pages}{9598--9608}.
\newblock


\bibitem[\protect\citeauthoryear{Huynh, Kornblith, Walter, Maire, and
  Khademi}{Huynh et~al\mbox{.}}{2022}]%
        {huynh2022boosting}
\bibfield{author}{\bibinfo{person}{Tri Huynh}, \bibinfo{person}{Simon
  Kornblith}, \bibinfo{person}{Matthew~R Walter}, \bibinfo{person}{Michael
  Maire}, {and} \bibinfo{person}{Maryam Khademi}.}
  \bibinfo{year}{2022}\natexlab{}.
\newblock \showarticletitle{Boosting contrastive self-supervised learning with
  false negative cancellation}. In \bibinfo{booktitle}{\emph{Proceedings of the
  IEEE/CVF Winter Conference on Applications of Computer Vision}}.
  \bibinfo{pages}{2785--2795}.
\newblock


\bibitem[\protect\citeauthoryear{Jain, Tang, Min, Kawsar, and Mathur}{Jain
  et~al\mbox{.}}{2022}]%
        {jain2022collossl}
\bibfield{author}{\bibinfo{person}{Yash Jain}, \bibinfo{person}{Chi~Ian Tang},
  \bibinfo{person}{Chulhong Min}, \bibinfo{person}{Fahim Kawsar}, {and}
  \bibinfo{person}{Akhil Mathur}.} \bibinfo{year}{2022}\natexlab{}.
\newblock \showarticletitle{ColloSSL: Collaborative Self-Supervised Learning
  for Human Activity Recognition}.
\newblock \bibinfo{journal}{\emph{Proceedings of the ACM on Interactive,
  Mobile, Wearable and Ubiquitous Technologies}} (\bibinfo{year}{2022}).
\newblock
\urldef\tempurl%
\url{https://doi.org/10.1145/1122445.1122456}
\showDOI{\tempurl}


\bibitem[\protect\citeauthoryear{Jaiswal, ramesh babu, Zadeh, Banerjee, and
  Makedon}{Jaiswal et~al\mbox{.}}{2020}]%
        {Jaiswal2020survey}
\bibfield{author}{\bibinfo{person}{Ashish Jaiswal}, \bibinfo{person}{Ashwin
  ramesh babu}, \bibinfo{person}{Mohammad Zadeh}, \bibinfo{person}{Debapriya
  Banerjee}, {and} \bibinfo{person}{Fillia Makedon}.}
  \bibinfo{year}{2020}\natexlab{}.
\newblock \showarticletitle{A Survey on Contrastive Self-Supervised Learning}.
\newblock \bibinfo{journal}{\emph{Technologies}}  \bibinfo{volume}{9}
  (\bibinfo{date}{12} \bibinfo{year}{2020}), \bibinfo{pages}{2}.
\newblock
\urldef\tempurl%
\url{https://doi.org/10.3390/technologies9010002}
\showDOI{\tempurl}


\bibitem[\protect\citeauthoryear{Jing, Chen, Zhang, He, and Tian}{Jing
  et~al\mbox{.}}{2020}]%
        {jing2020invariant}
\bibfield{author}{\bibinfo{person}{Longlong Jing}, \bibinfo{person}{Yucheng
  Chen}, \bibinfo{person}{Ling Zhang}, \bibinfo{person}{Mingyi He}, {and}
  \bibinfo{person}{Yingli Tian}.} \bibinfo{year}{2020}\natexlab{}.
\newblock \showarticletitle{Self-supervised modal and view invariant feature
  learning}.
\newblock \bibinfo{journal}{\emph{arXiv preprint arXiv:2005.14169}}
  (\bibinfo{year}{2020}).
\newblock


\bibitem[\protect\citeauthoryear{Jing and Tian}{Jing and Tian}{2020}]%
        {jing2020self}
\bibfield{author}{\bibinfo{person}{Longlong Jing} {and} \bibinfo{person}{Yingli
  Tian}.} \bibinfo{year}{2020}\natexlab{}.
\newblock \showarticletitle{Self-supervised visual feature learning with deep
  neural networks: A survey}.
\newblock \bibinfo{journal}{\emph{IEEE transactions on pattern analysis and
  machine intelligence}} (\bibinfo{year}{2020}).
\newblock


\bibitem[\protect\citeauthoryear{Jing, Zhang, and Tian}{Jing
  et~al\mbox{.}}{2021}]%
        {jing2021crossview}
\bibfield{author}{\bibinfo{person}{Longlong Jing}, \bibinfo{person}{Ling
  Zhang}, {and} \bibinfo{person}{Yingli Tian}.}
  \bibinfo{year}{2021}\natexlab{}.
\newblock \showarticletitle{Self-supervised feature learning by cross-modality
  and cross-view correspondences}. In \bibinfo{booktitle}{\emph{Proceedings of
  the IEEE/CVF Conference on Computer Vision and Pattern Recognition}}.
  \bibinfo{pages}{1581--1591}.
\newblock


\bibitem[\protect\citeauthoryear{Kalantidis, Sariyildiz, Pion, Weinzaepfel, and
  Larlus}{Kalantidis et~al\mbox{.}}{2020}]%
        {kalantidis2020hard}
\bibfield{author}{\bibinfo{person}{Yannis Kalantidis},
  \bibinfo{person}{Mert~Bulent Sariyildiz}, \bibinfo{person}{Noe Pion},
  \bibinfo{person}{Philippe Weinzaepfel}, {and} \bibinfo{person}{Diane
  Larlus}.} \bibinfo{year}{2020}\natexlab{}.
\newblock \showarticletitle{Hard negative mixing for contrastive learning}.
\newblock \bibinfo{journal}{\emph{Advances in Neural Information Processing
  Systems}}  \bibinfo{volume}{33} (\bibinfo{year}{2020}),
  \bibinfo{pages}{21798--21809}.
\newblock


\bibitem[\protect\citeauthoryear{Khurana, Laurent, and Glass}{Khurana
  et~al\mbox{.}}{2020}]%
        {khurana2020cstnet}
\bibfield{author}{\bibinfo{person}{Sameer Khurana}, \bibinfo{person}{Antoine
  Laurent}, {and} \bibinfo{person}{James Glass}.}
  \bibinfo{year}{2020}\natexlab{}.
\newblock \showarticletitle{Cstnet: Contrastive speech translation network for
  self-supervised speech representation learning}.
\newblock \bibinfo{journal}{\emph{arXiv preprint arXiv:2006.02814}}
  (\bibinfo{year}{2020}).
\newblock


\bibitem[\protect\citeauthoryear{Kim, Jeong, Kim, Kang, and Kwak}{Kim
  et~al\mbox{.}}{2021a}]%
        {Kim_Jeong2021vqa}
\bibfield{author}{\bibinfo{person}{Seonhoon Kim}, \bibinfo{person}{Seohyeong
  Jeong}, \bibinfo{person}{Eunbyul Kim}, \bibinfo{person}{Inho Kang}, {and}
  \bibinfo{person}{Nojun Kwak}.} \bibinfo{year}{2021}\natexlab{a}.
\newblock \showarticletitle{Self-supervised Pre-training and Contrastive
  Representation Learning for Multiple-choice Video QA}.
\newblock \bibinfo{journal}{\emph{Proceedings of the AAAI Conference on
  Artificial Intelligence}} \bibinfo{volume}{35}, \bibinfo{number}{14}
  (\bibinfo{date}{May} \bibinfo{year}{2021}), \bibinfo{pages}{13171--13179}.
\newblock


\bibitem[\protect\citeauthoryear{Kim, Yoon, Kim, and Kim}{Kim
  et~al\mbox{.}}{2021b}]%
        {kim2021unsupervised}
\bibfield{author}{\bibinfo{person}{Youngji Kim}, \bibinfo{person}{Sungho Yoon},
  \bibinfo{person}{Sujung Kim}, {and} \bibinfo{person}{Ayoung Kim}.}
  \bibinfo{year}{2021}\natexlab{b}.
\newblock \showarticletitle{Unsupervised balanced covariance learning for
  visual-inertial sensor fusion}.
\newblock \bibinfo{journal}{\emph{IEEE Robotics and Automation Letters}}
  \bibinfo{volume}{6}, \bibinfo{number}{2} (\bibinfo{year}{2021}),
  \bibinfo{pages}{819--826}.
\newblock


\bibitem[\protect\citeauthoryear{Kiyasseh, Zhu, and Clifton}{Kiyasseh
  et~al\mbox{.}}{2020}]%
        {kiyasseh2020clocs}
\bibfield{author}{\bibinfo{person}{Dani Kiyasseh}, \bibinfo{person}{Tingting
  Zhu}, {and} \bibinfo{person}{David~A Clifton}.}
  \bibinfo{year}{2020}\natexlab{}.
\newblock \showarticletitle{Clocs: Contrastive learning of cardiac signals}.
\newblock \bibinfo{journal}{\emph{arXiv preprint arXiv:2005.13249}}
  (\bibinfo{year}{2020}).
\newblock


\bibitem[\protect\citeauthoryear{Koch, Zemel, Salakhutdinov,
  et~al\mbox{.}}{Koch et~al\mbox{.}}{2015}]%
        {koch2015siamese}
\bibfield{author}{\bibinfo{person}{Gregory Koch}, \bibinfo{person}{Richard
  Zemel}, \bibinfo{person}{Ruslan Salakhutdinov}, {et~al\mbox{.}}}
  \bibinfo{year}{2015}\natexlab{}.
\newblock \showarticletitle{Siamese neural networks for one-shot image
  recognition}. In \bibinfo{booktitle}{\emph{ICML deep learning workshop}},
  Vol.~\bibinfo{volume}{2}. Lille, \bibinfo{pages}{0}.
\newblock


\bibitem[\protect\citeauthoryear{Korbar, Tran, and Torresani}{Korbar
  et~al\mbox{.}}{2018}]%
        {korbar2018cooperative}
\bibfield{author}{\bibinfo{person}{Bruno Korbar}, \bibinfo{person}{Du Tran},
  {and} \bibinfo{person}{Lorenzo Torresani}.} \bibinfo{year}{2018}\natexlab{}.
\newblock \showarticletitle{Cooperative learning of audio and video models from
  self-supervised synchronization}.
\newblock \bibinfo{journal}{\emph{Advances in Neural Information Processing
  Systems}}  \bibinfo{volume}{31} (\bibinfo{year}{2018}).
\newblock


\bibitem[\protect\citeauthoryear{Latif, Rana, Khalifa, Jurdak, Qadir, and
  Schuller}{Latif et~al\mbox{.}}{2020}]%
        {latif2020speechsurvey}
\bibfield{author}{\bibinfo{person}{Siddique Latif}, \bibinfo{person}{Rajib
  Rana}, \bibinfo{person}{Sara Khalifa}, \bibinfo{person}{Raja Jurdak},
  \bibinfo{person}{Junaid Qadir}, {and} \bibinfo{person}{Bj{\"o}rn~W
  Schuller}.} \bibinfo{year}{2020}\natexlab{}.
\newblock \showarticletitle{Deep representation learning in speech processing:
  Challenges, recent advances, and future trends}.
\newblock \bibinfo{journal}{\emph{arXiv preprint arXiv:2001.00378}}
  (\bibinfo{year}{2020}).
\newblock


\bibitem[\protect\citeauthoryear{Le-Khac, Healy, and Smeaton}{Le-Khac
  et~al\mbox{.}}{2020}]%
        {le2020contrastive}
\bibfield{author}{\bibinfo{person}{Phuc~H Le-Khac}, \bibinfo{person}{Graham
  Healy}, {and} \bibinfo{person}{Alan~F Smeaton}.}
  \bibinfo{year}{2020}\natexlab{}.
\newblock \showarticletitle{Contrastive representation learning: A framework
  and review}.
\newblock \bibinfo{journal}{\emph{IEEE Access}} (\bibinfo{year}{2020}).
\newblock


\bibitem[\protect\citeauthoryear{LeCun and Misra}{LeCun and Misra}{2021}]%
        {lecun2021dark}
\bibfield{author}{\bibinfo{person}{Yan LeCun} {and} \bibinfo{person}{Ishan
  Misra}.} \bibinfo{year}{2021}\natexlab{}.
\newblock \bibinfo{title}{Self-supervised Learning: The Dark Matter of
  Intelligence}.
\newblock
\newblock
\urldef\tempurl%
\url{https://ai.facebook.com/blog/self-supervised-learning-the-dark-matter-of-intelligence/}
\showURL{%
Retrieved Jan 7, 2021 from \tempurl}


\bibitem[\protect\citeauthoryear{Li, Zhou, Xiong, and Hoi}{Li
  et~al\mbox{.}}{2021}]%
        {li2021prototypical}
\bibfield{author}{\bibinfo{person}{Junnan Li}, \bibinfo{person}{Pan Zhou},
  \bibinfo{person}{Caiming Xiong}, {and} \bibinfo{person}{Steven Hoi}.}
  \bibinfo{year}{2021}\natexlab{}.
\newblock \showarticletitle{Prototypical Contrastive Learning of Unsupervised
  Representations}. In \bibinfo{booktitle}{\emph{International Conference on
  Learning Representations}}.
\newblock
\urldef\tempurl%
\url{https://openreview.net/forum?id=KmykpuSrjcq}
\showURL{%
\tempurl}


\bibitem[\protect\citeauthoryear{Li, Hu, Liu, Peng, Zhou, and Peng}{Li
  et~al\mbox{.}}{[n.d.]}]%
        {li2021contrastive}
\bibfield{author}{\bibinfo{person}{Yunfan Li}, \bibinfo{person}{Peng Hu},
  \bibinfo{person}{Zitao Liu}, \bibinfo{person}{Dezhong Peng},
  \bibinfo{person}{Joey~Tianyi Zhou}, {and} \bibinfo{person}{Xi Peng}.}
  \bibinfo{year}{[n.d.]}\natexlab{}.
\newblock \showarticletitle{Contrastive clustering}. In
  \bibinfo{booktitle}{\emph{2021 AAAI}}.
\newblock


\bibitem[\protect\citeauthoryear{Liu, Li, and Lee}{Liu et~al\mbox{.}}{2021a}]%
        {liu2021tera}
\bibfield{author}{\bibinfo{person}{Andy~T Liu}, \bibinfo{person}{Shang-Wen Li},
  {and} \bibinfo{person}{Hung-yi Lee}.} \bibinfo{year}{2021}\natexlab{a}.
\newblock \showarticletitle{Tera: Self-supervised learning of transformer
  encoder representation for speech}.
\newblock \bibinfo{journal}{\emph{IEEE/ACM Transactions on Audio, Speech, and
  Language Processing}}  \bibinfo{volume}{29} (\bibinfo{year}{2021}),
  \bibinfo{pages}{2351--2366}.
\newblock


\bibitem[\protect\citeauthoryear{Liu, Zhang, Hou, Mian, Wang, Zhang, and
  Tang}{Liu et~al\mbox{.}}{2021b}]%
        {liu2021genvscon}
\bibfield{author}{\bibinfo{person}{Xiao Liu}, \bibinfo{person}{Fanjin Zhang},
  \bibinfo{person}{Zhenyu Hou}, \bibinfo{person}{Li Mian},
  \bibinfo{person}{Zhaoyu Wang}, \bibinfo{person}{Jing Zhang}, {and}
  \bibinfo{person}{Jie Tang}.} \bibinfo{year}{2021}\natexlab{b}.
\newblock \showarticletitle{Self-supervised Learning: Generative or
  Contrastive}.
\newblock \bibinfo{journal}{\emph{IEEE Transactions on Knowledge and Data
  Engineering}} (\bibinfo{year}{2021}), \bibinfo{pages}{1--1}.
\newblock
\urldef\tempurl%
\url{https://doi.org/10.1109/TKDE.2021.3090866}
\showDOI{\tempurl}


\bibitem[\protect\citeauthoryear{Lu, Batra, Parikh, and Lee}{Lu
  et~al\mbox{.}}{2019}]%
        {lu2019vilbert}
\bibfield{author}{\bibinfo{person}{Jiasen Lu}, \bibinfo{person}{Dhruv Batra},
  \bibinfo{person}{Devi Parikh}, {and} \bibinfo{person}{Stefan Lee}.}
  \bibinfo{year}{2019}\natexlab{}.
\newblock \showarticletitle{Vilbert: Pretraining task-agnostic visiolinguistic
  representations for vision-and-language tasks}.
\newblock \bibinfo{journal}{\emph{Advances in neural information processing
  systems}}  \bibinfo{volume}{32} (\bibinfo{year}{2019}).
\newblock


\bibitem[\protect\citeauthoryear{Mao}{Mao}{2020}]%
        {mao2020survey}
\bibfield{author}{\bibinfo{person}{Huanru~Henry Mao}.}
  \bibinfo{year}{2020}\natexlab{}.
\newblock \showarticletitle{A survey on self-supervised pre-training for
  sequential transfer learning in neural networks}.
\newblock \bibinfo{journal}{\emph{arXiv preprint arXiv:2007.00800}}
  (\bibinfo{year}{2020}).
\newblock


\bibitem[\protect\citeauthoryear{Miech, Alayrac, Smaira, Laptev, Sivic, and
  Zisserman}{Miech et~al\mbox{.}}{2020}]%
        {Miech_2020_CVPR}
\bibfield{author}{\bibinfo{person}{Antoine Miech},
  \bibinfo{person}{Jean-Baptiste Alayrac}, \bibinfo{person}{Lucas Smaira},
  \bibinfo{person}{Ivan Laptev}, \bibinfo{person}{Josef Sivic}, {and}
  \bibinfo{person}{Andrew Zisserman}.} \bibinfo{year}{2020}\natexlab{}.
\newblock \showarticletitle{End-to-End Learning of Visual Representations From
  Uncurated Instructional Videos}. In \bibinfo{booktitle}{\emph{Proc. of the
  IEEE/CVF Conference on Computer Vision and Pattern Recognition (CVPR)}}.
\newblock


\bibitem[\protect\citeauthoryear{Miech, Zhukov, Alayrac, Tapaswi, Laptev, and
  Sivic}{Miech et~al\mbox{.}}{2019}]%
        {miech2019howto100m}
\bibfield{author}{\bibinfo{person}{Antoine Miech}, \bibinfo{person}{Dimitri
  Zhukov}, \bibinfo{person}{Jean-Baptiste Alayrac}, \bibinfo{person}{Makarand
  Tapaswi}, \bibinfo{person}{Ivan Laptev}, {and} \bibinfo{person}{Josef
  Sivic}.} \bibinfo{year}{2019}\natexlab{}.
\newblock \showarticletitle{Howto100m: Learning a text-video embedding by
  watching hundred million narrated video clips}. In
  \bibinfo{booktitle}{\emph{Proceedings of the IEEE/CVF International
  Conference on Computer Vision}}. \bibinfo{pages}{2630--2640}.
\newblock


\bibitem[\protect\citeauthoryear{Minderer, Bachem, Houlsby, and
  Tschannen}{Minderer et~al\mbox{.}}{2020}]%
        {minderer2020automatic}
\bibfield{author}{\bibinfo{person}{Matthias Minderer}, \bibinfo{person}{Olivier
  Bachem}, \bibinfo{person}{Neil Houlsby}, {and} \bibinfo{person}{Michael
  Tschannen}.} \bibinfo{year}{2020}\natexlab{}.
\newblock \showarticletitle{Automatic shortcut removal for self-supervised
  representation learning}. In \bibinfo{booktitle}{\emph{International
  Conference on Machine Learning}}. PMLR, \bibinfo{pages}{6927--6937}.
\newblock


\bibitem[\protect\citeauthoryear{Misra, Zitnick, and Hebert}{Misra
  et~al\mbox{.}}{2016}]%
        {misra2016shuffle}
\bibfield{author}{\bibinfo{person}{Ishan Misra}, \bibinfo{person}{C~Lawrence
  Zitnick}, {and} \bibinfo{person}{Martial Hebert}.}
  \bibinfo{year}{2016}\natexlab{}.
\newblock \showarticletitle{Shuffle and learn: unsupervised learning using
  temporal order verification}. In \bibinfo{booktitle}{\emph{European
  Conference on Computer Vision}}. Springer, \bibinfo{pages}{527--544}.
\newblock


\bibitem[\protect\citeauthoryear{Mitrovic, McWilliams, and Rey}{Mitrovic
  et~al\mbox{.}}{2020}]%
        {Mitrovic2020less}
\bibfield{author}{\bibinfo{person}{Jovana Mitrovic}, \bibinfo{person}{Brian
  McWilliams}, {and} \bibinfo{person}{Melanie Rey}.}
  \bibinfo{year}{2020}\natexlab{}.
\newblock \showarticletitle{Less can be more in contrastive learning}. In
  \bibinfo{booktitle}{\emph{Proceedings on "I Can't Believe It's Not Better!"
  at NeurIPS Workshops}} \emph{(\bibinfo{series}{Proceedings of Machine
  Learning Research}, Vol.~\bibinfo{volume}{137})}. \bibinfo{publisher}{PMLR},
  \bibinfo{pages}{70--75}.
\newblock


\bibitem[\protect\citeauthoryear{Morgado, Vasconcelos, and Misra}{Morgado
  et~al\mbox{.}}{2021}]%
        {morgado2021audio}
\bibfield{author}{\bibinfo{person}{Pedro Morgado}, \bibinfo{person}{Nuno
  Vasconcelos}, {and} \bibinfo{person}{Ishan Misra}.}
  \bibinfo{year}{2021}\natexlab{}.
\newblock \showarticletitle{Audio-visual instance discrimination with
  cross-modal agreement}. In \bibinfo{booktitle}{\emph{Proceedings of the
  IEEE/CVF Conference on Computer Vision and Pattern Recognition}}.
  \bibinfo{pages}{12475--12486}.
\newblock


\bibitem[\protect\citeauthoryear{Nagrani, Albanie, and Zisserman}{Nagrani
  et~al\mbox{.}}{2018}]%
        {nagrani2018learnable}
\bibfield{author}{\bibinfo{person}{Arsha Nagrani}, \bibinfo{person}{Samuel
  Albanie}, {and} \bibinfo{person}{Andrew Zisserman}.}
  \bibinfo{year}{2018}\natexlab{}.
\newblock \showarticletitle{Learnable pins: Cross-modal embeddings for person
  identity}. In \bibinfo{booktitle}{\emph{Proceedings of the European
  Conference on Computer Vision (ECCV)}}. \bibinfo{pages}{71--88}.
\newblock


\bibitem[\protect\citeauthoryear{Nguyen, Shu, Pham, Bui, and Ermon}{Nguyen
  et~al\mbox{.}}{2021}]%
        {nguyen2021temporal}
\bibfield{author}{\bibinfo{person}{Tung Nguyen}, \bibinfo{person}{Rui Shu},
  \bibinfo{person}{Tuan Pham}, \bibinfo{person}{Hung Bui}, {and}
  \bibinfo{person}{Stefano Ermon}.} \bibinfo{year}{2021}\natexlab{}.
\newblock \showarticletitle{Temporal Predictive Coding For Model-Based Planning
  In Latent Space}. In \bibinfo{booktitle}{\emph{International Conference on
  Machine Learning}}.
\newblock


\bibitem[\protect\citeauthoryear{Oh~Song, Xiang, Jegelka, and Savarese}{Oh~Song
  et~al\mbox{.}}{2016}]%
        {oh2016deep}
\bibfield{author}{\bibinfo{person}{Hyun Oh~Song}, \bibinfo{person}{Yu Xiang},
  \bibinfo{person}{Stefanie Jegelka}, {and} \bibinfo{person}{Silvio Savarese}.}
  \bibinfo{year}{2016}\natexlab{}.
\newblock \showarticletitle{Deep metric learning via lifted structured feature
  embedding}. In \bibinfo{booktitle}{\emph{Proceedings of the IEEE conference
  on computer vision and pattern recognition}}. \bibinfo{pages}{4004--4012}.
\newblock


\bibitem[\protect\citeauthoryear{Oord, Li, and Vinyals}{Oord
  et~al\mbox{.}}{2018}]%
        {oord2018representation}
\bibfield{author}{\bibinfo{person}{Aaron van~den Oord}, \bibinfo{person}{Yazhe
  Li}, {and} \bibinfo{person}{Oriol Vinyals}.} \bibinfo{year}{2018}\natexlab{}.
\newblock \showarticletitle{Representation learning with contrastive predictive
  coding}.
\newblock \bibinfo{journal}{\emph{arXiv preprint}} (\bibinfo{year}{2018}).
\newblock


\bibitem[\protect\citeauthoryear{Ott, R{\"u}gamer, Heublein, Bischl, and
  Mutschler}{Ott et~al\mbox{.}}{2022}]%
        {ott2022cross}
\bibfield{author}{\bibinfo{person}{Felix Ott}, \bibinfo{person}{David
  R{\"u}gamer}, \bibinfo{person}{Lucas Heublein}, \bibinfo{person}{Bernd
  Bischl}, {and} \bibinfo{person}{Christopher Mutschler}.}
  \bibinfo{year}{2022}\natexlab{}.
\newblock \showarticletitle{Cross-Modal Common Representation Learning with
  Triplet Loss Functions}.
\newblock \bibinfo{journal}{\emph{arXiv preprint arXiv:2202.07901}}
  (\bibinfo{year}{2022}).
\newblock


\bibitem[\protect\citeauthoryear{Owens and Efros}{Owens and Efros}{2018}]%
        {owens2018multisensory}
\bibfield{author}{\bibinfo{person}{Andrew Owens} {and}
  \bibinfo{person}{Alexei~A Efros}.} \bibinfo{year}{2018}\natexlab{}.
\newblock \showarticletitle{Audio-visual scene analysis with self-supervised
  multisensory features}. In \bibinfo{booktitle}{\emph{Proceedings of the
  European Conference on Computer Vision (ECCV)}}.
\newblock


\bibitem[\protect\citeauthoryear{Owens, Wu, McDermott, Freeman, and
  Torralba}{Owens et~al\mbox{.}}{2016}]%
        {owens2016ambient}
\bibfield{author}{\bibinfo{person}{Andrew Owens}, \bibinfo{person}{Jiajun Wu},
  \bibinfo{person}{Josh~H McDermott}, \bibinfo{person}{William~T Freeman},
  {and} \bibinfo{person}{Antonio Torralba}.} \bibinfo{year}{2016}\natexlab{}.
\newblock \showarticletitle{Ambient sound provides supervision for visual
  learning}. In \bibinfo{booktitle}{\emph{European conference on computer
  vision}}. Springer, \bibinfo{pages}{801--816}.
\newblock


\bibitem[\protect\citeauthoryear{Pascual, Ravanelli, Serr{\`{a}}, Bonafonte,
  and Bengio}{Pascual et~al\mbox{.}}{[n.d.]}]%
        {pascual2019learning}
\bibfield{author}{\bibinfo{person}{Santiago Pascual}, \bibinfo{person}{Mirco
  Ravanelli}, \bibinfo{person}{Joan Serr{\`{a}}}, \bibinfo{person}{Antonio
  Bonafonte}, {and} \bibinfo{person}{Yoshua Bengio}.}
  \bibinfo{year}{[n.d.]}\natexlab{}.
\newblock \showarticletitle{Learning Problem-Agnostic Speech Representations
  from Multiple Self-Supervised Tasks}. In
  \bibinfo{booktitle}{\emph{Interspeech 2019, 20th Annual Conference of the
  International Speech Communication Association 2019}}.
\newblock


\bibitem[\protect\citeauthoryear{Piergiovanni, Angelova, and Ryoo}{Piergiovanni
  et~al\mbox{.}}{2020}]%
        {piergiovanni2020evolving}
\bibfield{author}{\bibinfo{person}{AJ Piergiovanni}, \bibinfo{person}{Anelia
  Angelova}, {and} \bibinfo{person}{Michael~S Ryoo}.}
  \bibinfo{year}{2020}\natexlab{}.
\newblock \showarticletitle{Evolving losses for unsupervised video
  representation learning}. In \bibinfo{booktitle}{\emph{Proceedings of the
  IEEE/CVF Conference on Computer Vision and Pattern Recognition}}.
  \bibinfo{pages}{133--142}.
\newblock


\bibitem[\protect\citeauthoryear{Qian, Meng, Gong, Yang, Wang, Belongie, and
  Cui}{Qian et~al\mbox{.}}{2021}]%
        {qian2021spatiotemporal}
\bibfield{author}{\bibinfo{person}{Rui Qian}, \bibinfo{person}{Tianjian Meng},
  \bibinfo{person}{Boqing Gong}, \bibinfo{person}{Ming-Hsuan Yang},
  \bibinfo{person}{Huisheng Wang}, \bibinfo{person}{Serge Belongie}, {and}
  \bibinfo{person}{Yin Cui}.} \bibinfo{year}{2021}\natexlab{}.
\newblock \showarticletitle{Spatiotemporal contrastive video representation
  learning}. In \bibinfo{booktitle}{\emph{Proceedings of the IEEE/CVF
  Conference on Computer Vision and Pattern Recognition}}.
  \bibinfo{pages}{6964--6974}.
\newblock


\bibitem[\protect\citeauthoryear{Radford, Kim, Hallacy, Ramesh, Goh, Agarwal,
  Sastry, Askell, Mishkin, Clark, et~al\mbox{.}}{Radford et~al\mbox{.}}{2021}]%
        {radford2021learning}
\bibfield{author}{\bibinfo{person}{Alec Radford}, \bibinfo{person}{Jong~Wook
  Kim}, \bibinfo{person}{Chris Hallacy}, \bibinfo{person}{Aditya Ramesh},
  \bibinfo{person}{Gabriel Goh}, \bibinfo{person}{Sandhini Agarwal},
  \bibinfo{person}{Girish Sastry}, \bibinfo{person}{Amanda Askell},
  \bibinfo{person}{Pamela Mishkin}, \bibinfo{person}{Jack Clark},
  {et~al\mbox{.}}} \bibinfo{year}{2021}\natexlab{}.
\newblock \showarticletitle{Learning transferable visual models from natural
  language supervision}. In \bibinfo{booktitle}{\emph{International Conference
  on Machine Learning}}. PMLR.
\newblock


\bibitem[\protect\citeauthoryear{Rao, Visin, Rusu, Pascanu, Teh, and
  Hadsell}{Rao et~al\mbox{.}}{2019}]%
        {rao2019continual}
\bibfield{author}{\bibinfo{person}{Dushyant Rao}, \bibinfo{person}{Francesco
  Visin}, \bibinfo{person}{Andrei Rusu}, \bibinfo{person}{Razvan Pascanu},
  \bibinfo{person}{Yee~Whye Teh}, {and} \bibinfo{person}{Raia Hadsell}.}
  \bibinfo{year}{2019}\natexlab{}.
\newblock \showarticletitle{Continual Unsupervised Representation Learning}.
\newblock \bibinfo{journal}{\emph{Advances in Neural Information Processing
  Systems}}  \bibinfo{volume}{32} (\bibinfo{year}{2019}),
  \bibinfo{pages}{7647--7657}.
\newblock


\bibitem[\protect\citeauthoryear{Richemond, Grill, Altch{\'e}, Tallec, Strub,
  Brock, Smith, De, Pascanu, Piot, et~al\mbox{.}}{Richemond
  et~al\mbox{.}}{2020}]%
        {richemond2020byol}
\bibfield{author}{\bibinfo{person}{Pierre~H Richemond},
  \bibinfo{person}{Jean-Bastien Grill}, \bibinfo{person}{Florent Altch{\'e}},
  \bibinfo{person}{Corentin Tallec}, \bibinfo{person}{Florian Strub},
  \bibinfo{person}{Andrew Brock}, \bibinfo{person}{Samuel Smith},
  \bibinfo{person}{Soham De}, \bibinfo{person}{Razvan Pascanu},
  \bibinfo{person}{Bilal Piot}, {et~al\mbox{.}}}
  \bibinfo{year}{2020}\natexlab{}.
\newblock \showarticletitle{BYOL works even without batch statistics}.
\newblock \bibinfo{journal}{\emph{arXiv preprint arXiv:2010.10241}}
  (\bibinfo{year}{2020}).
\newblock


\bibitem[\protect\citeauthoryear{Robinson, Chuang, Sra, and Jegelka}{Robinson
  et~al\mbox{.}}{2021}]%
        {robinson2021hardcontrastive}
\bibfield{author}{\bibinfo{person}{Joshua~David Robinson},
  \bibinfo{person}{Ching-Yao Chuang}, \bibinfo{person}{Suvrit Sra}, {and}
  \bibinfo{person}{Stefanie Jegelka}.} \bibinfo{year}{2021}\natexlab{}.
\newblock \showarticletitle{Contrastive Learning with Hard Negative Samples}.
  In \bibinfo{booktitle}{\emph{International Conference on Learning
  Representations}}.
\newblock
\urldef\tempurl%
\url{https://openreview.net/forum?id=CR1XOQ0UTh-}
\showURL{%
\tempurl}


\bibitem[\protect\citeauthoryear{Rubanova, Chen, and Duvenaud}{Rubanova
  et~al\mbox{.}}{2019}]%
        {rubanova2019latent}
\bibfield{author}{\bibinfo{person}{Yulia Rubanova}, \bibinfo{person}{Ricky~TQ
  Chen}, {and} \bibinfo{person}{David~K Duvenaud}.}
  \bibinfo{year}{2019}\natexlab{}.
\newblock \showarticletitle{Latent ordinary differential equations for
  irregularly-sampled time series}.
\newblock \bibinfo{journal}{\emph{Advances in neural information processing
  systems}}  \bibinfo{volume}{32} (\bibinfo{year}{2019}).
\newblock


\bibitem[\protect\citeauthoryear{Saeed, Grangier, and Zeghidour}{Saeed
  et~al\mbox{.}}{2021a}]%
        {saeed2021contrastive}
\bibfield{author}{\bibinfo{person}{Aaqib Saeed}, \bibinfo{person}{David
  Grangier}, {and} \bibinfo{person}{Neil Zeghidour}.}
  \bibinfo{year}{2021}\natexlab{a}.
\newblock \showarticletitle{Contrastive learning of general-purpose audio
  representations}. In \bibinfo{booktitle}{\emph{ICASSP 2021-2021 IEEE
  International Conference on Acoustics, Speech and Signal Processing
  (ICASSP)}}. IEEE, \bibinfo{pages}{3875--3879}.
\newblock


\bibitem[\protect\citeauthoryear{Saeed, Ozcelebi, and Lukkien}{Saeed
  et~al\mbox{.}}{2019}]%
        {saeed2019multi}
\bibfield{author}{\bibinfo{person}{Aaqib Saeed}, \bibinfo{person}{Tanir
  Ozcelebi}, {and} \bibinfo{person}{Johan Lukkien}.}
  \bibinfo{year}{2019}\natexlab{}.
\newblock \showarticletitle{Multi-task Self-Supervised Learning for Human
  Activity Detection}.
\newblock \bibinfo{journal}{\emph{Proceedings of the ACM on Interactive,
  Mobile, Wearable and Ubiquitous Technologies}} \bibinfo{volume}{3},
  \bibinfo{number}{2} (\bibinfo{year}{2019}), \bibinfo{pages}{61}.
\newblock


\bibitem[\protect\citeauthoryear{Saeed, Salim, Ozcelebi, and Lukkien}{Saeed
  et~al\mbox{.}}{2021b}]%
        {saeed2020fed}
\bibfield{author}{\bibinfo{person}{Aaqib Saeed}, \bibinfo{person}{Flora~D.
  Salim}, \bibinfo{person}{Tanir Ozcelebi}, {and} \bibinfo{person}{Johan
  Lukkien}.} \bibinfo{year}{2021}\natexlab{b}.
\newblock \showarticletitle{Federated Self-Supervised Learning of Multisensor
  Representations for Embedded Intelligence}.
\newblock \bibinfo{journal}{\emph{IEEE Internet of Things Journal}}
  \bibinfo{volume}{8}, \bibinfo{number}{2} (\bibinfo{year}{2021}),
  \bibinfo{pages}{1030--1040}.
\newblock
\urldef\tempurl%
\url{https://doi.org/10.1109/JIOT.2020.3009358}
\showDOI{\tempurl}


\bibitem[\protect\citeauthoryear{Saeed, Ungureanu, and Gfeller}{Saeed
  et~al\mbox{.}}{2021c}]%
        {saeed2021sense}
\bibfield{author}{\bibinfo{person}{Aaqib Saeed}, \bibinfo{person}{Victor
  Ungureanu}, {and} \bibinfo{person}{Beat Gfeller}.}
  \bibinfo{year}{2021}\natexlab{c}.
\newblock \showarticletitle{Sense and Learn: Self-supervision for omnipresent
  sensors}.
\newblock \bibinfo{journal}{\emph{Machine Learning with Applications}}
  (\bibinfo{year}{2021}).
\newblock


\bibitem[\protect\citeauthoryear{Sahoo, Kanungo, Behera, and Sabut}{Sahoo
  et~al\mbox{.}}{2017}]%
        {SAHOO201755}
\bibfield{author}{\bibinfo{person}{Santanu Sahoo}, \bibinfo{person}{Bhupen
  Kanungo}, \bibinfo{person}{Suresh Behera}, {and} \bibinfo{person}{Sukanta
  Sabut}.} \bibinfo{year}{2017}\natexlab{}.
\newblock \showarticletitle{Multiresolution wavelet transform based feature
  extraction and ECG classification to detect cardiac abnormalities}.
\newblock \bibinfo{journal}{\emph{Measurement}}  \bibinfo{volume}{108}
  (\bibinfo{year}{2017}), \bibinfo{pages}{55--66}.
\newblock
\showISSN{0263-2241}
\urldef\tempurl%
\url{https://doi.org/10.1016/j.measurement.2017.05.022}
\showDOI{\tempurl}


\bibitem[\protect\citeauthoryear{Schroeder and Foxe}{Schroeder and
  Foxe}{2005}]%
        {schroeder2005multisensory}
\bibfield{author}{\bibinfo{person}{Charles~E Schroeder} {and}
  \bibinfo{person}{John Foxe}.} \bibinfo{year}{2005}\natexlab{}.
\newblock \showarticletitle{Multisensory contributions to
  low-level,‘unisensory’processing}.
\newblock \bibinfo{journal}{\emph{Current opinion in neurobiology}}
  (\bibinfo{year}{2005}).
\newblock


\bibitem[\protect\citeauthoryear{Schroff, Kalenichenko, and Philbin}{Schroff
  et~al\mbox{.}}{2015}]%
        {schroff2015facenet}
\bibfield{author}{\bibinfo{person}{Florian Schroff}, \bibinfo{person}{Dmitry
  Kalenichenko}, {and} \bibinfo{person}{James Philbin}.}
  \bibinfo{year}{2015}\natexlab{}.
\newblock \showarticletitle{Facenet: A unified embedding for face recognition
  and clustering}. In \bibinfo{booktitle}{\emph{Proceedings of the IEEE
  conference on computer vision and pattern recognition}}.
  \bibinfo{pages}{815--823}.
\newblock


\bibitem[\protect\citeauthoryear{Schwarzer, Anand, Goel, Hjelm, Courville, and
  Bachman}{Schwarzer et~al\mbox{.}}{2021}]%
        {schwarzer2020data}
\bibfield{author}{\bibinfo{person}{Max Schwarzer}, \bibinfo{person}{Ankesh
  Anand}, \bibinfo{person}{Rishab Goel}, \bibinfo{person}{R.~Devon Hjelm},
  \bibinfo{person}{Aaron~C. Courville}, {and} \bibinfo{person}{Philip
  Bachman}.} \bibinfo{year}{2021}\natexlab{}.
\newblock \showarticletitle{Data-Efficient Reinforcement Learning with
  Self-Predictive Representations}. In \bibinfo{booktitle}{\emph{9th
  International Conference on Learning Representations, {ICLR} 2021, Virtual
  Event}}.
\newblock


\bibitem[\protect\citeauthoryear{Sermanet, Lynch, Chebotar, Hsu, Jang, Schaal,
  and Levine}{Sermanet et~al\mbox{.}}{2018}]%
        {Sermanet2017TCN}
\bibfield{author}{\bibinfo{person}{Pierre Sermanet}, \bibinfo{person}{Corey
  Lynch}, \bibinfo{person}{Yevgen Chebotar}, \bibinfo{person}{Jasmine Hsu},
  \bibinfo{person}{Eric Jang}, \bibinfo{person}{Stefan Schaal}, {and}
  \bibinfo{person}{Sergey Levine}.} \bibinfo{year}{2018}\natexlab{}.
\newblock \showarticletitle{Time-Contrastive Networks: Self-Supervised Learning
  from Video}.
\newblock \bibinfo{journal}{\emph{Proceedings of International Conference in
  Robotics and Automation (ICRA)}} (\bibinfo{year}{2018}).
\newblock


\bibitem[\protect\citeauthoryear{Sermanet, Lynch, Hsu, and Levine}{Sermanet
  et~al\mbox{.}}{2017}]%
        {sermanet2017tcnmultiview}
\bibfield{author}{\bibinfo{person}{Pierre Sermanet}, \bibinfo{person}{Corey
  Lynch}, \bibinfo{person}{Jasmine Hsu}, {and} \bibinfo{person}{Sergey
  Levine}.} \bibinfo{year}{2017}\natexlab{}.
\newblock \showarticletitle{Time-Contrastive Networks: Self-Supervised Learning
  from Multi-view Observation}. In \bibinfo{booktitle}{\emph{2017 IEEE
  Conference on Computer Vision and Pattern Recognition Workshops (CVPRW)}}.
\newblock
\urldef\tempurl%
\url{https://doi.org/10.1109/CVPRW.2017.69}
\showDOI{\tempurl}


\bibitem[\protect\citeauthoryear{Shi, Paige, Torr, and N}{Shi
  et~al\mbox{.}}{2021}]%
        {shi2021relating}
\bibfield{author}{\bibinfo{person}{Yuge Shi}, \bibinfo{person}{Brooks Paige},
  \bibinfo{person}{Philip Torr}, {and} \bibinfo{person}{Siddharth N}.}
  \bibinfo{year}{2021}\natexlab{}.
\newblock \showarticletitle{Relating by Contrasting: A Data-efficient Framework
  for Multimodal Generative Models}. In \bibinfo{booktitle}{\emph{International
  Conference on Learning Representations}}.
\newblock
\urldef\tempurl%
\url{https://openreview.net/forum?id=vhKe9UFbrJo}
\showURL{%
\tempurl}


\bibitem[\protect\citeauthoryear{Simo-Serra, Trulls, Ferraz, Kokkinos, Fua, and
  Moreno-Noguer}{Simo-Serra et~al\mbox{.}}{2015}]%
        {simo2015discriminative}
\bibfield{author}{\bibinfo{person}{Edgar Simo-Serra}, \bibinfo{person}{Eduard
  Trulls}, \bibinfo{person}{Luis Ferraz}, \bibinfo{person}{Iasonas Kokkinos},
  \bibinfo{person}{Pascal Fua}, {and} \bibinfo{person}{Francesc
  Moreno-Noguer}.} \bibinfo{year}{2015}\natexlab{}.
\newblock \showarticletitle{Discriminative learning of deep convolutional
  feature point descriptors}. In \bibinfo{booktitle}{\emph{Proceedings of the
  IEEE International Conference on Computer Vision}}.
  \bibinfo{pages}{118--126}.
\newblock


\bibitem[\protect\citeauthoryear{Smith and Gasser}{Smith and Gasser}{2005}]%
        {smith2005sixlesson}
\bibfield{author}{\bibinfo{person}{Linda Smith} {and} \bibinfo{person}{Michael
  Gasser}.} \bibinfo{year}{2005}\natexlab{}.
\newblock \showarticletitle{The development of embodied cognition: Six lessons
  from babies}.
\newblock \bibinfo{journal}{\emph{Artificial life}} \bibinfo{volume}{11},
  \bibinfo{number}{1-2} (\bibinfo{year}{2005}), \bibinfo{pages}{13--29}.
\newblock


\bibitem[\protect\citeauthoryear{Sohn}{Sohn}{2016}]%
        {sohn2016improved}
\bibfield{author}{\bibinfo{person}{Kihyuk Sohn}.}
  \bibinfo{year}{2016}\natexlab{}.
\newblock \showarticletitle{Improved deep metric learning with multi-class
  n-pair loss objective}. In \bibinfo{booktitle}{\emph{Advances in neural
  information processing systems}}.
\newblock


\bibitem[\protect\citeauthoryear{Stroud, Lu, Sun, Deng, Sukthankar, Schmid, and
  Ross}{Stroud et~al\mbox{.}}{2020}]%
        {stroud2020learning}
\bibfield{author}{\bibinfo{person}{Jonathan~C Stroud}, \bibinfo{person}{Zhichao
  Lu}, \bibinfo{person}{Chen Sun}, \bibinfo{person}{Jia Deng},
  \bibinfo{person}{Rahul Sukthankar}, \bibinfo{person}{Cordelia Schmid}, {and}
  \bibinfo{person}{David~A Ross}.} \bibinfo{year}{2020}\natexlab{}.
\newblock \showarticletitle{Learning video representations from textual web
  supervision}.
\newblock \bibinfo{journal}{\emph{arXiv preprint arXiv:2007.14937}}
  (\bibinfo{year}{2020}).
\newblock


\bibitem[\protect\citeauthoryear{Sun, Myers, Vondrick, Murphy, and Schmid}{Sun
  et~al\mbox{.}}{2019}]%
        {sun2019videobert}
\bibfield{author}{\bibinfo{person}{Chen Sun}, \bibinfo{person}{Austin Myers},
  \bibinfo{person}{Carl Vondrick}, \bibinfo{person}{Kevin Murphy}, {and}
  \bibinfo{person}{Cordelia Schmid}.} \bibinfo{year}{2019}\natexlab{}.
\newblock \showarticletitle{Videobert: A joint model for video and language
  representation learning}. In \bibinfo{booktitle}{\emph{Proceedings of the
  IEEE/CVF International Conference on Computer Vision}}.
  \bibinfo{pages}{7464--7473}.
\newblock


\bibitem[\protect\citeauthoryear{Tamkin, Liu, Lu, Fein, Schultz, and
  Goodman}{Tamkin et~al\mbox{.}}{2021}]%
        {tamkin2021dabs}
\bibfield{author}{\bibinfo{person}{Alex Tamkin}, \bibinfo{person}{Vincent Liu},
  \bibinfo{person}{Rongfei Lu}, \bibinfo{person}{Daniel Fein},
  \bibinfo{person}{Colin Schultz}, {and} \bibinfo{person}{Noah Goodman}.}
  \bibinfo{year}{2021}\natexlab{}.
\newblock \showarticletitle{{DABS}: a Domain-Agnostic Benchmark for
  Self-Supervised Learning}. In \bibinfo{booktitle}{\emph{Thirty-fifth
  Conference on Neural Information Processing Systems Datasets and Benchmarks
  Track (Round 1)}}.
\newblock


\bibitem[\protect\citeauthoryear{Tao, Wang, and Yamasaki}{Tao
  et~al\mbox{.}}{2020}]%
        {tao2020self}
\bibfield{author}{\bibinfo{person}{Li Tao}, \bibinfo{person}{Xueting Wang},
  {and} \bibinfo{person}{Toshihiko Yamasaki}.} \bibinfo{year}{2020}\natexlab{}.
\newblock \showarticletitle{Self-supervised video representation learning using
  inter-intra contrastive framework}. In \bibinfo{booktitle}{\emph{Proceedings
  of the 28th ACM International Conference on Multimedia}}.
  \bibinfo{pages}{2193--2201}.
\newblock


\bibitem[\protect\citeauthoryear{Tian, Henaff, and van~den Oord}{Tian
  et~al\mbox{.}}{2021}]%
        {tian2021divide}
\bibfield{author}{\bibinfo{person}{Yonglong Tian}, \bibinfo{person}{Olivier~J
  Henaff}, {and} \bibinfo{person}{A{\"a}ron van~den Oord}.}
  \bibinfo{year}{2021}\natexlab{}.
\newblock \showarticletitle{Divide and contrast: Self-supervised learning from
  uncurated data}. In \bibinfo{booktitle}{\emph{Proceedings of the IEEE/CVF
  International Conference on Computer Vision}}. \bibinfo{pages}{10063--10074}.
\newblock


\bibitem[\protect\citeauthoryear{Tian, Krishnan, and Isola}{Tian
  et~al\mbox{.}}{2020a}]%
        {tian2020contrastive}
\bibfield{author}{\bibinfo{person}{Yonglong Tian}, \bibinfo{person}{Dilip
  Krishnan}, {and} \bibinfo{person}{Phillip Isola}.}
  \bibinfo{year}{2020}\natexlab{a}.
\newblock \showarticletitle{Contrastive Multiview Coding}. In
  \bibinfo{booktitle}{\emph{Computer Vision -- ECCV 2020}},
  \bibfield{editor}{\bibinfo{person}{Andrea Vedaldi}, \bibinfo{person}{Horst
  Bischof}, \bibinfo{person}{Thomas Brox}, {and} \bibinfo{person}{Jan-Michael
  Frahm}} (Eds.). \bibinfo{publisher}{Springer International Publishing},
  \bibinfo{address}{Cham}, \bibinfo{pages}{776--794}.
\newblock
\showISBNx{978-3-030-58621-8}


\bibitem[\protect\citeauthoryear{Tian, Sun, Poole, Krishnan, Schmid, and
  Isola}{Tian et~al\mbox{.}}{2020b}]%
        {tian2020makes}
\bibfield{author}{\bibinfo{person}{Yonglong Tian}, \bibinfo{person}{Chen Sun},
  \bibinfo{person}{Ben Poole}, \bibinfo{person}{Dilip Krishnan},
  \bibinfo{person}{Cordelia Schmid}, {and} \bibinfo{person}{Phillip Isola}.}
  \bibinfo{year}{2020}\natexlab{b}.
\newblock \showarticletitle{What makes for good views for contrastive
  learning?}
\newblock \bibinfo{journal}{\emph{NeurIPS}} (\bibinfo{year}{2020}).
\newblock
\urldef\tempurl%
\url{https://ai.googleblog.com/2020/08/understanding-view-selection-for.html}
\showURL{%
\tempurl}


\bibitem[\protect\citeauthoryear{Tian, Yu, Chen, and Ganguli}{Tian
  et~al\mbox{.}}{2020c}]%
        {tian2020understanding}
\bibfield{author}{\bibinfo{person}{Yuandong Tian}, \bibinfo{person}{Lantao Yu},
  \bibinfo{person}{Xinlei Chen}, {and} \bibinfo{person}{Surya Ganguli}.}
  \bibinfo{year}{2020}\natexlab{c}.
\newblock \showarticletitle{Understanding self-supervised learning with dual
  deep networks}.
\newblock \bibinfo{journal}{\emph{arXiv preprint arXiv:2010.00578}}
  (\bibinfo{year}{2020}).
\newblock


\bibitem[\protect\citeauthoryear{Toering, Gatopoulos, Stol, and Hu}{Toering
  et~al\mbox{.}}{2022}]%
        {toering2022self}
\bibfield{author}{\bibinfo{person}{Martine Toering}, \bibinfo{person}{Ioannis
  Gatopoulos}, \bibinfo{person}{Maarten Stol}, {and}
  \bibinfo{person}{Vincent~Tao Hu}.} \bibinfo{year}{2022}\natexlab{}.
\newblock \showarticletitle{Self-supervised Video Representation Learning with
  Cross-Stream Prototypical Contrasting}. In
  \bibinfo{booktitle}{\emph{Proceedings of the IEEE/CVF Winter Conference on
  Applications of Computer Vision}}. \bibinfo{pages}{108--118}.
\newblock


\bibitem[\protect\citeauthoryear{Tonekaboni, Eytan, and Goldenberg}{Tonekaboni
  et~al\mbox{.}}{2021}]%
        {tonekaboni2021unsupervised}
\bibfield{author}{\bibinfo{person}{Sana Tonekaboni}, \bibinfo{person}{Danny
  Eytan}, {and} \bibinfo{person}{Anna Goldenberg}.}
  \bibinfo{year}{2021}\natexlab{}.
\newblock \showarticletitle{Unsupervised Representation Learning for Time
  Series with Temporal Neighborhood Coding}. In
  \bibinfo{booktitle}{\emph{International Conference on Learning
  Representations}}.
\newblock
\urldef\tempurl%
\url{https://openreview.net/forum?id=8qDwejCuCN}
\showURL{%
\tempurl}


\bibitem[\protect\citeauthoryear{Tononi, Sporns, and Edelman}{Tononi
  et~al\mbox{.}}{1999}]%
        {tononi1999degeneracy}
\bibfield{author}{\bibinfo{person}{Giulio Tononi}, \bibinfo{person}{Olaf
  Sporns}, {and} \bibinfo{person}{Gerald~M Edelman}.}
  \bibinfo{year}{1999}\natexlab{}.
\newblock \showarticletitle{Measures of degeneracy and redundancy in biological
  networks}.
\newblock \bibinfo{journal}{\emph{Proceedings of the National Academy of
  Sciences}} \bibinfo{volume}{96}, \bibinfo{number}{6} (\bibinfo{year}{1999}),
  \bibinfo{pages}{3257--3262}.
\newblock


\bibitem[\protect\citeauthoryear{Wang and Liu}{Wang and Liu}{2021}]%
        {Wang_2021_understandingcl}
\bibfield{author}{\bibinfo{person}{Feng Wang} {and} \bibinfo{person}{Huaping
  Liu}.} \bibinfo{year}{2021}\natexlab{}.
\newblock \showarticletitle{Understanding the Behaviour of Contrastive Loss}.
  In \bibinfo{booktitle}{\emph{Proceedings of the IEEE/CVF Conference on
  Computer Vision and Pattern Recognition (CVPR)}}.
  \bibinfo{pages}{2495--2504}.
\newblock


\bibitem[\protect\citeauthoryear{Wang, Lam, Su, and Yu}{Wang
  et~al\mbox{.}}{2021a}]%
        {wang2021contrastive}
\bibfield{author}{\bibinfo{person}{Jun Wang}, \bibinfo{person}{Max W~Y Lam},
  \bibinfo{person}{Dan Su}, {and} \bibinfo{person}{Dong Yu}.}
  \bibinfo{year}{2021}\natexlab{a}.
\newblock \showarticletitle{Contrastive Separative Coding for Self-Supervised
  Representation Learning}. In \bibinfo{booktitle}{\emph{ICASSP 2021-2021 IEEE
  International Conference on Acoustics, Speech and Signal Processing
  (ICASSP)}}. IEEE, \bibinfo{pages}{3865--3869}.
\newblock


\bibitem[\protect\citeauthoryear{Wang, Luc, Recasens, Alayrac, and Oord}{Wang
  et~al\mbox{.}}{2021b}]%
        {wang2021multimodal}
\bibfield{author}{\bibinfo{person}{Luyu Wang}, \bibinfo{person}{Pauline Luc},
  \bibinfo{person}{Adria Recasens}, \bibinfo{person}{Jean-Baptiste Alayrac},
  {and} \bibinfo{person}{Aaron van~den Oord}.}
  \bibinfo{year}{2021}\natexlab{b}.
\newblock \showarticletitle{Multimodal Self-Supervised Learning of General
  Audio Representations}.
\newblock \bibinfo{journal}{\emph{arXiv preprint arXiv:2104.12807}}
  (\bibinfo{year}{2021}).
\newblock


\bibitem[\protect\citeauthoryear{Wang and Isola}{Wang and Isola}{2020}]%
        {wang2020understanding}
\bibfield{author}{\bibinfo{person}{Tongzhou Wang} {and}
  \bibinfo{person}{Phillip Isola}.} \bibinfo{year}{2020}\natexlab{}.
\newblock \showarticletitle{Understanding contrastive representation learning
  through alignment and uniformity on the hypersphere}. In
  \bibinfo{booktitle}{\emph{International Conference on Machine Learning}}.
  PMLR, \bibinfo{pages}{9929--9939}.
\newblock


\bibitem[\protect\citeauthoryear{Wang, Hua, Kodirov, Hu, Garnier, and
  Robertson}{Wang et~al\mbox{.}}{2019}]%
        {wang2019ranked}
\bibfield{author}{\bibinfo{person}{Xinshao Wang}, \bibinfo{person}{Yang Hua},
  \bibinfo{person}{Elyor Kodirov}, \bibinfo{person}{Guosheng Hu},
  \bibinfo{person}{Romain Garnier}, {and} \bibinfo{person}{Neil~M Robertson}.}
  \bibinfo{year}{2019}\natexlab{}.
\newblock \showarticletitle{Ranked list loss for deep metric learning}. In
  \bibinfo{booktitle}{\emph{Proceedings of the IEEE Conference on Computer
  Vision and Pattern Recognition}}.
\newblock


\bibitem[\protect\citeauthoryear{Wei, Lim, Zisserman, and Freeman}{Wei
  et~al\mbox{.}}{2018}]%
        {wei2018learning}
\bibfield{author}{\bibinfo{person}{Donglai Wei}, \bibinfo{person}{Joseph~J
  Lim}, \bibinfo{person}{Andrew Zisserman}, {and} \bibinfo{person}{William~T
  Freeman}.} \bibinfo{year}{2018}\natexlab{}.
\newblock \showarticletitle{Learning and using the arrow of time}. In
  \bibinfo{booktitle}{\emph{Proceedings of the IEEE Conference on Computer
  Vision and Pattern Recognition}}. \bibinfo{pages}{8052--8060}.
\newblock


\bibitem[\protect\citeauthoryear{Wen, Sun, Yang, Song, Gao, Wang, and Xu}{Wen
  et~al\mbox{.}}{2021}]%
        {wen2021augmentation}
\bibfield{author}{\bibinfo{person}{Qingsong Wen}, \bibinfo{person}{Liang Sun},
  \bibinfo{person}{Fan Yang}, \bibinfo{person}{Xiaomin Song},
  \bibinfo{person}{Jingkun Gao}, \bibinfo{person}{Xue Wang}, {and}
  \bibinfo{person}{Huan Xu}.} \bibinfo{year}{2021}\natexlab{}.
\newblock \showarticletitle{Time Series Data Augmentation for Deep Learning: A
  Survey}. In \bibinfo{booktitle}{\emph{Proceedings of the Thirtieth
  International Joint Conference on Artificial Intelligence, {IJCAI-21}}},
  \bibfield{editor}{\bibinfo{person}{Zhi-Hua Zhou}} (Ed.).
  \bibinfo{publisher}{International Joint Conferences on Artificial
  Intelligence Organization}, \bibinfo{pages}{4653--4660}.
\newblock
\newblock
\shownote{Survey Track.}


\bibitem[\protect\citeauthoryear{Wu, Manmatha, Smola, and Krahenbuhl}{Wu
  et~al\mbox{.}}{2017}]%
        {wu2017sampling}
\bibfield{author}{\bibinfo{person}{Chao-Yuan Wu}, \bibinfo{person}{R Manmatha},
  \bibinfo{person}{Alexander~J Smola}, {and} \bibinfo{person}{Philipp
  Krahenbuhl}.} \bibinfo{year}{2017}\natexlab{}.
\newblock \showarticletitle{Sampling matters in deep embedding learning}. In
  \bibinfo{booktitle}{\emph{Proceedings of the IEEE International Conference on
  Computer Vision}}.
\newblock


\bibitem[\protect\citeauthoryear{Xiao, Wang, Ye, Zhang, Bu, Zhang, and Wu}{Xiao
  et~al\mbox{.}}{2021}]%
        {xiao2021sleep}
\bibfield{author}{\bibinfo{person}{Qinfeng Xiao}, \bibinfo{person}{Jing Wang},
  \bibinfo{person}{Jianan Ye}, \bibinfo{person}{Hongjun Zhang},
  \bibinfo{person}{Yuyan Bu}, \bibinfo{person}{Yiqiong Zhang}, {and}
  \bibinfo{person}{Hao Wu}.} \bibinfo{year}{2021}\natexlab{}.
\newblock \showarticletitle{Self-Supervised Learning for Sleep Stage
  Classification with Predictive and Discriminative Contrastive Coding}. In
  \bibinfo{booktitle}{\emph{International Conf. on Acoustics, Speech and Signal
  Processing (ICASSP)}}.
\newblock


\bibitem[\protect\citeauthoryear{Xie, Sun, Huang, Tu, and Murphy}{Xie
  et~al\mbox{.}}{2017}]%
        {xie2017rethinking}
\bibfield{author}{\bibinfo{person}{Saining Xie}, \bibinfo{person}{Chen Sun},
  \bibinfo{person}{Jonathan Huang}, \bibinfo{person}{Zhuowen Tu}, {and}
  \bibinfo{person}{Kevin Murphy}.} \bibinfo{year}{2017}\natexlab{}.
\newblock \showarticletitle{Rethinking spatiotemporal feature learning for
  video understanding}.
\newblock \bibinfo{journal}{\emph{arXiv preprint arXiv:1712.04851}}
  \bibinfo{volume}{1}, \bibinfo{number}{2} (\bibinfo{year}{2017}),
  \bibinfo{pages}{5}.
\newblock


\bibitem[\protect\citeauthoryear{Yang, Miech, Sivic, Laptev, and Schmid}{Yang
  et~al\mbox{.}}{2021a}]%
        {yang2021just}
\bibfield{author}{\bibinfo{person}{Antoine Yang}, \bibinfo{person}{Antoine
  Miech}, \bibinfo{person}{Josef Sivic}, \bibinfo{person}{Ivan Laptev}, {and}
  \bibinfo{person}{Cordelia Schmid}.} \bibinfo{year}{2021}\natexlab{a}.
\newblock \showarticletitle{Just ask: Learning to answer questions from
  millions of narrated videos}. In \bibinfo{booktitle}{\emph{Proceedings of the
  IEEE/CVF International Conference on Computer Vision}}.
  \bibinfo{pages}{1686--1697}.
\newblock


\bibitem[\protect\citeauthoryear{Yang, Ramesh, Chitta, Madhvanath, Bernal, and
  Luo}{Yang et~al\mbox{.}}{2017}]%
        {yang2017deep}
\bibfield{author}{\bibinfo{person}{Xitong Yang}, \bibinfo{person}{Palghat
  Ramesh}, \bibinfo{person}{Radha Chitta}, \bibinfo{person}{Sriganesh
  Madhvanath}, \bibinfo{person}{Edgar~A Bernal}, {and} \bibinfo{person}{Jiebo
  Luo}.} \bibinfo{year}{2017}\natexlab{}.
\newblock \showarticletitle{Deep multimodal representation learning from
  temporal data}. In \bibinfo{booktitle}{\emph{Proceedings of the IEEE
  conference on computer vision and pattern recognition}}.
  \bibinfo{pages}{5447--5455}.
\newblock


\bibitem[\protect\citeauthoryear{Yang, Song, King, and Xu}{Yang
  et~al\mbox{.}}{2021b}]%
        {yang2021survey}
\bibfield{author}{\bibinfo{person}{Xiangli Yang}, \bibinfo{person}{Zixing
  Song}, \bibinfo{person}{Irwin King}, {and} \bibinfo{person}{Zenglin Xu}.}
  \bibinfo{year}{2021}\natexlab{b}.
\newblock \showarticletitle{A Survey on Deep Semi-supervised Learning}.
\newblock \bibinfo{journal}{\emph{arXiv preprint arXiv:2103.00550}}
  (\bibinfo{year}{2021}).
\newblock


\bibitem[\protect\citeauthoryear{YM., C., and A.}{YM. et~al\mbox{.}}{2020}]%
        {YM2020Selflabelling}
\bibfield{author}{\bibinfo{person}{Asano YM.}, \bibinfo{person}{Rupprecht C.},
  {and} \bibinfo{person}{Vedaldi A.}} \bibinfo{year}{2020}\natexlab{}.
\newblock \showarticletitle{Self-labelling via simultaneous clustering and
  representation learning}. In \bibinfo{booktitle}{\emph{International
  Conference on Learning Representations}}.
\newblock
\urldef\tempurl%
\url{https://openreview.net/forum?id=Hyx-jyBFPr}
\showURL{%
\tempurl}


\bibitem[\protect\citeauthoryear{Yuan, Lin, Kuen, Zhang, Wang, Maire, Kale, and
  Faieta}{Yuan et~al\mbox{.}}{2021}]%
        {yuan2021multimodal}
\bibfield{author}{\bibinfo{person}{Xin Yuan}, \bibinfo{person}{Zhe Lin},
  \bibinfo{person}{Jason Kuen}, \bibinfo{person}{Jianming Zhang},
  \bibinfo{person}{Yilin Wang}, \bibinfo{person}{Michael Maire},
  \bibinfo{person}{Ajinkya Kale}, {and} \bibinfo{person}{Baldo Faieta}.}
  \bibinfo{year}{2021}\natexlab{}.
\newblock \showarticletitle{Multimodal Contrastive Training for Visual
  Representation Learning}. In \bibinfo{booktitle}{\emph{Proceedings of the
  IEEE/CVF Conference on Computer Vision and Pattern Recognition}}.
  \bibinfo{pages}{6995--7004}.
\newblock


\bibitem[\protect\citeauthoryear{Yue, Wang, Duan, Yang, Huang, Tong, and
  Xu}{Yue et~al\mbox{.}}{2021}]%
        {yue2021ts2vec}
\bibfield{author}{\bibinfo{person}{Zhihan Yue}, \bibinfo{person}{Yujing Wang},
  \bibinfo{person}{Juanyong Duan}, \bibinfo{person}{Tianmeng Yang},
  \bibinfo{person}{Congrui Huang}, \bibinfo{person}{Yunhai Tong}, {and}
  \bibinfo{person}{Bixiong Xu}.} \bibinfo{year}{2021}\natexlab{}.
\newblock \showarticletitle{TS2Vec: Towards Universal Representation of Time
  Series}.
\newblock \bibinfo{journal}{\emph{arXiv preprint arXiv:2106.10466}}
  (\bibinfo{year}{2021}).
\newblock


\bibitem[\protect\citeauthoryear{Zaiem, Parcollet, and Essid}{Zaiem
  et~al\mbox{.}}{2021}]%
        {zaiem2021pretext}
\bibfield{author}{\bibinfo{person}{Salah Zaiem}, \bibinfo{person}{Titouan
  Parcollet}, {and} \bibinfo{person}{Slim Essid}.}
  \bibinfo{year}{2021}\natexlab{}.
\newblock \showarticletitle{Pretext Tasks selection for multitask
  self-supervised speech representation learning}.
\newblock \bibinfo{journal}{\emph{arXiv preprint arXiv:2107.00594}}
  (\bibinfo{year}{2021}).
\newblock


\bibitem[\protect\citeauthoryear{Zbontar, Jing, Misra, LeCun, and Deny}{Zbontar
  et~al\mbox{.}}{2021}]%
        {zbontar2021barlow}
\bibfield{author}{\bibinfo{person}{Jure Zbontar}, \bibinfo{person}{Li Jing},
  \bibinfo{person}{Ishan Misra}, \bibinfo{person}{Yann LeCun}, {and}
  \bibinfo{person}{St{\'{e}}phane Deny}.} \bibinfo{year}{2021}\natexlab{}.
\newblock \showarticletitle{Barlow Twins: Self-Supervised Learning via
  Redundancy Reduction}. In \bibinfo{booktitle}{\emph{Proceedings of the 38th
  International Conference on Machine Learning, {ICML}}},
  \bibfield{editor}{\bibinfo{person}{Marina Meila} {and} \bibinfo{person}{Tong
  Zhang}} (Eds.), Vol.~\bibinfo{volume}{139}. \bibinfo{publisher}{{PMLR}}.
\newblock


\bibitem[\protect\citeauthoryear{Zerveas, Jayaraman, Patel, Bhamidipaty, and
  Eickhoff}{Zerveas et~al\mbox{.}}{2021}]%
        {zerveas2020transformer}
\bibfield{author}{\bibinfo{person}{George Zerveas}, \bibinfo{person}{Srideepika
  Jayaraman}, \bibinfo{person}{Dhaval Patel}, \bibinfo{person}{Anuradha
  Bhamidipaty}, {and} \bibinfo{person}{Carsten Eickhoff}.}
  \bibinfo{year}{2021}\natexlab{}.
\newblock \showarticletitle{A Transformer-Based Framework for Multivariate Time
  Series Representation Learning}. In \bibinfo{booktitle}{\emph{Proceedings of
  the 27th ACM SIGKDD Conference on Knowledge Discovery and Data Mining}}.
\newblock


\bibitem[\protect\citeauthoryear{Zhan, Xie, Liu, Ong, and Loy}{Zhan
  et~al\mbox{.}}{2020}]%
        {zhan2020ODC}
\bibfield{author}{\bibinfo{person}{Xiaohang Zhan}, \bibinfo{person}{Jiahao
  Xie}, \bibinfo{person}{Ziwei Liu}, \bibinfo{person}{Yew-Soon Ong}, {and}
  \bibinfo{person}{Chen~Change Loy}.} \bibinfo{year}{2020}\natexlab{}.
\newblock \showarticletitle{Online deep clustering for unsupervised
  representation learning}. In \bibinfo{booktitle}{\emph{Proceedings of the
  IEEE/CVF conference on computer vision and pattern recognition}}.
  \bibinfo{pages}{6688--6697}.
\newblock


\bibitem[\protect\citeauthoryear{Zhao, Cheng, Zhang, and Peng}{Zhao
  et~al\mbox{.}}{2020}]%
        {zhao2020ecg}
\bibfield{author}{\bibinfo{person}{Yunxiang Zhao}, \bibinfo{person}{Jinyong
  Cheng}, \bibinfo{person}{Ping Zhang}, {and} \bibinfo{person}{Xueping Peng}.}
  \bibinfo{year}{2020}\natexlab{}.
\newblock \showarticletitle{ECG classification using deep CNN improved by
  wavelet transform}.
\newblock \bibinfo{journal}{\emph{Computers, Materials and Continua}}
  (\bibinfo{year}{2020}).
\newblock


\end{thebibliography}



\end{document}